%% file: paper.tex
\documentclass[sigconf]{acmart}

\usepackage{hyperref}
\usepackage{cleveref}
\usepackage{algorithmic}
\usepackage{algorithm}
\usepackage{subcaption}
\usepackage{xcolor}
\usepackage{multirow}

\definecolor{modelinversion}{HTML}{1f77b4}
\definecolor{gradientinversion}{HTML}{ff7f0e}
\definecolor{reference}{HTML}{2ca02c}
\definecolor{random}{HTML}{131111}
\definecolor{drop}{HTML}{EC1919}
\definecolor{noaction}{HTML}{888888}

\AtBeginDocument{%
  }

\setcopyright{rightsretained}
\copyrightyear{2025}
\acmYear{2025}
\acmConference[FedKDD @ KDD '25] {FedKDD: International Joint Workshop on Federated Learning for Data Mining and Graph Analytics, 31st ACM SIGKDD Conference on Knowledge Discovery and Data Mining}{August 3--7, 2025}{Toronto, ON, Canada.}
\acmBooktitle{FedKDD: International Joint Workshop on Federated Learning for Data Mining and Graph Analytics at 31st ACM SIGKDD Conference on Knowledge Discovery and Data Mining, 
 August 3--7, 2025, Toronto, ON, Canada}

\begin{document}

\title[Mitigating Persistent Client Dropout]{Mitigating Persistent Client Dropout\\ 
in Asynchronous Decentralized Federated Learning}

\author{Ignacy St\k{e}pka}
\email{istepka@andrew.cmu.edu}
\orcid{0009-0004-4575-0689}
\affiliation{%
  \institution{Carnegie Mellon University}
  \city{Pittsburgh}
  \state{PA}
  \country{USA}
}

\author{Nick Gisolfi}
\email{ngisolfi@andrew.cmu.edu}
\orcid{0000-0002-9258-6285}
\affiliation{%
  \institution{Carnegie Mellon University}
  \city{Pittsburgh}
  \state{PA}
  \country{USA}
}

\author{Kacper Tr\k{e}bacz}
\email{ktrebacz@andrew.cmu.edu}
\orcid{0009-0006-5615-5935}
\affiliation{%
  \institution{Carnegie Mellon University}
  \city{Pittsburgh}
  \state{PA}
  \country{USA}
}

\author{Artur Dubrawski}
\email{awd@andrew.cmu.edu}
\orcid{0000-0002-2372-0831}
\affiliation{%
  \institution{Carnegie Mellon University}
  \city{Pittsburgh}
  \state{PA}
  \country{USA}
}

\renewcommand{\shortauthors}{St\k{e}pka et al.}

\begin{abstract}
We consider the problem of persistent client dropout in asynchronous Decentralized Federated Learning (DFL).  Asynchronicity and decentralization obfuscate information about model updates among federation peers, making recovery from a client dropout difficult. Access to the number of learning epochs, data distributions, and all the information necessary to precisely reconstruct the missing neighbor's loss functions is limited. We show that obvious mitigations do not adequately address the problem and introduce adaptive strategies based on client reconstruction. We show that these strategies can effectively recover some performance loss caused by dropout. Our work focuses on asynchronous DFL with local regularization and differs substantially from that in the existing literature. We evaluate the proposed methods on tabular and image datasets, involve three DFL algorithms, and three data heterogeneity scenarios (iid, non-iid, class-focused non-iid). Our experiments show that the proposed adaptive strategies can be effective in maintaining robustness of federated learning, even if they do not reconstruct the missing client's data precisely.  We also discuss the limitations and identify future avenues for tackling the problem of client dropout.
\end{abstract}




\maketitle

\input{chapters/1.intro}

\input{chapters/2.related}

\input{chapters/3.background}

\input{chapters/4.methods}

\input{chapters/5.experiments}

\input{chapters/6.conclusions}

\begin{acks}
This work was partially supported by the NSF (awards 2406231 and 2427948), NIH (awards R01NS124642 and R01NR013912), DARPA (HR00112420329), and the US Army (W911NF-20-D0002).
\end{acks}

\bibliographystyle{ACM-Reference-Format}
\bibliography{ref2}

\appendix

\input{chapters/7.0.appendix}

\input{chapters/7.1.appendix}

\input{chapters/7.2.appendix}

\input{chapters/7.3.appendix}

\input{chapters/7.4.appendix}

\input{chapters/7.5.appendix}

\end{document}

%% file: chapters/1.intro.tex
\section{Introduction}
The most widely used form of decentralized learning is Federated Learning (FL), where data remains distributed across clients. Each client performs local training and shares gradients with a central server, which aggregates them using, e.g., the FedAvg algorithm~\cite{mcmahan_communication-efficient_2023} and broadcasts the updated model back to the clients.
In this work, we focus on a decentralized variant of this paradigm, Decentralized Federated Learning (DFL), where no central server exists. Instead, clients interact in a peer-to-peer, asynchronous manner \cite{sun_decentralized_2021}.

One of the core challenges in such systems is unequal client participation. This is especially common in practical applications involving mobile devices or unreliable networks. Unequal participation and the presence of stragglers in synchronous settings have been shown to degrade overall system performance \cite{zhu_client_2023}. While prior work addresses transient client dropout in centralized setups \cite{sun_mimic_2024,shao_dres-fl_2022,wang_friends_2024}, persistent dropout in asynchronous decentralized environments remains unexplored.

DFL environments are characterized by low observability, unknown local step counts, and high communication asynchronicity, rendering many existing solutions incompatible. To this end, we investigate adaptive mitigation strategies that respond to persistent client dropout and aim to recover performance that would otherwise be lost.

Our findings indicate that simply doing nothing after a dropout significantly reduces federation performance. Similarly, na\"{i}ve strategies, such as removing the dropped client, lead to poor outcomes, especially under non-iid data distributions. Motivated by these observations, we propose adaptive strategies that reconstruct the missing client’s data and instantiate a \textit{virtual client} to continue participating in the optimization process.

We employ gradient inversion \cite{geiping_inverting_2020,zhu_deep_2019} and model inversion attacks \cite{haim_reconstructing_2022} to approximate the dropped client’s data. Empirically, we demonstrate the effectiveness of these adaptive methods across three different DFL algorithms and data heterogeneity scenarios.

To sum up, in this work we make two key contributions. First, we identify the persistent client dropout problem in asynchronous DFL under low-information assumptions and introduce mitigation strategies that do not require modifying the core optimization algorithm. Second, we demonstrate that gradient and model inversion attacks can recover useful approximations of lost client data even when gradients reflect multiple local steps and data points, enabling virtual clients to rejoin the optimization and significantly improve final performance.

    
    
    

%% file: chapters/2.related.tex
\section{Related Work}

Our focus is on decentralized learning, where the presence of a central server is explicitly disallowed. Recent advances, such as DFedAvgM, a decentralized FL algorithm employing momentum-based updates~\cite{sun_decentralized_2021}, have extended FL-style learning to fully decentralized settings. In another line of work, each client optimizes its local objective asynchronously and engages in peer-to-peer communication rounds, exchanging local models and incorporating neighbors' models into their optimization processes. Unlike traditional FL approaches that share gradients with a central server, they adopt a model-sharing paradigm with equivalent communication costs.  Notable decentralized optimization approaches include ADMM-based methods \cite{vanhaesebrouck_decentralized_2017}, AIDE \cite{reddi_aide_2016}, DJAM \cite{almeida_djam_2018}, and FSR \cite{good2024trustworthy}. In this work, we use DFedAvgM, DJAM, and FSR as representative baselines to assess the effectiveness of our proposed adaptive strategies across a range of learning paradigms.

\paragraph{Client Dropout and Unequal Participation}

The client dropout problem can be viewed as an extreme case of unequal participation in federated systems, and is closely related to the issue of stragglers \cite{zhu_client_2023}. 
Although it has been investigated in centralized FL settings, the assumptions and architectures in those works are fundamentally different from ours. For instance, MimiC \cite{sun_mimic_2024} proposes a correction-based method where the server adjusts updates to account for absent clients. DReS-FL \cite{shao_dres-fl_2022} introduces a secure mechanism for client replacement via Lagrange-coded data sharing, but it relies on specialized cryptographic primitives and restricts model design (e.g., requiring integer polynomial neural networks). Friends-to-Help \cite{wang_friends_2024} substitutes dropped client updates using neighboring clients with similar data distributions, but assumes transient dropout and the ability to exchange metadata about local data.

Our setting differs from that background in two important ways: (1) we assume persistent client dropout, where a lost client never returns to the federation, and (2) we operate in an asynchronous, decentralized environment without a central server. These make prior methods ill-suited for direct adoption, as they either depend on central orchestration, require modification of the underlying learning algorithms, or are infeasible in low-information regimes. In contrast, our approach works with arbitrary DFL methods and relies only on observing model updates. Additionally, our methods are compatible with setups involving local regularization/aggregation rather than centralized aggregation.

\paragraph{Data Extraction Attacks}

A large body of work has studied data extraction in static centralized learning settings, often by exploiting the loss function of the attacked model~\cite{fang2024privacy}. These attacks typically initialize synthetic inputs and optimize them to minimize model loss, often including domain-specific regularizers such as total variation to enhance the realism of reconstructed samples. Haim et al.~\cite{haim_reconstructing_2022} propose a KKT-based method to reconstruct real training data, under the assumption that the model has converged to a stationary point satisfying optimality conditions. However, enforcing these constraints requires architectural restrictions, which substantially harm the model performance in practice~\cite{ji_directional_2020}.

Gradient inversion attacks offer an alternative by working directly with shared gradients rather than full models. These methods aim to match real and synthetic gradients via second-order optimization, and have been explored in the context of centralized FL \cite{zhu_deep_2019,geiping_inverting_2020,yin_see_2021,wang_beyond_2019}. However, most prior work studies them in idealized scenarios where gradients are exchanged frequently and correspond to single batches or epochs. In contrast, DFL settings involve model updates that are the cumulative result of many local steps, making the inversion problem more difficult.

Further complicating matters, DFL algorithms often include private regularization terms, such as neighbor model alignment in DJAM \cite{almeida_djam_2018} and FSR \cite{good2024trustworthy}, introducing additional noise and obfuscation in the exchanged updates. Thus, applying gradient or model inversion in this context is more challenging and less explored. Our work takes a first step in this direction by empirically evaluating whether such inversion methods can still extract useful proxy data or biases for dropout mitigation in DFL.

%% file: chapters/3.background.tex
\section{Background}
\label{sec:background}







\subsection{System Model}
\label{sec:system-model}

We consider a network of $m$ clients, where each client $i$ holds private data $(X_i, Y_i)$ and trains a local model with parameters $\theta_i$. The collective goal is to minimize the average loss across clients while encouraging alignment between models:

\begin{equation}
\label{eq:global-objective}
    \underset{\theta_1, \dots, \theta_m}{\text{minimize}} \quad \frac{1}{m} \sum_{i=1}^{m} \mathcal{L}_i(\theta_i, X_i, Y_i)
    \quad \text{s.t.} \quad \mathcal{R}(\theta_1, \dots, \theta_m) = 0
\end{equation}

Here, $\mathcal{L}_i$ is the local data loss for client $i$, and $\mathcal{R}$ is a regularization term that encourages consistency between clients. Traditionally, $\mathcal{R}$ is defined in parameter space enforcing $\theta_0 + \theta_1  ... + \theta_{m-1} \approx 0$ as in DJAM or DFedAvgM. However, it can also operate in function space, as in FSR, where alignment is measured in the function space $f_i$, where $f_i := \mathcal{F}(\theta_i)$. $\mathcal{F}(\cdot)$ maps parameters to functions, e.g., a neural network forward pass.

Each client independently minimizes its local loss:
\begin{equation}
\label{eq:local-objective}
    \underset{\theta_i}{\text{minimize}} \quad \mathcal{L}_i(\theta_i, X_i, Y_i)
\end{equation}

\noindent but the choice of a particular optimization strategy varies by algorithm.

\subsection{Algorithms}
We consider three decentralized algorithms that represent distinct design choices in DFL: DJAM, FSR, and DFedAvgM.

\subsubsection{DJAM} 
\cite{almeida_djam_2018} is an asynchronous method that learns personal models through parameter-space regularization. Locally, each client minimizes the following objective:
\begin{equation}
    \mathcal{L}_{\text{DJAM}} = \mathcal{L}_d +  \Vert \theta_i^{t} - \theta^{t-1}_i \Vert_2 + \frac{1}{2} \sum_{j=1}^N g_{ij} \Vert \theta_i - \theta^t_{ij} \Vert_2
\end{equation}
where $\mathcal{L}_d$ is the standard data loss (e.g., cross-entropy). The second and third loss terms are, respectively, for self and neighbor regularization and are calulated as an L2 norm between model parameters.
DJAM requires model design homogeneity across clients to enable the L2-based parameter space regularization.

\subsubsection{DFedAvgM}  
\cite{sun_decentralized_2021} implements FedAvg-style updates in a decentralized and, in our case, asynchronous setting. In each round, a client first averages the latest received models from its neighbors according to the communication graph: 
\begin{equation}
    \theta^{t+1}_i = \sum_{j=1}^N g_{ij} \theta^t_j
\end{equation}
Then, it performs local updates using the data loss $\mathcal{L}_d$, and finally broadcasts the updated model $\theta^{t+1}_i$ to its neighbors.

\subsubsection{FSR}
\cite{good2024trustworthy} supports heterogeneous models by regularizing in function space. Its loss function includes data loss, self-regularization acting as an adaptive learning rate, and neighbor regularization enforcing neighbor similarity in function space:
\begin{equation}
\label{eq:fsr-loss-function}
    \mathcal{L}_{\text{FSR}} = \mathcal{L}_d + \frac{1 - \omega}{\omega} \Vert f^t_i - f^{t-1}_i \Vert_2 + \lambda \frac{1}{N} \sum_{j=1}^N g_{ij} \Vert f_i - f^t_{ij} \Vert_2 
\end{equation}
This allows FSR to operate under model heterogeneity while adapting learning rates.

\subsection{Communication}
\label{sec:communication}

We assume an asynchronous peer-to-peer setting, where clients communicate via a graph $G \in \mathbb{R}^{m \times m}$. Each entry $g_{ij} > 0$ indicates a communication link between clients $i$ and $j$. The optimization proceeds in alternating local and communication rounds. In each communication round, a random pair of connected clients exchange their current models. The procedure is outlined in \Cref{alg:distributed-optimization}.

%% file: chapters/4.methods.tex
\begin{figure*}[htb]
    \centering
    \begin{subfigure}[b]{0.33\textwidth}
        \includegraphics[width=\linewidth]{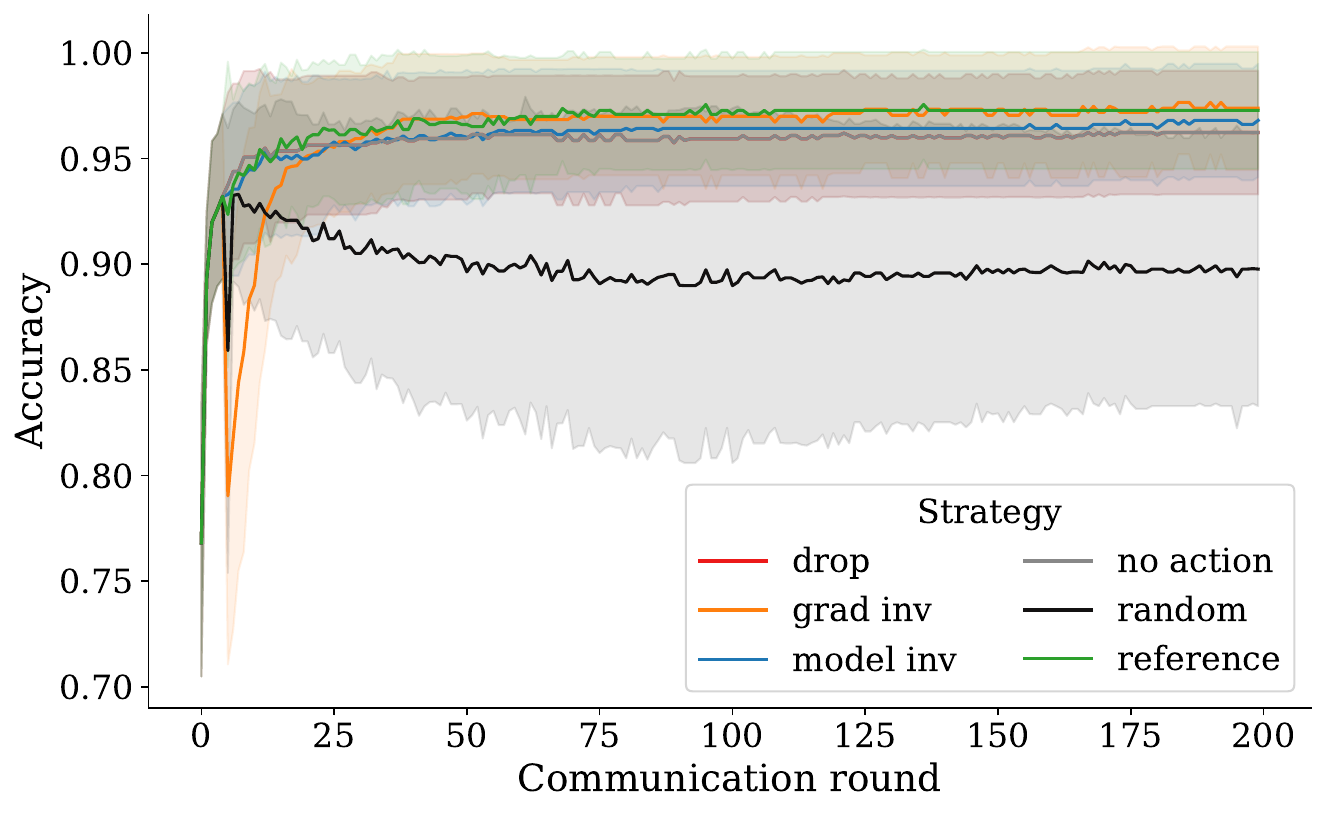}
        \caption{iid}
    \end{subfigure}
    \hfill
    \begin{subfigure}[b]{0.33\textwidth}
        \includegraphics[width=\linewidth]{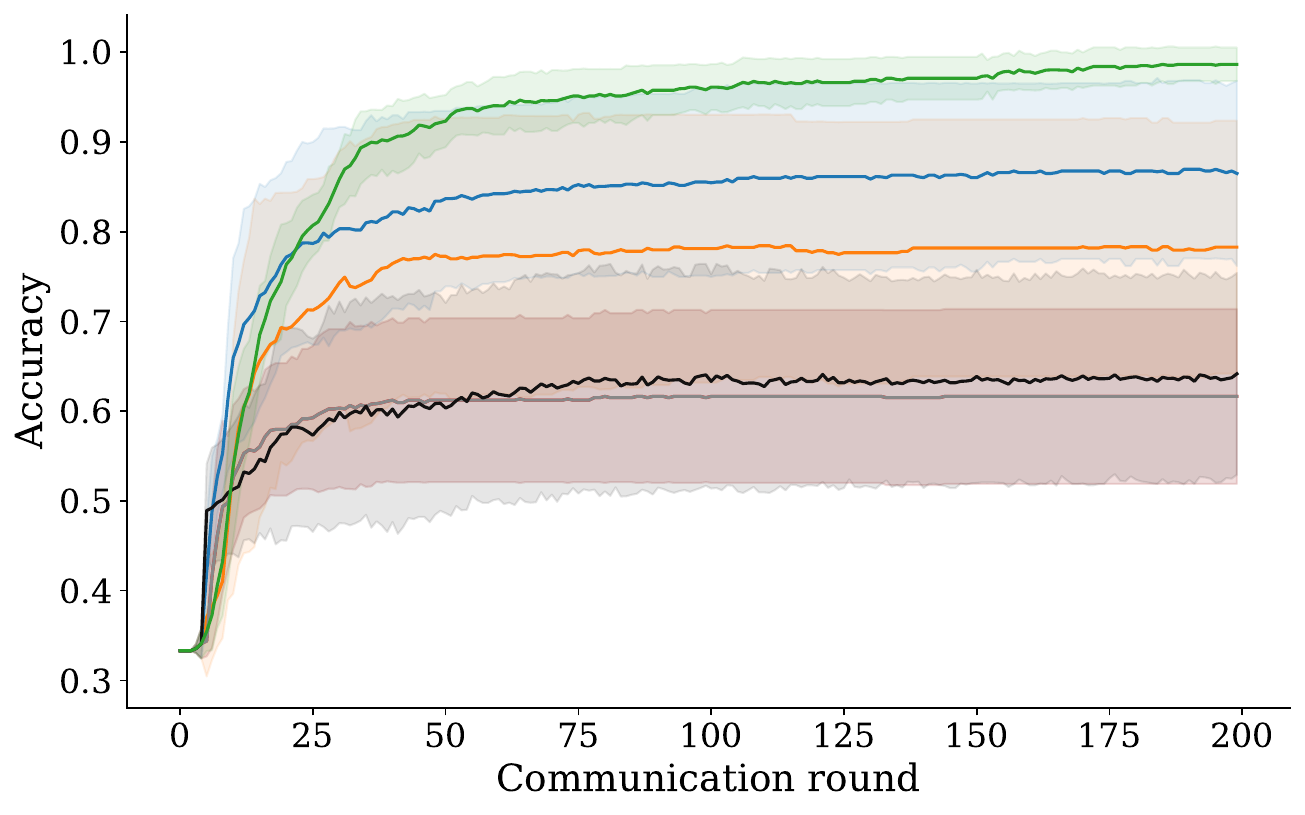}
        \caption{non-iid (clusters)}
    \end{subfigure}
    \hfill
    \begin{subfigure}[b]{0.33\textwidth}
        \includegraphics[width=\linewidth]{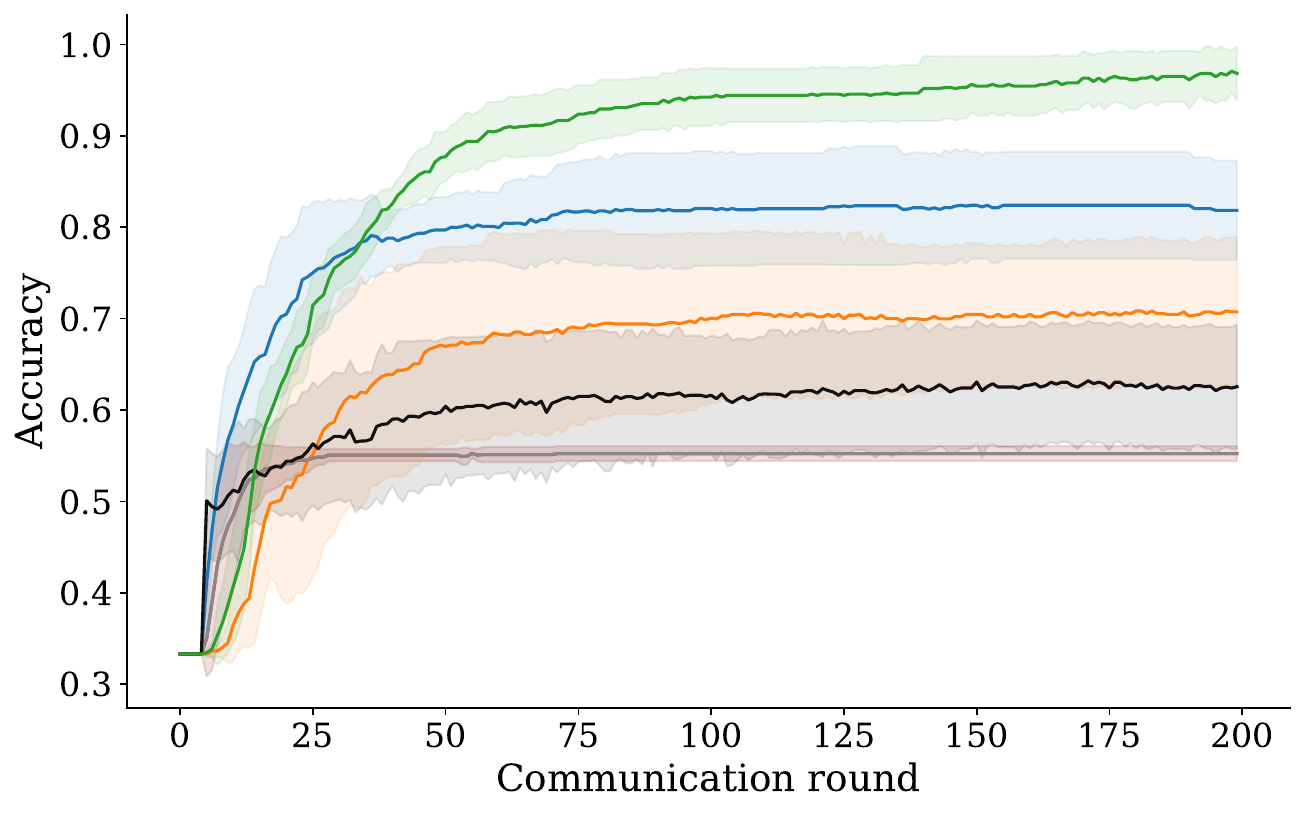}
        \caption{non-iid (classes)}
    \end{subfigure}
    \caption{Convergence plots for DFedAvgM algorithm on the wine dataset. Shaded regions represent mean $\pm$ standard deviation across 10 folds, over 200 communication rounds. A random client is dropped persistently after the 5th round. Colors: {\color{modelinversion}model inversion}, {\color{gradientinversion}gradient inversion}, {\color{reference}reference}, {\color{random}random}, {\color{drop}drop}, {\color{noaction}no action}}.
    \label{fig:convergence:DFedAvgM:wine}
\end{figure*}

\section{Methods}

In this section, we introduce the concept of \textit{client dropout} and present three strategies designed to mitigate its negative effects. A client dropout occurs when a client $i$ becomes permanently unavailable and can no longer participate in the joint optimization process. This implies that $i$ stops responding to peer-to-peer communications and is effectively excluded from further updates to the global objective~\cref{eq:global-objective}. Furthermore, the local data of the client $(X_i, Y_i)$ are lost, making continued optimization more difficult. In settings with iid data, such dropouts could be tolerable, as every client contributes similar information. However, in more realistic scenarios with class imbalance or non-iid data distributions, the absence of even a single client may significantly harm global performance. We provide empirical support for this observation in~\Cref{sec:experiments}.


\subsection{Baseline Strategy: No Reaction}
\label{sec:last-copy}

The first baseline comes down to taking no action when a client $i$ drops out. Optimization continues under the assumption that $i$ is still participating. In practice, this means that during any communication round involving $i$, no model exchange occurs, and the other clients retain and use the last known version of $\theta_i$. While simple, this approach often significantly hurts the performance because that model $\theta_i$ will never change and thus convergence (in either parameter or function space) is infeasible.  

\subsection{Baseline Strategy: Forget the Dropped Client}
\label{sec:forget-client}

The second baseline removes the dropped client completely from the federation. This is implemented in two steps: (1) client $i$ is disconnected from the communication graph $G$, and (2) all clients that previously interacted with $i$ delete their stored copy of $\theta_i$ and exclude it from future updates. For algorithms relying on neighbor information (e.g., DJAM, FSR), this also means that in their local objectives they will not consider the $i$-th neighbor in regularization. With this approach, the largest downside is that in non-iid settings, we simply lose all utility that could have otherwise been provided by the lost client. 

\subsection{Adaptive Strategies}
\label{sec:client-reconstr}

Unlike the baselines, our adaptive strategies attempt to \textit{reconstruct} the dropped client and restore its presence in the federation. Since the client's data is unavailable, we approximate it using synthetic data derived from the last known model. A new \textit{virtual client} is instantiated using the recovered data and continues participating in training as if it were the original.

\subsubsection{Random Images as a Sanity Check}

As a simple adaptive technique, we simulate the lost client using its last known model $\theta_i$ and completely random data. We generate synthetic inputs $X' \sim \mathcal{U}[0, 1]^d$ and assign labels $Y' \sim \mathcal{U}\{1, C\}$, where $C$ is the number of classes. The virtual client then continues local training on $(X', Y')$ as if it were real data. 

\subsubsection{Gradient Inversion}
\label{sec:grad-inversion}

Gradient inversion aims to reconstruct a synthetic dataset whose gradients closely match those of the lost client. We initialize $X' \sim \mathcal{U}[0,1]^d$ and assign either random or trainable labels $Y'$, then solve the following optimization:

\begin{equation}
    \mathcal{L}_{\text{GI}} = d(\nabla W' - \nabla W)^2 + \lambda \mathcal{L}_{\text{prior}}
\end{equation}

where $\nabla W' = \nabla_{\theta} \mathcal{L}_d(\theta, X', Y)$
and $\nabla W$ is the observed gradient from the lost client. The term $d(\cdot)$ denotes a distance metric (e.g., L2 norm \cite{zhu_deep_2019} or cosine similarity \cite{geiping_inverting_2020}), and $\mathcal{L}_{\text{prior}}$ includes domain-specific priors (e.g., total variation for image smoothness~\cite{yin_see_2021}).

\paragraph{Label Strategies.} While labels $Y'$ can be sampled randomly, performance improves when labels are optimized jointly with $X'$. One approach~\cite{wang_beyond_2019} includes an additional loss term:
\[
\mathcal{L}_{\text{labels}} = \| \theta(X') - Y' \|_2 \quad \text{where} \quad Y' = \text{softmax}(Y')
\]

In our experiments, we use both L2 and cosine variants and adopt joint label optimization, which has proven effective in prior work~\cite{geiping_inverting_2020, yin_see_2021,wang_beyond_2019}.

\paragraph{Limitations.} Gradient inversion becomes difficult when gradients reflect updates from multiple local epochs, mini-batches, and data points, denoted by \textit{ENB} (Epochs, Number of samples, Batch size). In asynchronous DFL, \textit{ENB} is often large, making $\nabla W$ less informative. For this reason, we also explore model inversion techniques, which do not rely on access to the client's gradients, but rather assume a static model.

\subsubsection{Model Inversion}
\label{sec:model-inversion}

Model inversion assumes that the last available model $\theta_i$ is close to a stationary point for its local objective. Although we do not enforce strict optimality, we assume that:

\begin{equation}
\label{eq:grad-loss-approx-zero}
    \nabla_{\theta} \mathcal{L}_d(\theta, X, Y) \approx 0
\end{equation}

\begin{equation}
\label{eq:grad-loss-approx-zero}
    \mathcal{L}_{MI} = \nabla_{\theta} \mathcal{L}_d(\theta, X, Y)
\end{equation}

To recover training data, we generate a synthetic dataset $(X', Y')$ such that this approximate condition holds. The optimization problem becomes:

\begin{equation}
        \underset{X'}{\text{minimize}} \quad \mathcal{L}_d(\theta, X', Y') \quad
        \text{s.t.} \quad \nabla_{\theta} \mathcal{L}_d(\theta, X', Y') \leq \epsilon
\end{equation}

\noindent This can be solved by any first-order optimizer (e.g., SGD, Adam) performing iterative gradient descent, where the model $\theta$ is frozen and the gradients are propagated onto input data $X'$
\begin{equation}
    X'_{t+1} = X'_t - \eta \nabla_{X'_t} \mathcal{L}(\theta, X', Y')
\end{equation}
Here, $X'$ is initialized from a uniform or normal distribution, and $Y'$ from a uniform categorical distribution.

\paragraph{Domain Constraints.} For both gradient and model inversion attacks we apply $\mathcal{L}_{\text{prior}} = \sum_{i=1}^{d} \text{ReLU}(x_i - 1) + \text{ReLU}(-x_i)$, a prior loss to ensure inputs remain within a valid range - in our case $x \in [0,1]^d$. Additionally, we clamp $X'$ to the valid domain after each optimization step.

\paragraph{Remarks.} Compared to gradient inversion, model inversion may be less sensitive to large \textit{ENB}, as it does not rely on access to $\nabla W$. Instead, it exploits the structure of $\theta_i$ itself. We empirically assess the viability of this approach across different data distributions and DFL algorithms.

%% file: chapters/5.experiments.tex
\begin{table*}[htb]
    \centering
    \input{figures/fedKDD-tabular2-plots/tables/DFedAvgM-table}
    \caption{Mean ($\pm$ std) accuracy of clients on a test set after 200 communication rounds for DFedAvgM algorithm over 10 folds. }
    \label{tab:results:dfedavgm}
\end{table*}

\section{Experiments}
\label{sec:experiments}

We evaluate the effectiveness of our dropout mitigation strategies across different DFL algorithms and data distributions. All experiments are conducted using logistic regression models implemented in PyTorch. We use a fully connected communication graph $G$, with $g_{ij} = 1$ for all $i \neq j$, and $g_{ii} = 0$ to avoid self-regularization. The federation consists of three clients, and we set the number of peer exchanges per communication round (parameter $k$ in \Cref{alg:distributed-optimization}) to 2. Client dropout occurs persistently at the 5th communication round. From that point onward, the dropped client ceases all participation, and its data becomes inaccessible.

Each experiment uses one of three datasets from the UCI repository\footnote{\href{https://archive.ics.uci.edu}{https://archive.ics.uci.edu}}: \textit{Wine}, \textit{Iris}, and \textit{Digits}. We apply 10-fold cross-validation and limit training to 200 communication rounds. Early stopping is triggered if all clients converge to the same accuracy on a holdout test set for 10 consecutive rounds. Full details on hyperparameters and implementation can be found in \Cref{app:more-details}.

We study three different data partitioning schemes: (1) \textbf{iid}, where data is uniformly distributed among clients; (2) \textbf{non-iid (clusters)}, where data is partitioned using $k$-means clustering; and (3) \textbf{non-iid (classes)}, where each client receives data from a distinct subset of classes.

\subsection{Empirical Results}

We first present convergence results (e.g., \Cref{fig:convergence:DFedAvgM:wine}, full set in \Cref{app:exp:all-conv-plots}) that show mean test accuracy across all clients over time. Each curve represents a different dropout-handling strategy: the \textcolor{reference}{\textbf{Reference}} strategy (green) represents the ideal scenario without client dropout; the \textcolor{drop}{\textbf{Drop}} (forget) strategy (red) permanently removes the lost client; the \textcolor{noaction}{\textbf{No Action}} strategy (gray) retains a frozen copy of the dropped client’s model; the \textcolor{random}{\textbf{Random}} strategy (black) reinstates the client with random data; the \textcolor{gradientinversion}{\textbf{Gradient Inversion}} strategy (orange) reconstructs the client using matched gradients; and the \textcolor{modelinversion}{\textbf{Model Inversion}} strategy (blue) attempts reconstruction based solely on the final model weights. 

Below, we first analyze the results for each method separately, and then provide a general summary of findings.

\paragraph{DFedAvgM} For all datasets and data partitions, model inversion consistently achieves the best performance, often matched closely by gradient inversion. The random strategy ranks third, with a negligible performance gap in the iid setting. In contrast, the drop strategy performs well only under iid assumptions, where clients still have reasonably representative data. In non-iid scenarios, removing the client significantly reduces global performance.

\paragraph{FSR} The trend is similar: model inversion typically performs best, with gradient inversion closely following. The random strategy again performs better than baseline strategies but falls short of the inversion-based methods.

\paragraph{DJAM} While model and gradient inversion still outperform the baselines, the performance gap relative to the random strategy is less pronounced. We hypothesize that this is due to the unobserved neighbor regularization loss in DJAM. Since we do not have access to this loss component for the dropped client, the data reconstruction becomes less precise, reducing the effectiveness of our adaptive attacks. This is supported by qualitative analysis of the recovered images from the Digits dataset (\Cref{app:extracted-data}), where samples extracted for DJAM resemble random noise, unlike the more coherent samples seen with DFedAvgM.

\paragraph{Summary} On average, adaptive strategies based on data reconstruction  outperform both baselines and the random strategy. The performance gain is most pronounced in non-iid scenarios, highlighting the importance of recovering client-specific information in heterogeneous federations.

\subsection{Model Similarity}
\label{sec:model-similarity}

\begin{figure}[tb]
    \centering
    \includegraphics[width=0.9\linewidth]{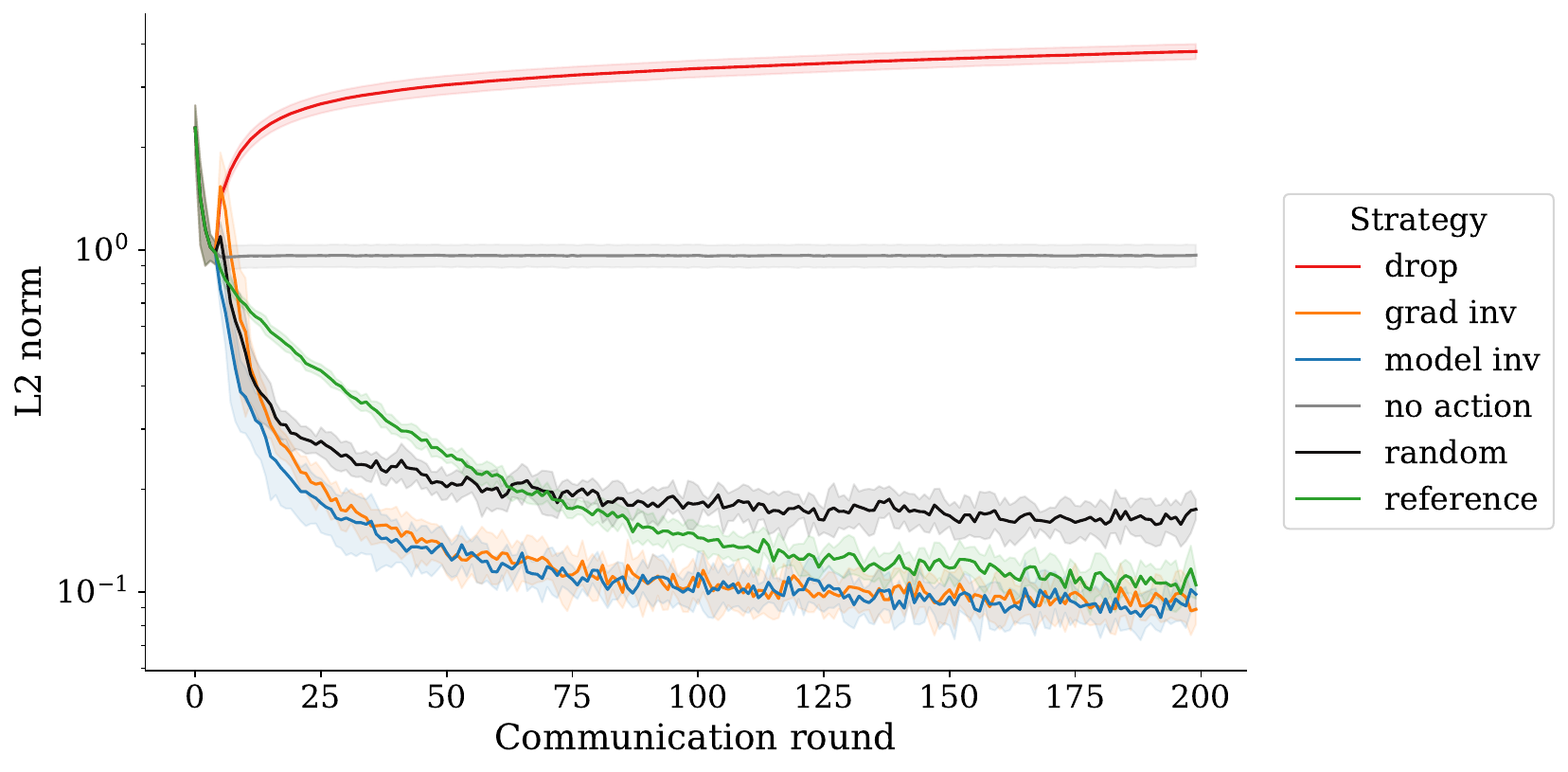}
    \caption{Model similarity over time (L2 norm between model parameters) for DJAM on the Digits dataset with non-iid (class) partitioning. Lower values indicate higher similarity. Note, that the y-axis is in logarithmic scale.}
    \label{fig:similarity:DJAM:digits}
\end{figure}

To further investigate system dynamics post-dropout, we analyze how model similarity evolves across clients during training. Since the goal of DFL is to reach consensus, we expect inter-client model distances (in parameter space) to decrease over time.

In \Cref{fig:similarity:DJAM:digits}, we plot the L2 norm between client models across communication rounds for DJAM on the Digits dataset. The \textcolor{reference}{reference} strategy shows a steady decrease in distance, indicating convergence. In the \textcolor{drop}{drop} scenario, similarity drops off as the remaining clients exclude the lost one, leading to model divergence. In the \textcolor{noaction}{no action} case, similarity may stagnate or fluctuate, depending on how aligned the stale model is with the evolving federation. 

For all adaptive strategies, we observe a consistent reduction in distance, as the reconstructed (virtual) client re-engages in learning, pulling models toward a shared solution. These trends reinforce the conclusion that our adaptive strategies enable more cohesive behavior of the federation under communication loss or other client failures.

%% file: figures/fedKDD-tabular2-plots/tables/DFedAvgM-table.tex
\begin{tabular}{ll|ccccc|c}
 \textbf{Dataset}   & \textbf{Distribution}       & \textbf{No action}       & \textbf{Forget}            & \textbf{Random}          & \textbf{Grad inv}        & \textbf{Model inv}       & \textbf{Reference}       \\
\hline
 \multirow{3}{*}{wine}      & iid                & $0.96 \pm 0.03$ & $0.96 \pm 0.03$ & $0.90 \pm 0.06$ & $0.97 \pm 0.03$ & $0.97 \pm 0.03$ & $0.97 \pm 0.03$ \\
       & non-iid (clusters) & $0.62 \pm 0.10$ & $0.62 \pm 0.10$ & $0.64 \pm 0.11$ & $0.78 \pm 0.14$ & \textbf{0.86 $\pm$ 0.10 }& $0.99 \pm 0.02$ \\
       & non-iid (class)    & $0.55 \pm 0.01$ & $0.55 \pm 0.01$ & $0.63 \pm 0.07$ & $0.71 \pm 0.08$ & \textbf{0.82 $\pm$ 0.05} & $0.97 \pm 0.03$ \\ \hline
 \multirow{3}{*}{iris}       & iid               & $0.90 \pm 0.04$ & $0.90 \pm 0.04$ & $0.89 \pm 0.09$ & $0.92 \pm 0.09$ & \textbf{0.95 $\pm $0.04} & $0.97 \pm 0.04$ \\
       & non-iid (clusters) & $0.64 \pm 0.11$ & $0.64 \pm 0.11$ & $0.70 \pm 0.17$ & $0.79 \pm 0.17$ & \textbf{0.87 $\pm$ 0.12 }& $0.94 \pm 0.05$ \\
       & non-iid (class)    & $0.57 \pm 0.04$ & $0.57 \pm 0.04$ & $0.57 \pm 0.13$ & $0.62 \pm 0.10$ & \textbf{0.73 $\pm$ 0.08} & $0.84 \pm 0.04$ \\ \hline
 \multirow{3}{*}{digits}     & iid                & $0.94 \pm 0.01$ & $0.94 \pm 0.01$ & $0.94 \pm 0.01$ & $0.95 \pm 0.02$ & $0.94 \pm 0.02$ & $0.95 \pm 0.01$ \\
     & non-iid (clusters) & $0.75 \pm 0.04$ & $0.75 \pm 0.04$ & $0.76 \pm 0.04$ & $0.84 \pm 0.06$ & \textbf{0.86 $\pm$ 0.04} & $0.95 \pm 0.02$ \\
     & non-iid (class)    & $0.55 \pm 0.02$ & $0.55 \pm 0.02$ & $0.63 \pm 0.05$ & $0.69 \pm 0.04$ & \textbf{0.75 $\pm$ 0.04 }& $0.93 \pm 0.02$ \\
\end{tabular}

%% file: chapters/6.conclusions.tex
\section{Conclusions}
\label{sec:conclusions}

We introduced adaptive mitigation strategies for persistent client dropout in asynchronous decentralized federated learning, leveraging model and gradient inversion techniques to reconstruct lost clients. Our experiments demonstrate that these strategies consistently improve performance across various data heterogeneity conditions (especially in non-iid settings), optimization algorithms, and datasets, when compared to simpler alternatives.

In the future, we would like to expand this investigation to more complex scenarios, such as high-resolution image classification tasks, and conduct a more detailed analysis of the fidelity and privacy implications of the reconstructed data.

Although we conducted some preliminary experiments beyond the main results, this study does not yet systematically examine the impact of the size of the federation, network topology, or optimization hyperparameters. We provide select exploratory findings on these aspects in the Appendix, and we leave a more comprehensive investigation to future work.

%% file: chapters/7.0.appendix.tex
\section{Experiment Setup Details}
\label{app:more-details}

\subsection{Federation Parameters}

We run all experiments with three participating clients, and cap each experiment at 200 communication rounds to ensure a uniform communication budget and fair comparison across methods. In every round, two client pairs are randomly selected to exchange their latest models. Regardless of participation in exchanges, each client performs a randomly chosen number of local optimization steps, uniformly sampled from the following categorical interval: $\mathcal{U}\{5, 10\}$. The outline of the global optimization procedure is presented in \Cref{alg:distributed-optimization}.

\begin{algorithm}[h]
\caption{Asynchronous Decentralized Federated Learning framework}
\label{alg:distributed-optimization}
\begin{algorithmic}[1]
    \REQUIRE No of clients $m$, initial models $\{\theta_i\}_{i=1}^m$, comm. graph $G$
    \STATE Initialize each client $i$ with model $\theta_i$
    \WHILE{not converged}
        \FORALL{clients $i$ in parallel}
            \STATE Sample $E_i \sim \mathcal{U}\{5, 10\}$
            \FOR{$e = 1$ to $E_i$}
                \STATE $\theta_i \leftarrow \arg\min_{\theta_i} \mathcal{L}_i(\theta_i, X_i, Y_i)$
            \ENDFOR
        \ENDFOR
        \STATE Randomly select $k$ pairs $(i, j)$ such that $g_{ij} \neq 0$
        \FORALL{selected pairs $(i, j)$}
            \STATE Clients $i$ and $j$ exchange and update their models
        \ENDFOR
    \ENDWHILE
    \RETURN $\{\theta_i\}_{i=1}^m$
\end{algorithmic}
\end{algorithm}

\subsection{Handling Data}

Each data silo contains at most 200 samples. No datapoint is duplicated across clients. In the \textbf{iid} setting, data is randomly assigned to clients. In the \textbf{non-iid (clusters)} setting, we perform $k$-means clustering with $k$ equal to the number of clients and assign one cluster per client. In the \textbf{non-iid (classes)} setting, each client is assigned a unique subset of the available classes. If the number of classes exceeds the number of clients, classes are distributed as evenly as possible, but preserving the constraint of each client having a unique subset of classes.

Each client splits its local dataset into training and validation subsets using an 80-20 split. All experiments are performed using 10-fold cross-validation, where 9 folds are used for client training and 1 held-out fold is used as a global test set, which we use to calculate metrics reported in experiments.

\subsection{Algorithm Parameters}

All three algorithms (FSR, DJAM, DFedAvgM) use the SGD optimizer for local training. For DFedAvgM, we set the learning rate to $0.01$ and the momentum for self-regularization to $0.9$. For FSR and DJAM, the learning rate is set to $0.1$.

In FSR, we approximate the L2 distance in Hilbert space using random samples drawn from a uniform distribution. For each local data batch, we generate 500 synthetic points to estimate the function space regularization terms. While this approximation is admittedly suboptimal (particularly in high-dimensional input spaces) we consider it sufficient for the scope of this study. As FSR was only recently introduced by Good et al.~\cite{good2024trustworthy}, there is currently no established method in the literature for computing neighbor regularization coefficients more accurately. Additional hyperparameters include $\omega = 0.01$ and the neighbor regularization weight $\lambda = 50$.

\subsection{Reconstruction Parameters}

For all adaptive strategies, we reconstruct 50 synthetic data points per lost client. This number is arbitrarily fixed and does not assume any knowledge about the original client’s dataset size or local training steps between model updates.

In the \textit{random} baseline, the virtual client is trained for 10 epochs on randomly generated data before being introduced back to the federation. For both \textit{gradient inversion} and \textit{model inversion}, we use a batch size of 16, the Adam optimizer, and include auxiliary losses: Total Variation ( $\lambda_{TV}=0.01$) and Domain Constraint ($\lambda_{domain} = 0.1$).

For \textit{model inversion}, we set the learning rate to $0.01$, apply weight decay of $0.01$, and train for 1000 epochs. The generated data is class-balanced.

For \textit{gradient inversion}, the learning rate is set to $0.05$ with 2000 training epochs. We run both L2 and cosine similarity variants and report results based on the best-performing variant.

%% file: chapters/7.1.appendix.tex
\section{All convergence plots}
\label{app:exp:all-conv-plots}

\begin{figure*}[htbp]
    \centering
    \begin{subfigure}[b]{0.33\textwidth}
        \includegraphics[width=\linewidth]{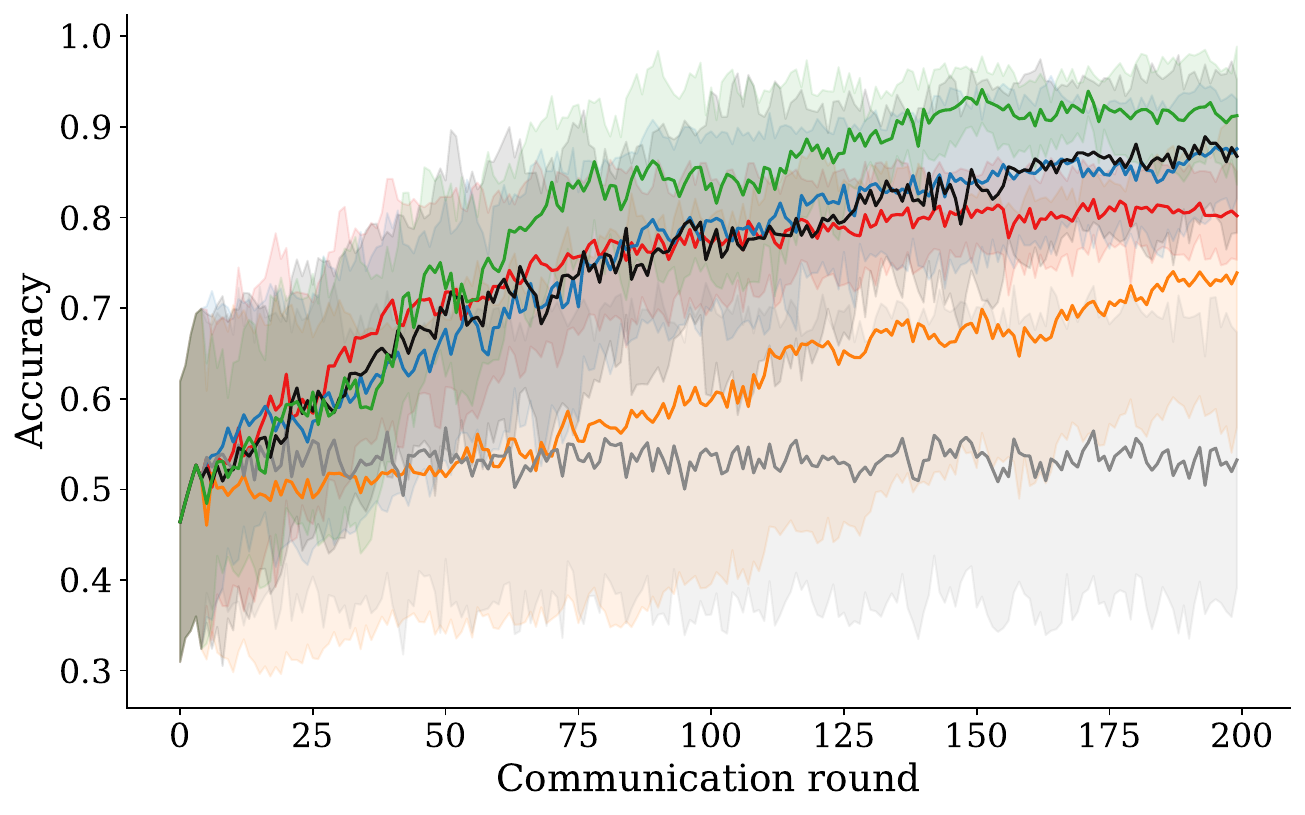}
        \caption{DJAM - iid}
       
    \end{subfigure}
    \hfill
    \begin{subfigure}[b]{0.33\textwidth}
        \includegraphics[width=\linewidth]{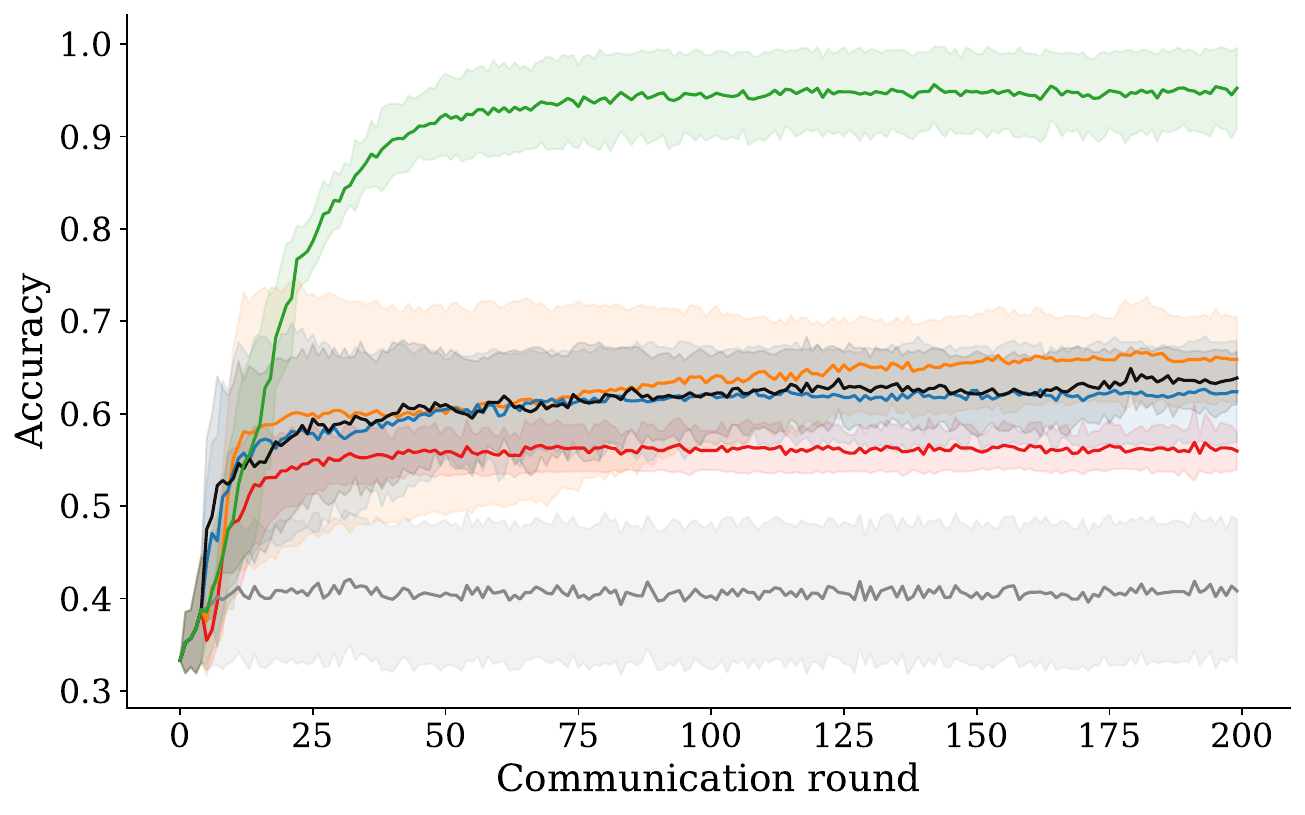}
        \caption{DJAM - non-iid (clusters)}
     
    \end{subfigure}
    \hfill
    \begin{subfigure}[b]{0.33\textwidth}
        \includegraphics[width=\linewidth]{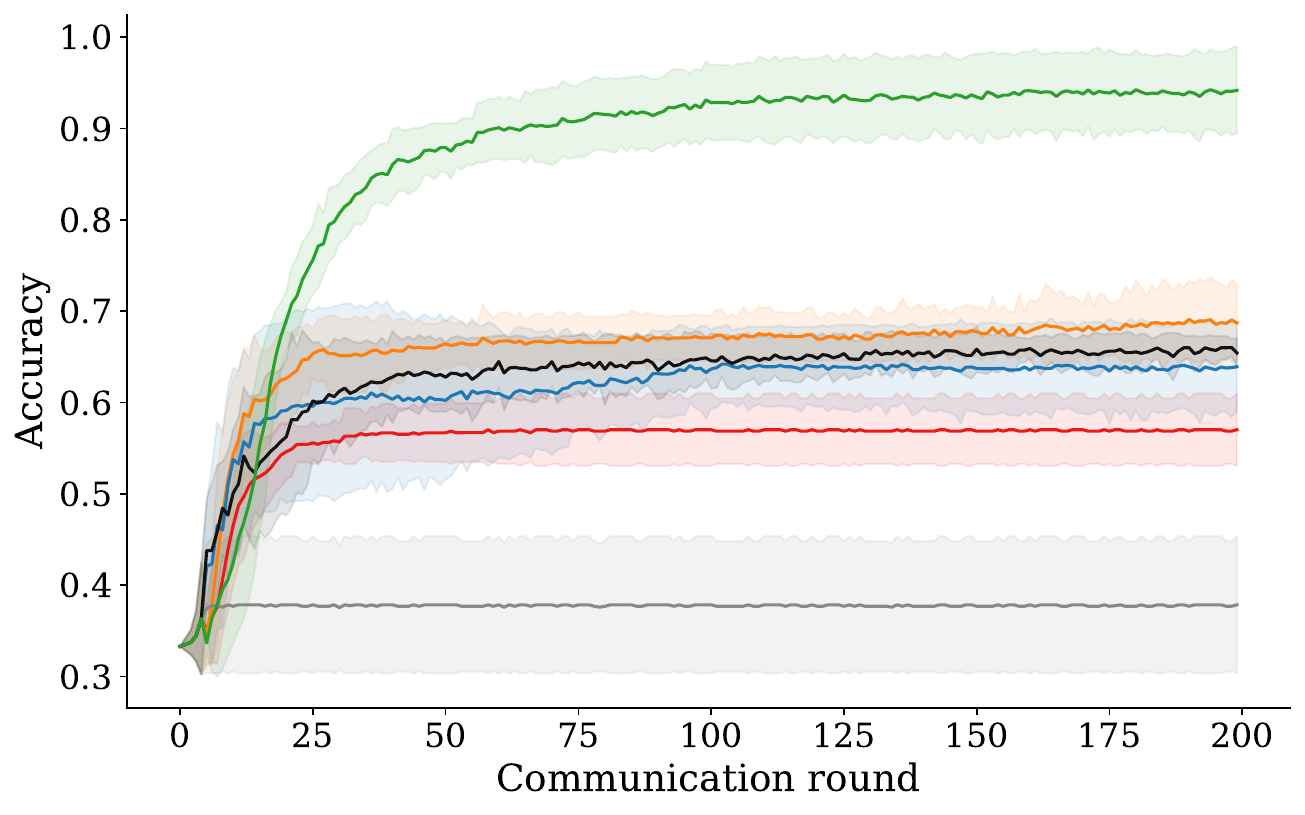}
        \caption{DJAM - non-iid (classes)}
       
    \end{subfigure}

    \begin{subfigure}[b]{0.33\textwidth}
        \includegraphics[width=\linewidth]{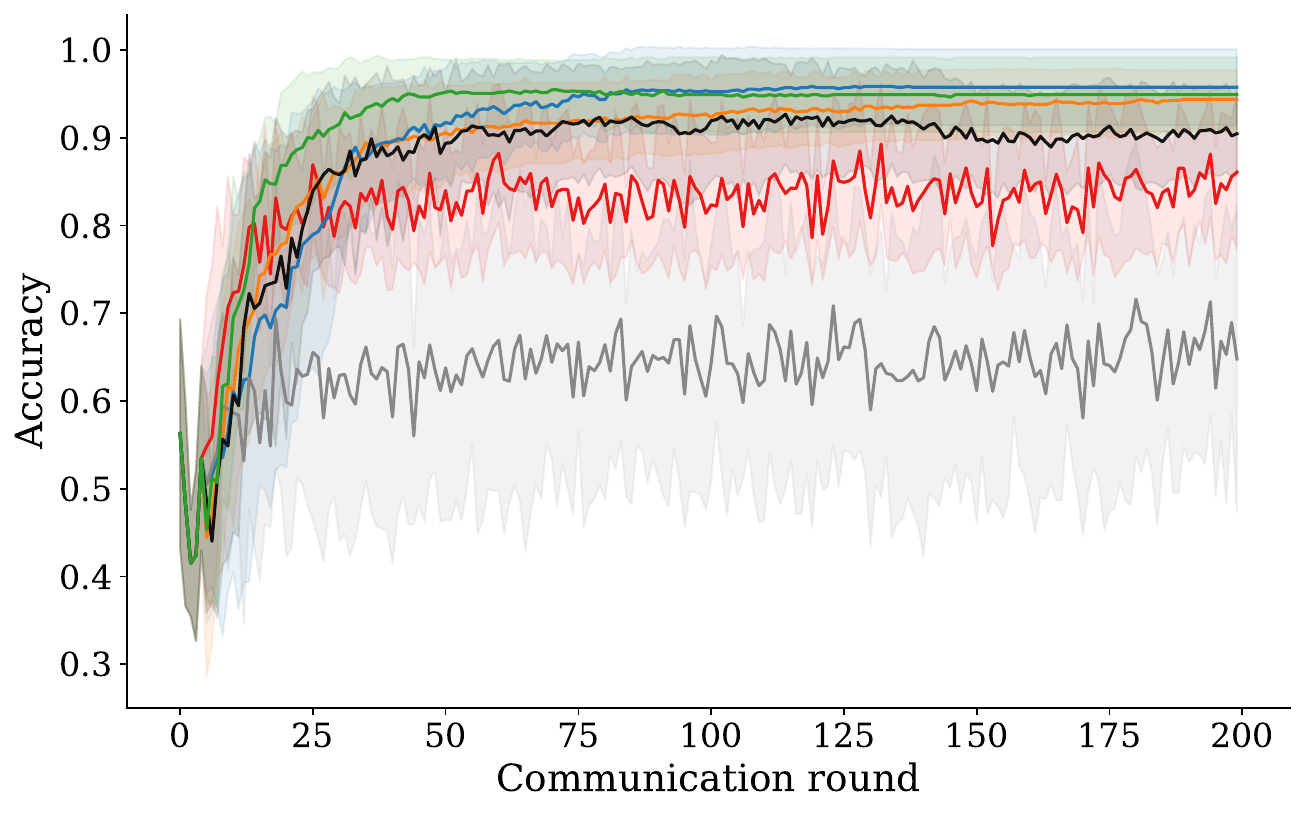}
        \caption{FSR - iid}
       
    \end{subfigure}
    \hfill
    \begin{subfigure}[b]{0.33\textwidth}
        \includegraphics[width=\linewidth]{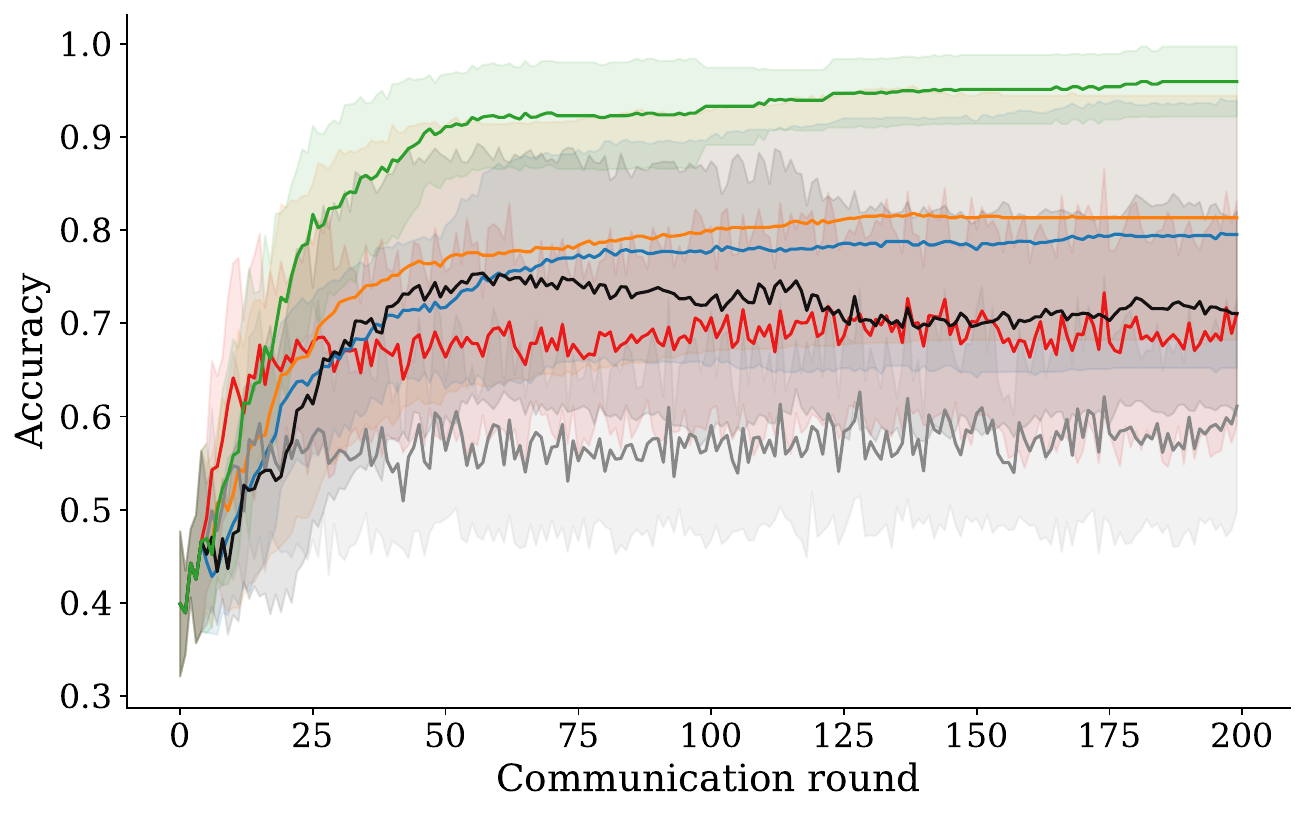}
        \caption{FSR - non-iid (clusters)}
     
    \end{subfigure}
    \hfill
    \begin{subfigure}[b]{0.33\textwidth}
        \includegraphics[width=\linewidth]{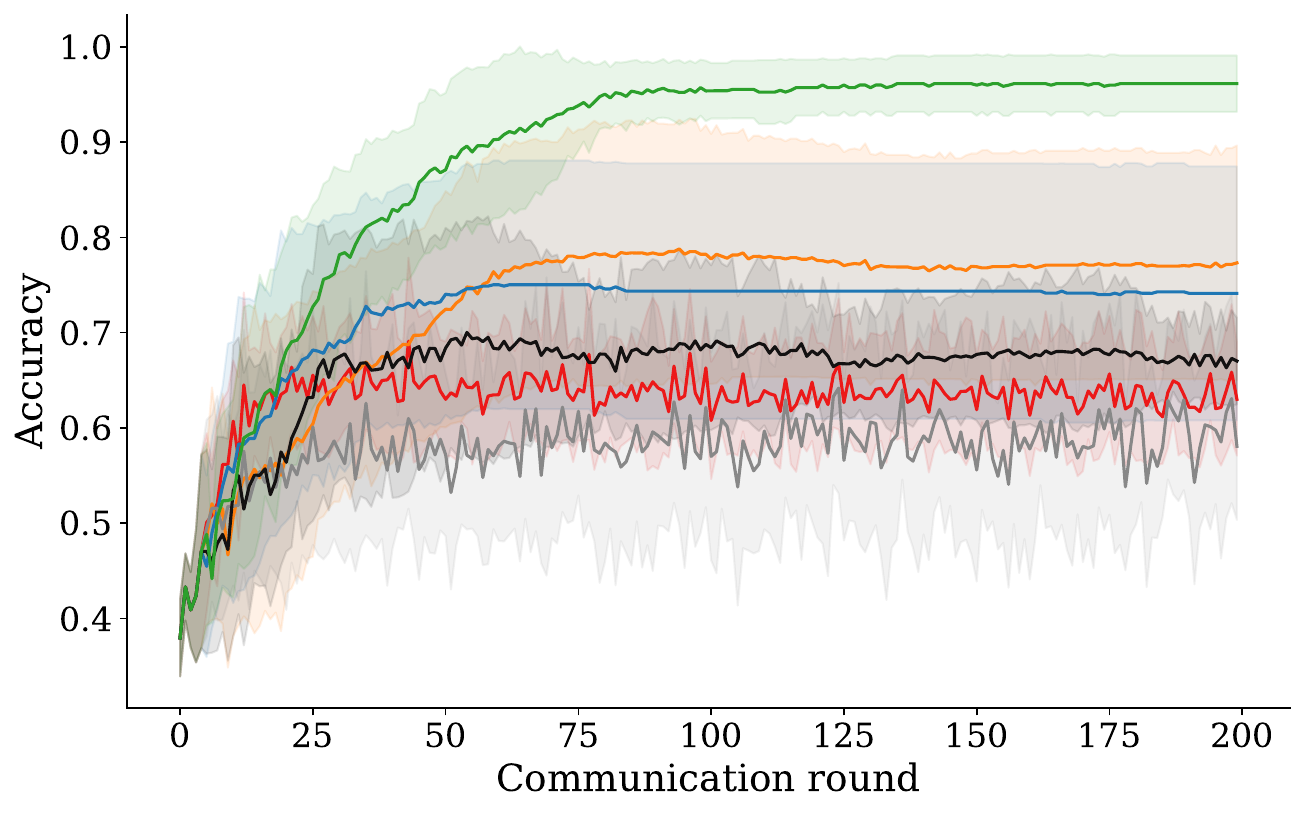}
        \caption{FSR - non-iid (classes)}
        
    \end{subfigure}

    \begin{subfigure}[b]{0.33\textwidth}
        \includegraphics[width=\linewidth]{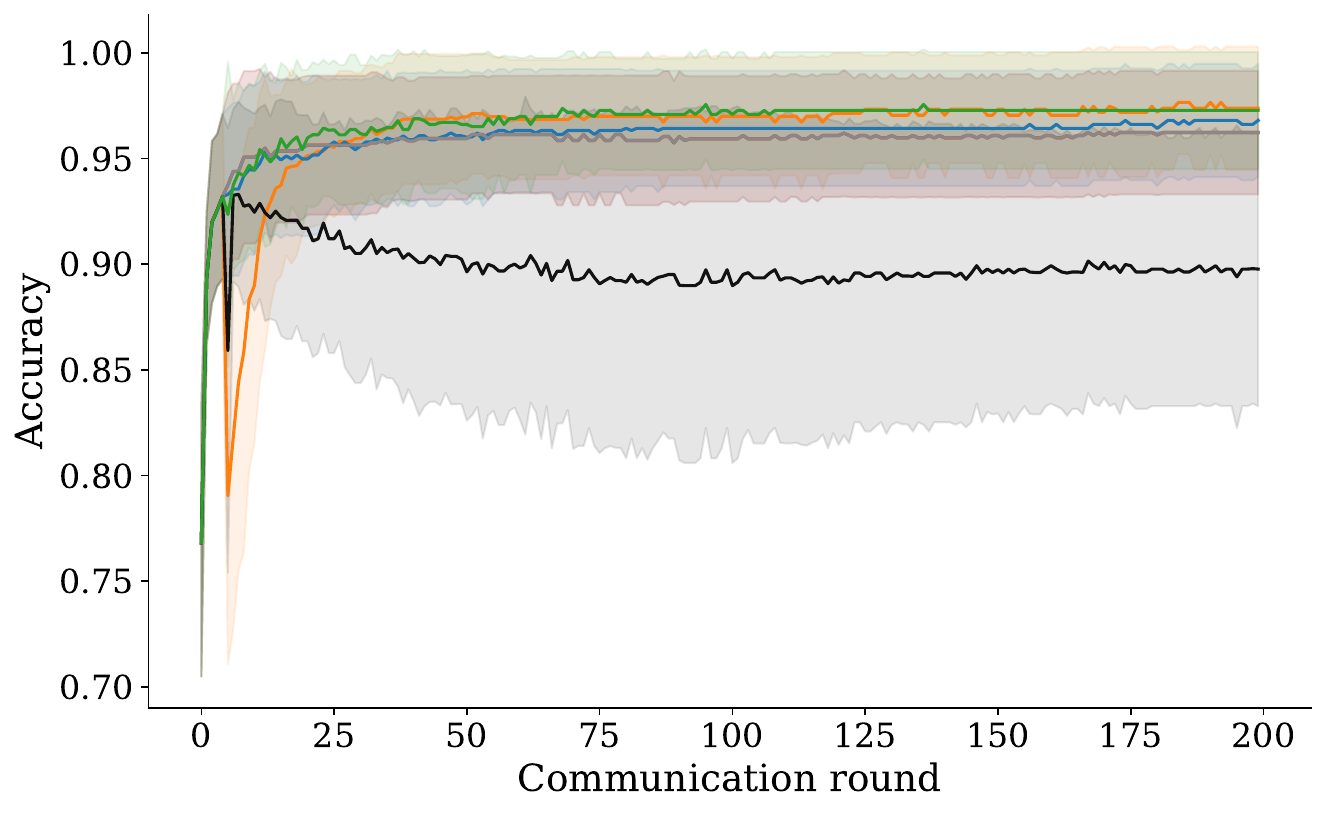}
        \caption{DFedAvgM - iid}
       
    \end{subfigure}
    \hfill
    \begin{subfigure}[b]{0.33\textwidth}
        \includegraphics[width=\linewidth]{figures/fedKDD-tabular2-plots/convergence/wine_DFedAvgM_clusters.pdf}
        \caption{DFedAvgM - non-iid (clusters)}
        
    \end{subfigure}
    \hfill
    \begin{subfigure}[b]{0.33\textwidth}
        \includegraphics[width=\linewidth]{figures/fedKDD-tabular2-plots/convergence/wine_DFedAvgM_extreme_imbalance.pdf}
        \caption{DFedAvgM - non-iid (classes)}
     
    \end{subfigure}

    \caption{Convergence plots for DJAM, FSR, and DFedAvgM algorithms on the wine dataset. Shaded regions represent mean $\pm$ standard deviation across 10 folds, over 200 communication rounds. A random client is dropped persistently after the 5th round. Colors: {\color{modelinversion}model inversion}, {\color{gradientinversion}gradient inversion}, {\color{reference}reference}, {\color{random}random}, {\color{drop}drop}, {\color{noaction}no action}}.
    \label{fig:convergence:wine}
\end{figure*}

\begin{figure*}[htbp]
    \centering
    \begin{subfigure}[b]{0.33\textwidth}
        \includegraphics[width=\linewidth]{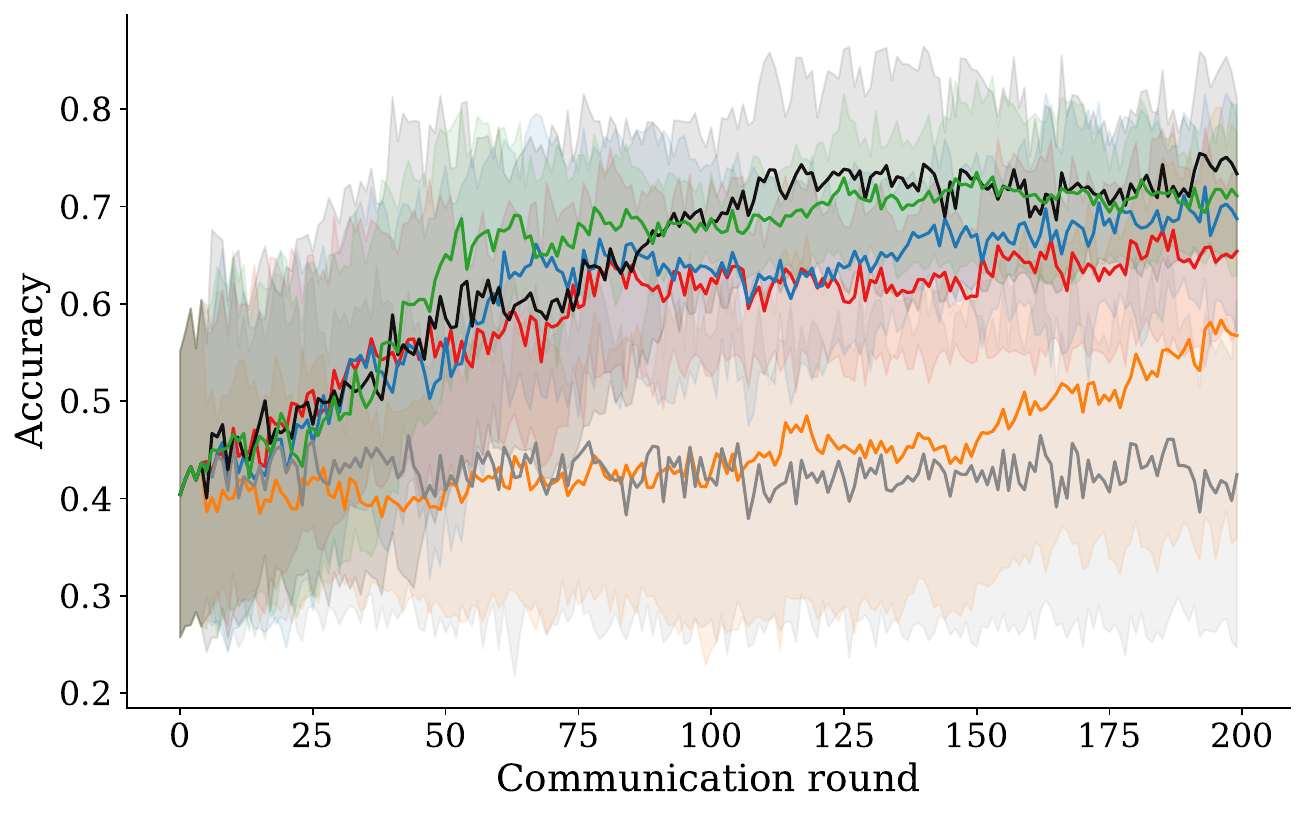}
        \caption{DJAM - iid}
     
    \end{subfigure}
    \hfill
    \begin{subfigure}[b]{0.33\textwidth}
        \includegraphics[width=\linewidth]{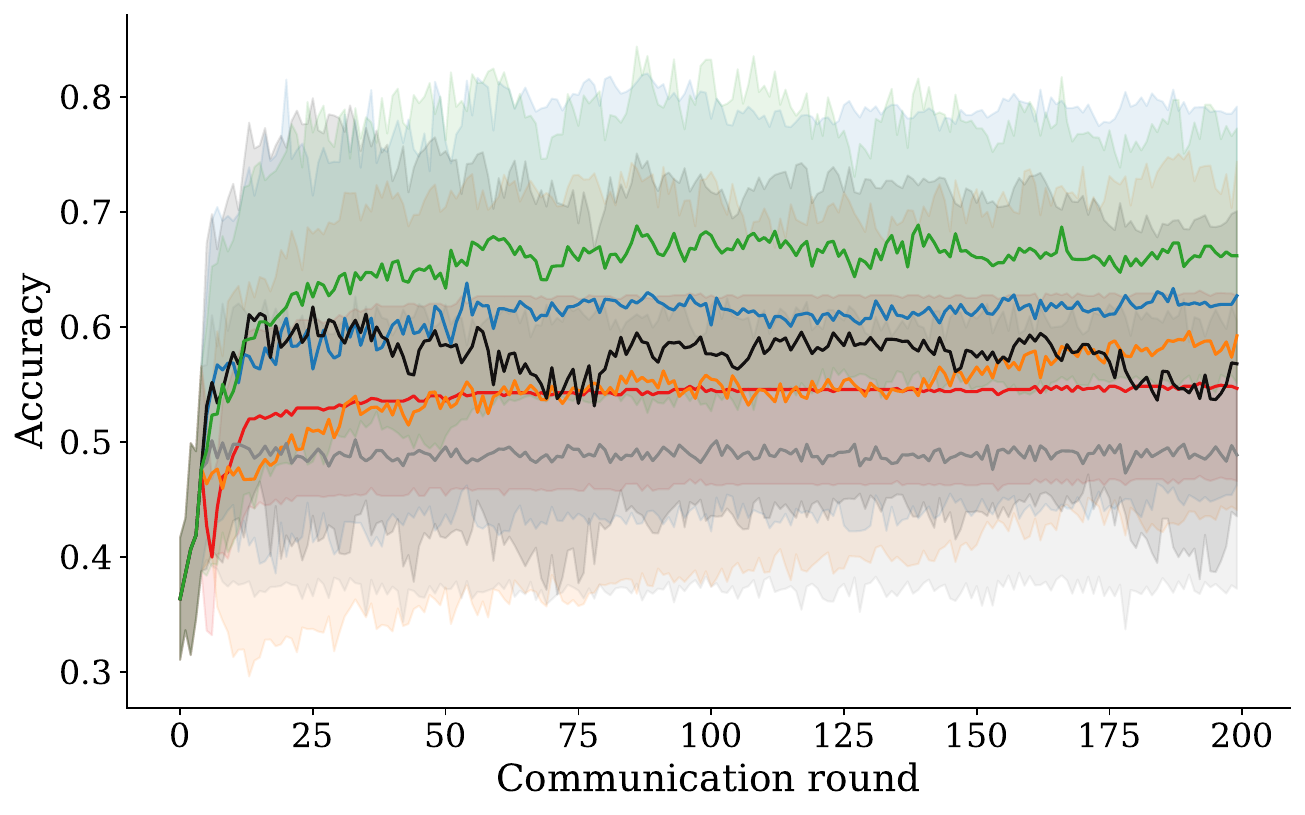}
        \caption{DJAM - non-iid (clusters)}
       
    \end{subfigure}
    \hfill
    \begin{subfigure}[b]{0.33\textwidth}
        \includegraphics[width=\linewidth]{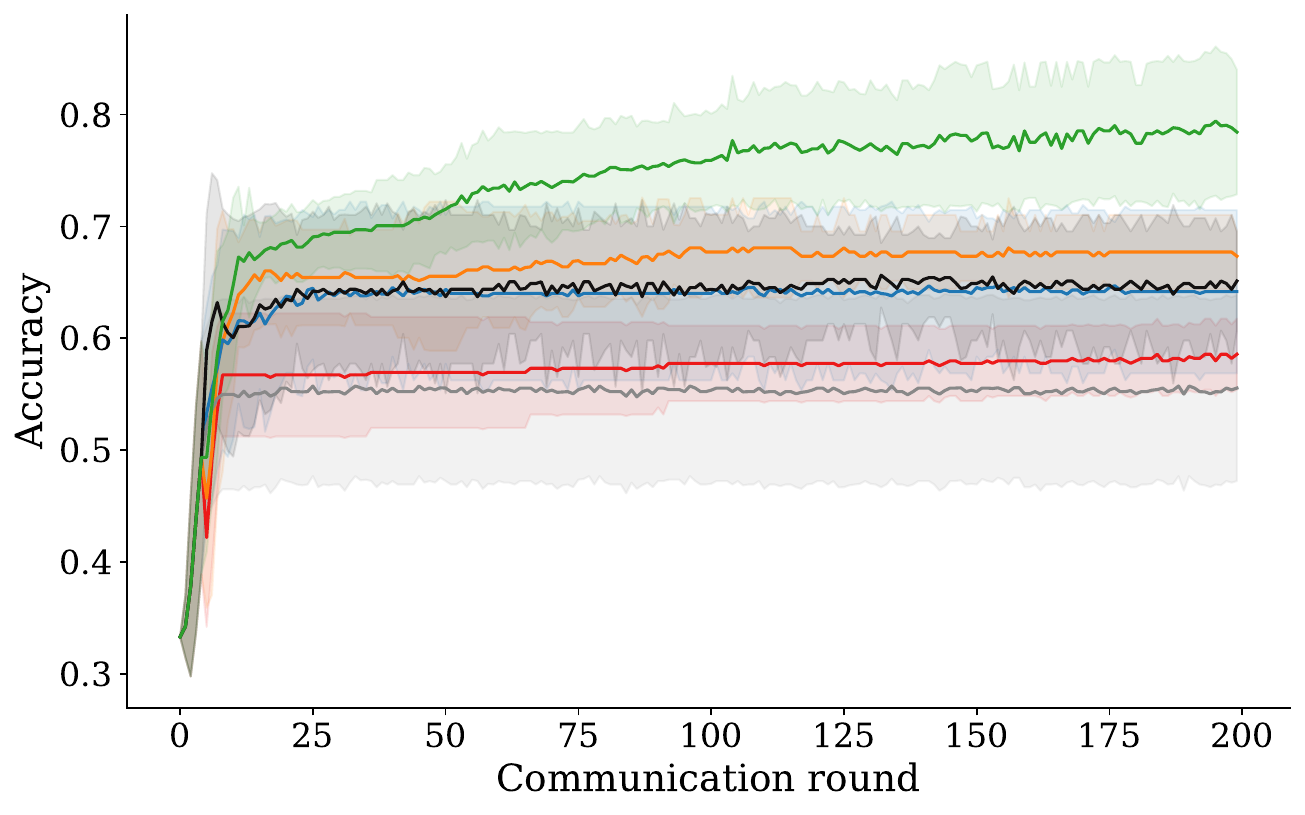}
        \caption{DJAM - non-iid (classes)}
       
    \end{subfigure}

    \begin{subfigure}[b]{0.33\textwidth}
        \includegraphics[width=\linewidth]{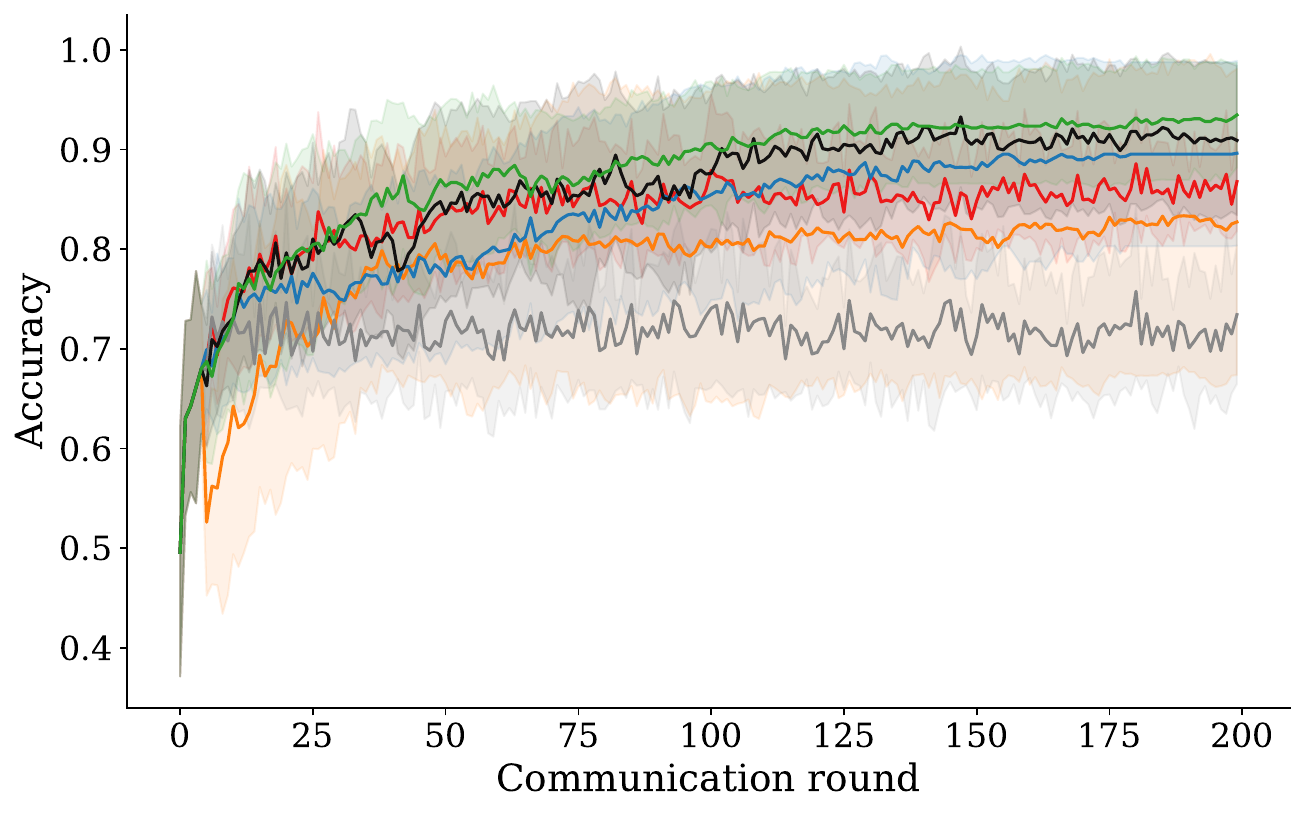}
        \caption{FSR - iid}
       
    \end{subfigure}
    \hfill
    \begin{subfigure}[b]{0.33\textwidth}
        \includegraphics[width=\linewidth]{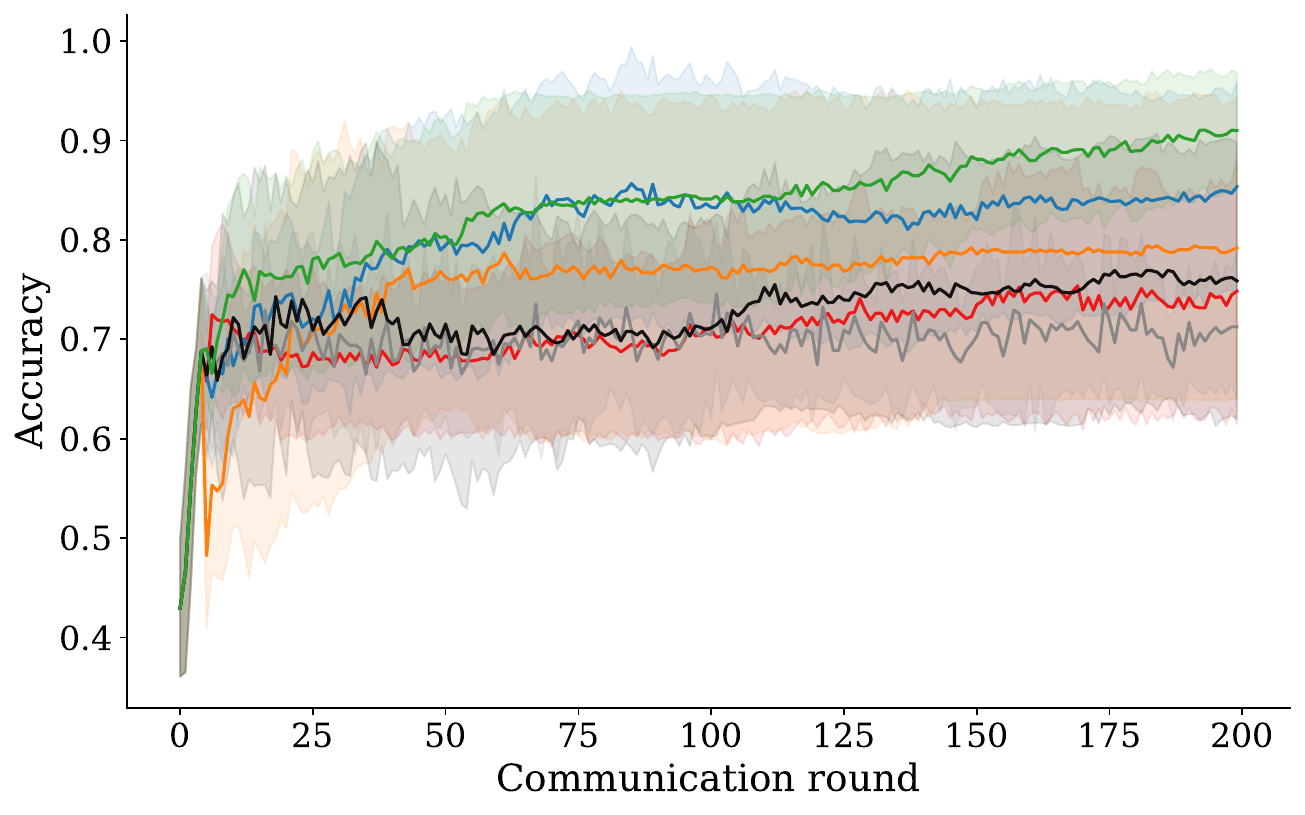}
        \caption{FSR - non-iid (clusters)}
      
    \end{subfigure}
    \hfill
    \begin{subfigure}[b]{0.33\textwidth}
        \includegraphics[width=\linewidth]{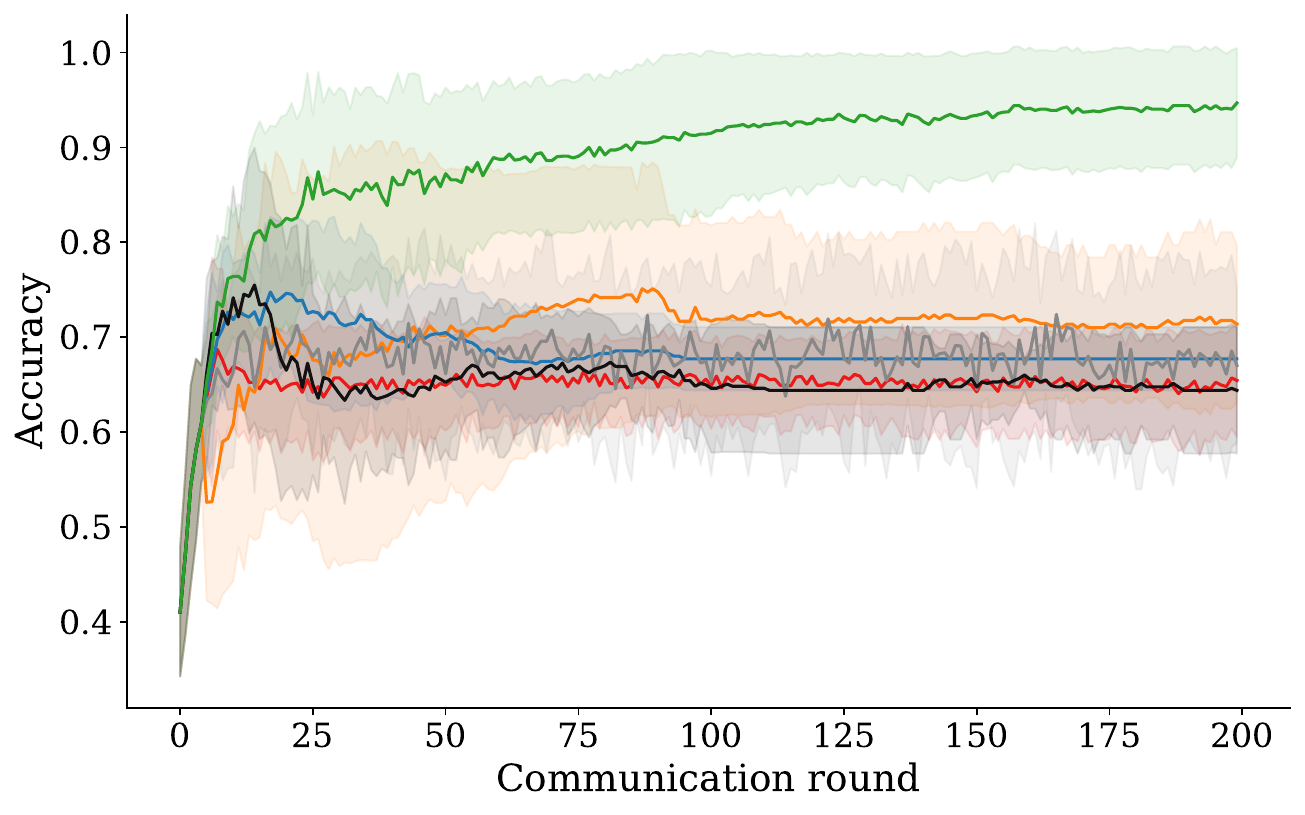}
        \caption{FSR - non-iid (classes)}
       
    \end{subfigure}

    \begin{subfigure}[b]{0.33\textwidth}
        \includegraphics[width=\linewidth]{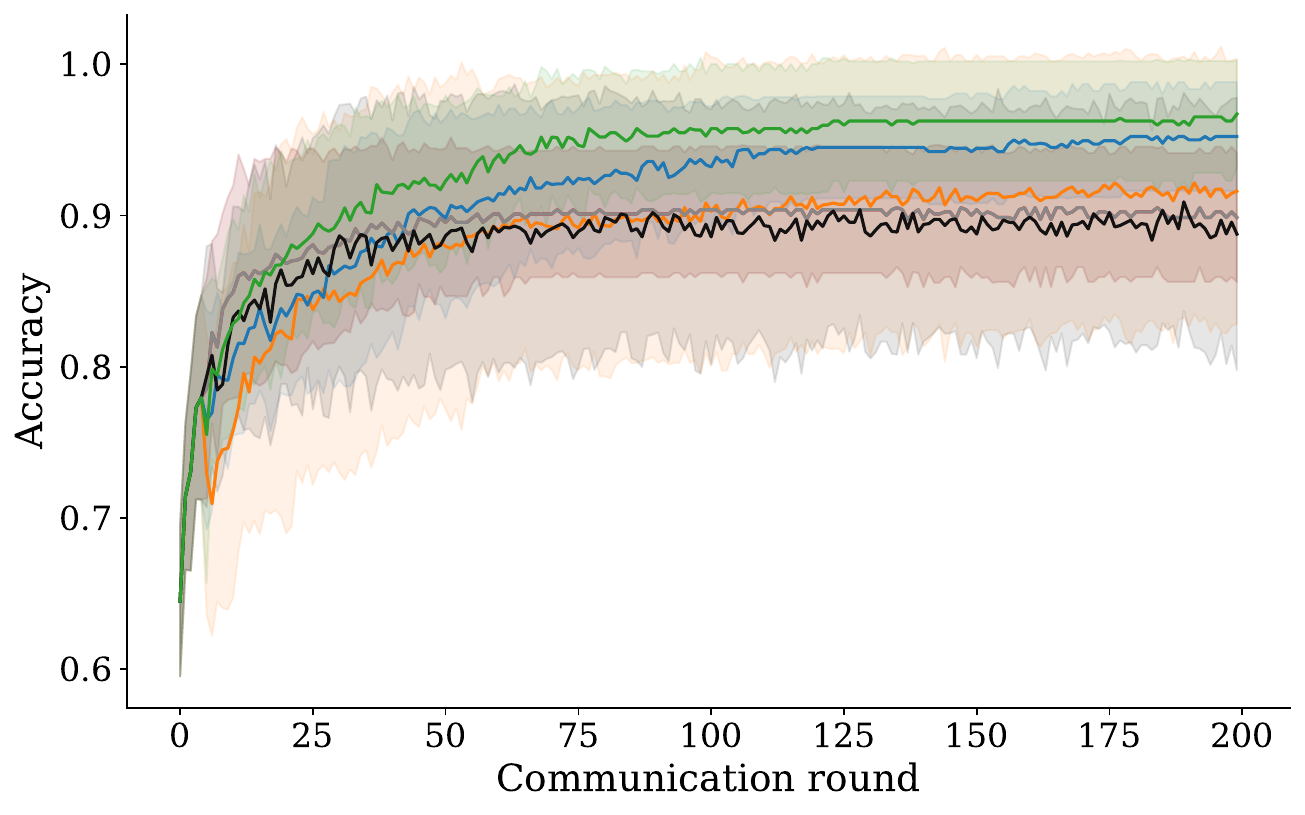}
        \caption{DFedAvgM - iid}
     
    \end{subfigure}
    \hfill
    \begin{subfigure}[b]{0.33\textwidth}
        \includegraphics[width=\linewidth]{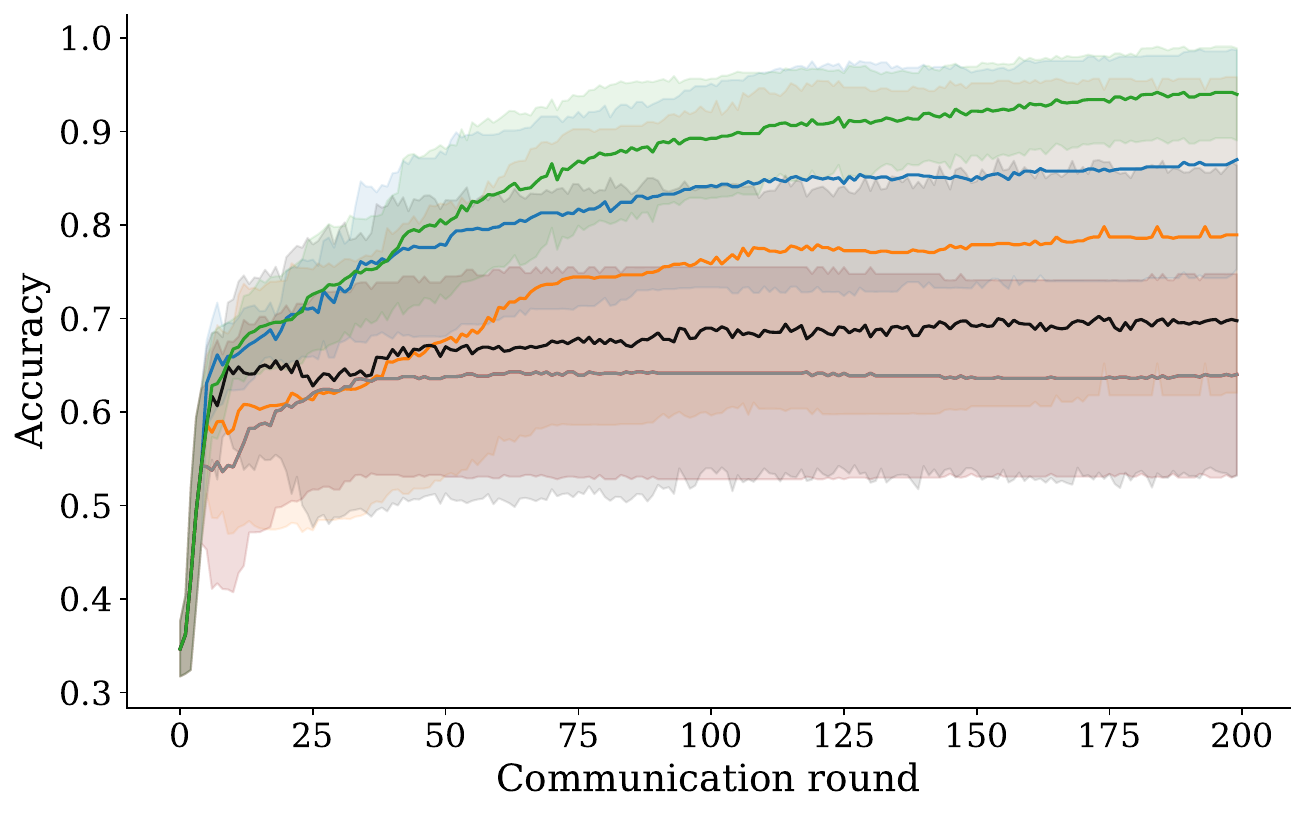}
        \caption{DFedAvgM - non-iid (clusters)}
       
    \end{subfigure}
    \hfill
    \begin{subfigure}[b]{0.33\textwidth}
        \includegraphics[width=\linewidth]{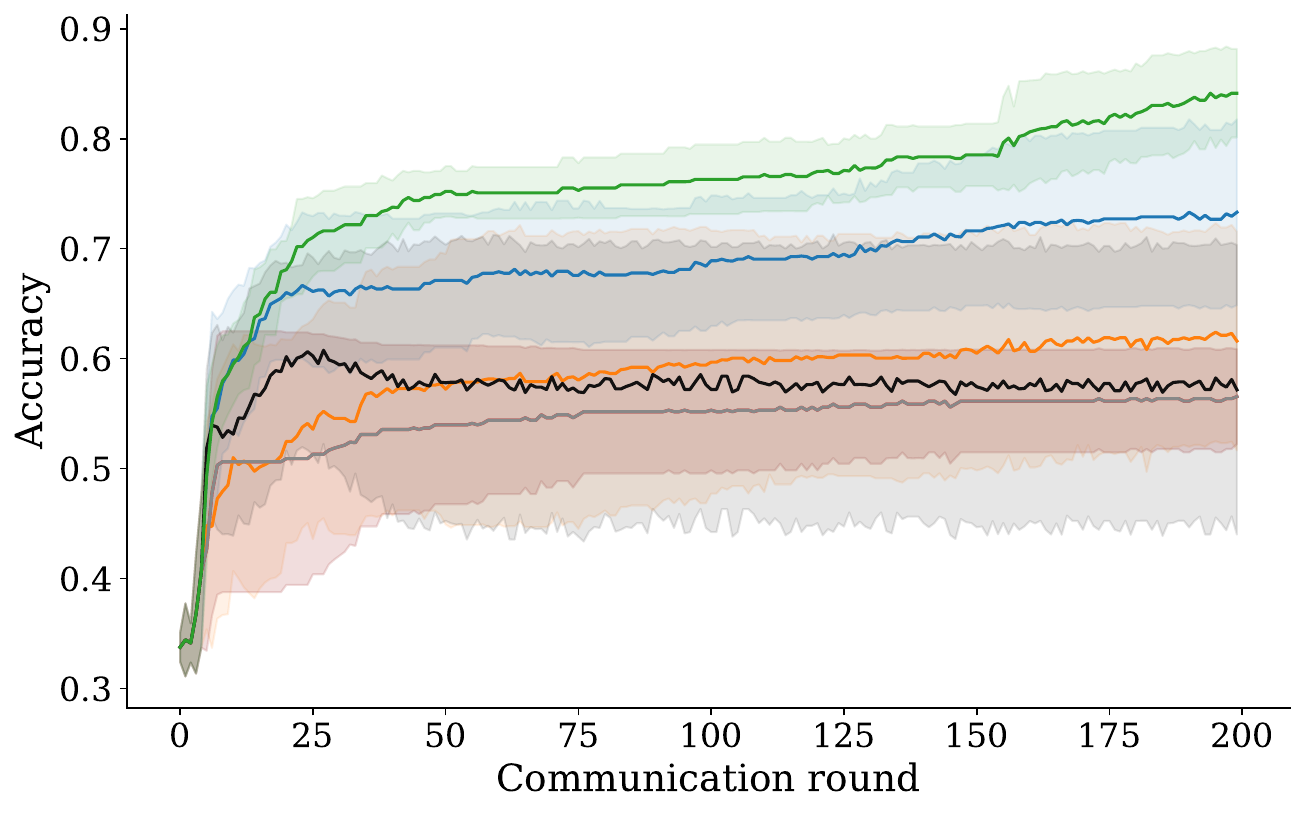}
        \caption{DFedAvgM - non-iid (classes)}
      
    \end{subfigure}

    \caption{Convergence plots for DJAM, FSR, and DFedAvgM algorithms on the iris dataset. Shaded regions represent mean $\pm$ standard deviation across 10 folds, over 200 communication rounds. A random client is dropped persistently after the 5th round. Colors: {\color{modelinversion}model inversion}, {\color{gradientinversion}gradient inversion}, {\color{reference}reference}, {\color{random}random}, {\color{drop}drop}, {\color{noaction}no action}}.
    \label{fig:convergence:iris}
\end{figure*}

\begin{figure*}[htbp]
    \centering
    \begin{subfigure}[b]{0.33\textwidth}
        \includegraphics[width=\linewidth]{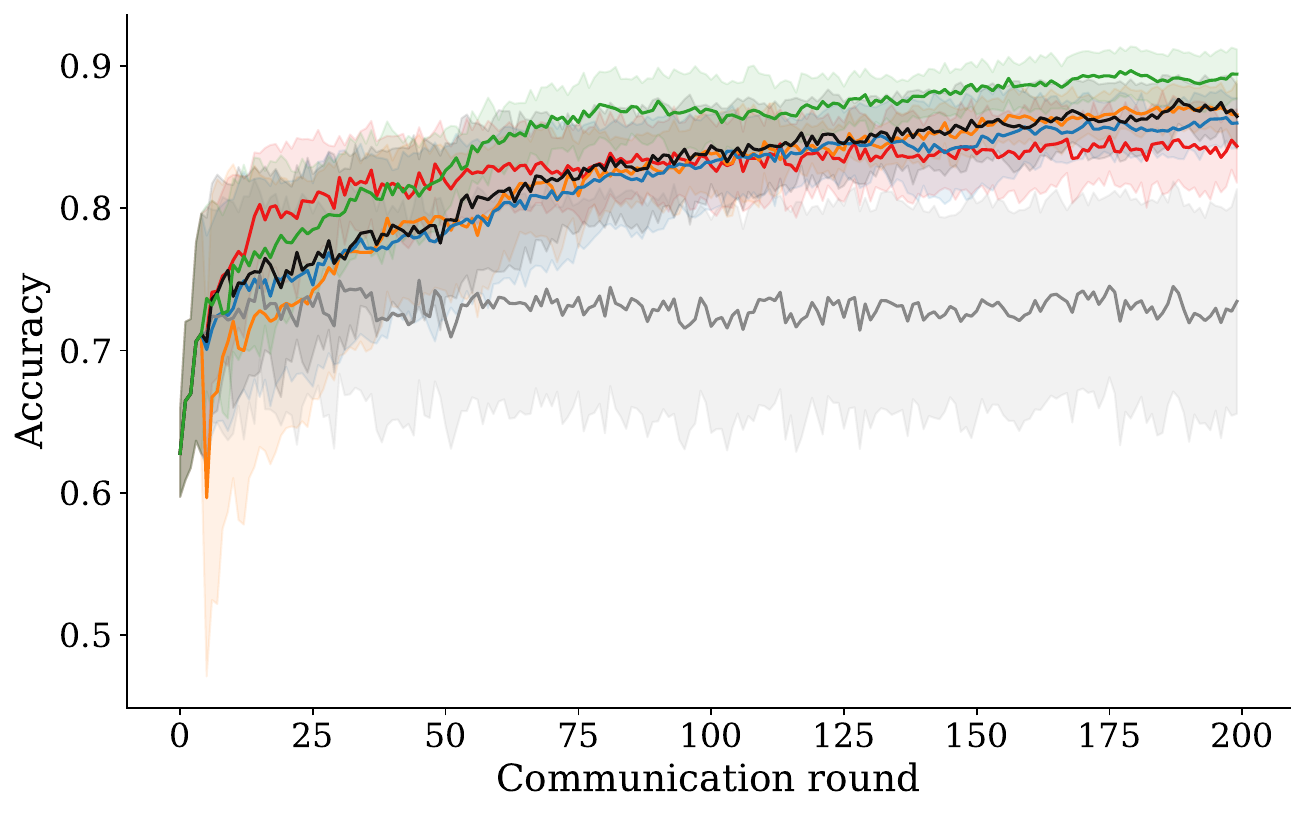}
        \caption{DJAM - iid}
      
    \end{subfigure}
    \hfill
    \begin{subfigure}[b]{0.33\textwidth}
        \includegraphics[width=\linewidth]{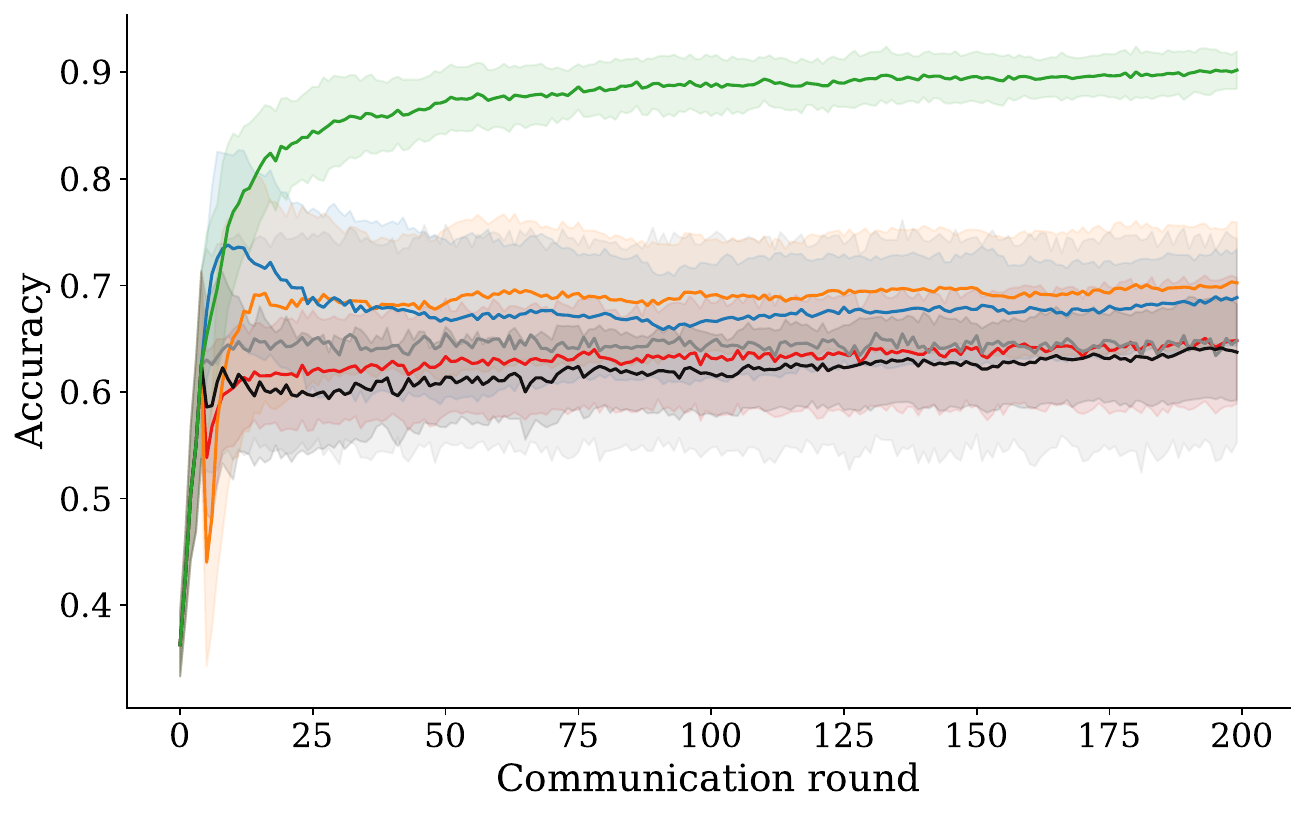}
        \caption{DJAM - non-iid (clusters)}
  
    \end{subfigure}
    \hfill
    \begin{subfigure}[b]{0.33\textwidth}
        \includegraphics[width=\linewidth]{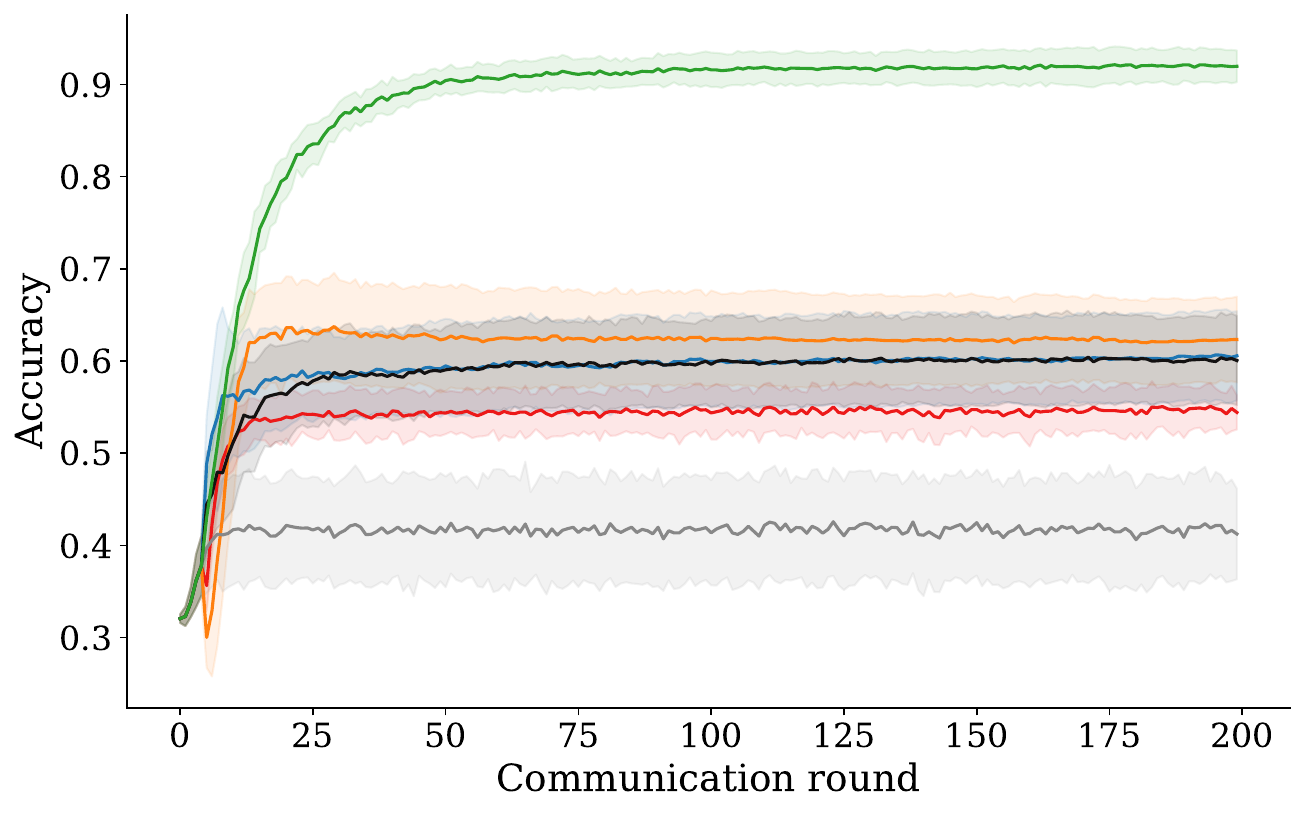}
        \caption{DJAM - non-iid (classes)}
      
    \end{subfigure}

    \begin{subfigure}[b]{0.33\textwidth}
        \includegraphics[width=\linewidth]{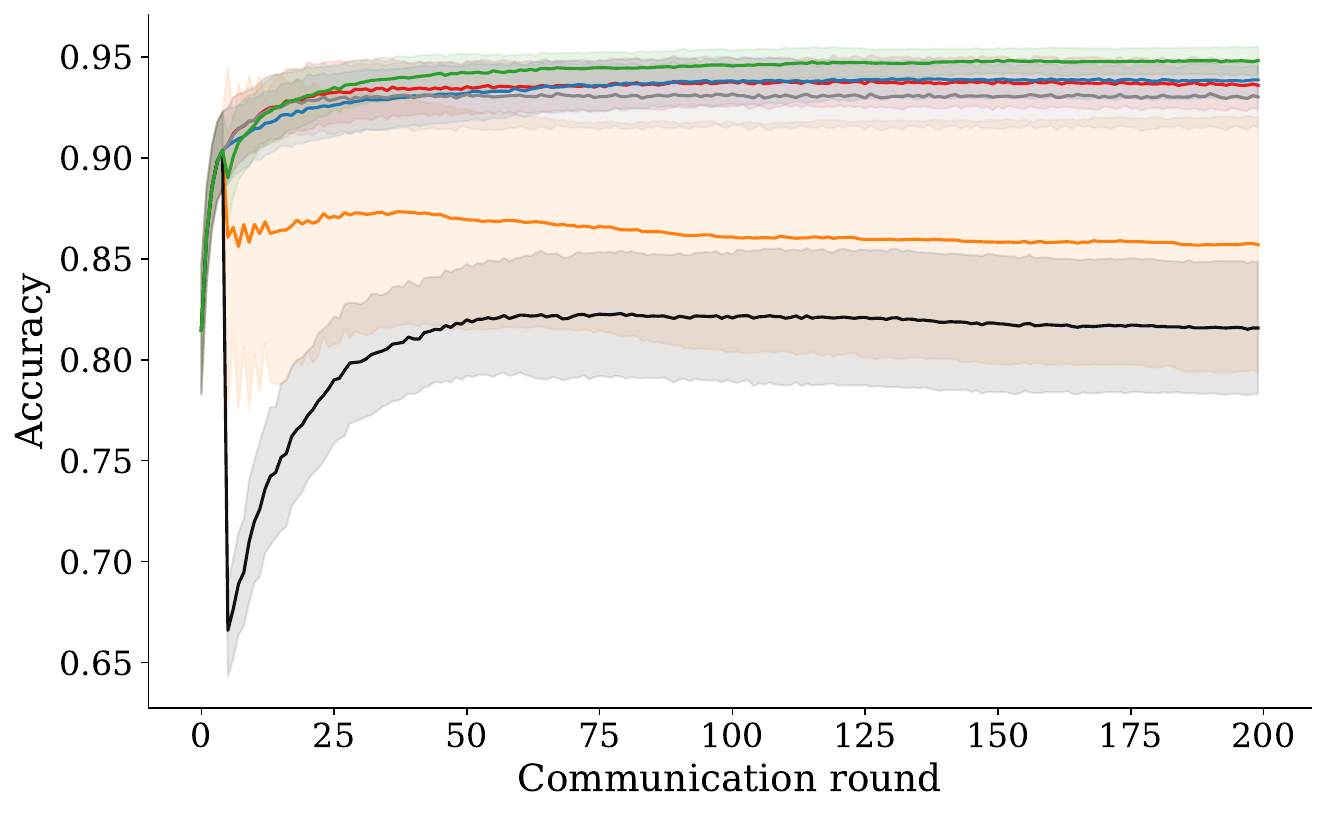}
        \caption{FSR - iid}
       
    \end{subfigure}
    \hfill
    \begin{subfigure}[b]{0.33\textwidth}
        \includegraphics[width=\linewidth]{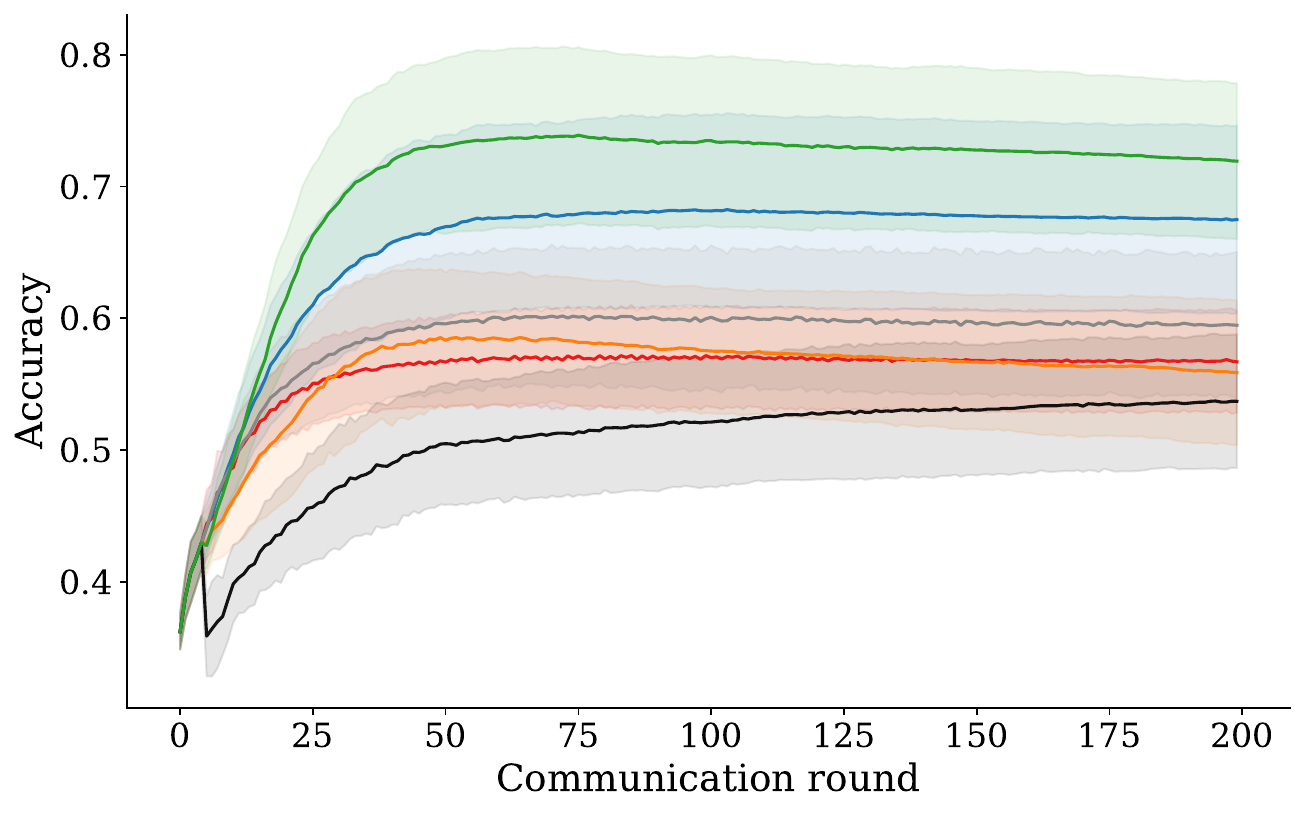}
        \caption{FSR - non-iid (clusters)}
        
    \end{subfigure}
    \hfill
    \begin{subfigure}[b]{0.33\textwidth}
        \includegraphics[width=\linewidth]{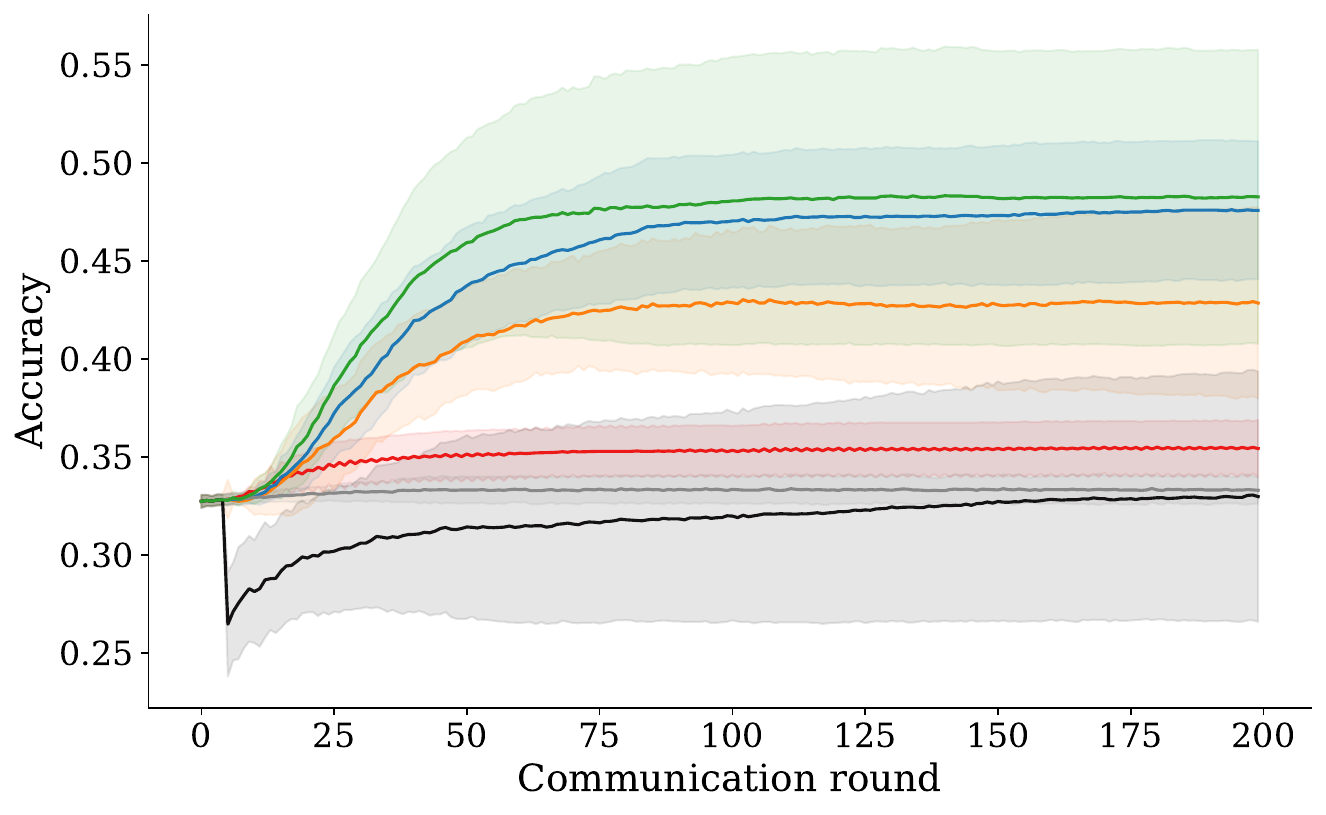}
        \caption{FSR - non-iid (classes)}
       
    \end{subfigure}

    \begin{subfigure}[b]{0.33\textwidth}
        \includegraphics[width=\linewidth]{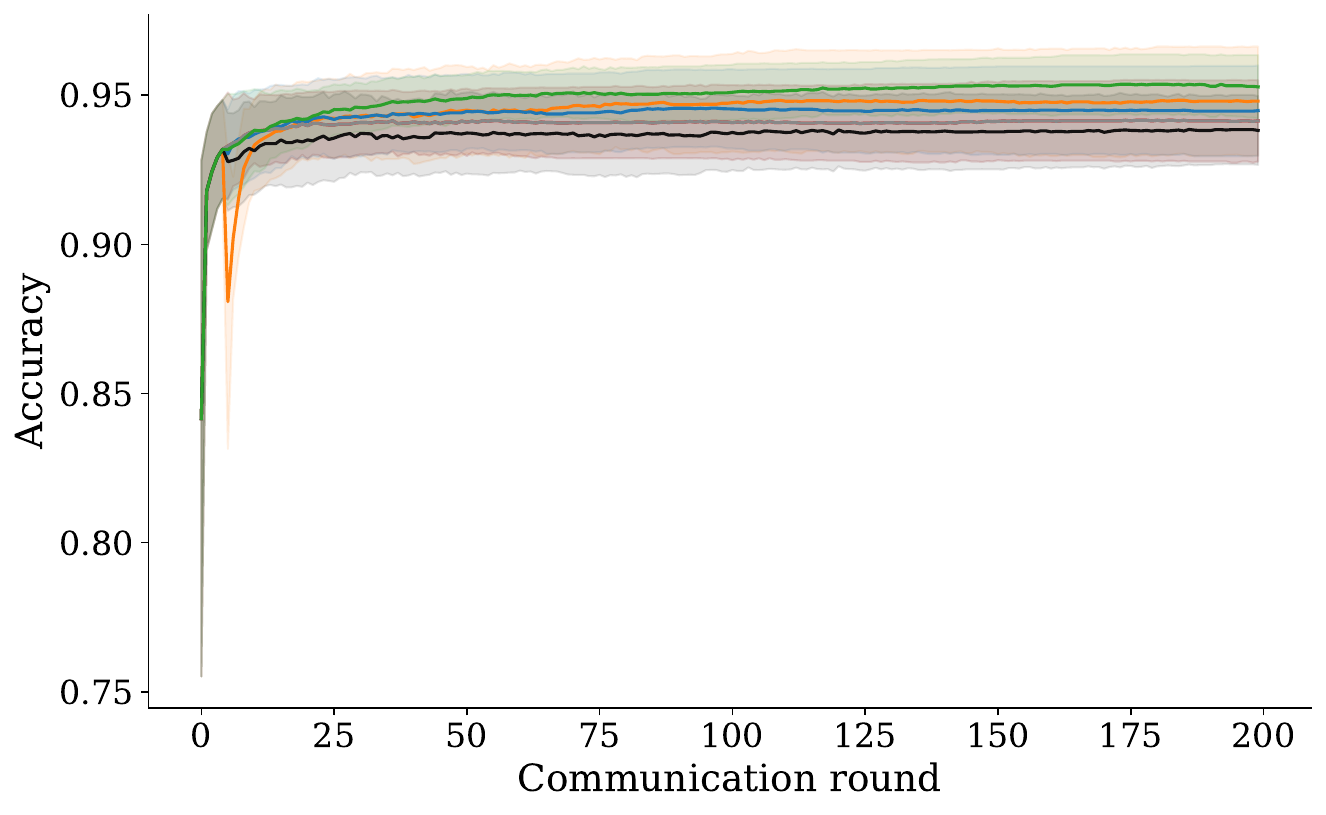}
        \caption{DFedAvgM - iid}
       
    \end{subfigure}
    \hfill
    \begin{subfigure}[b]{0.33\textwidth}
        \includegraphics[width=\linewidth]{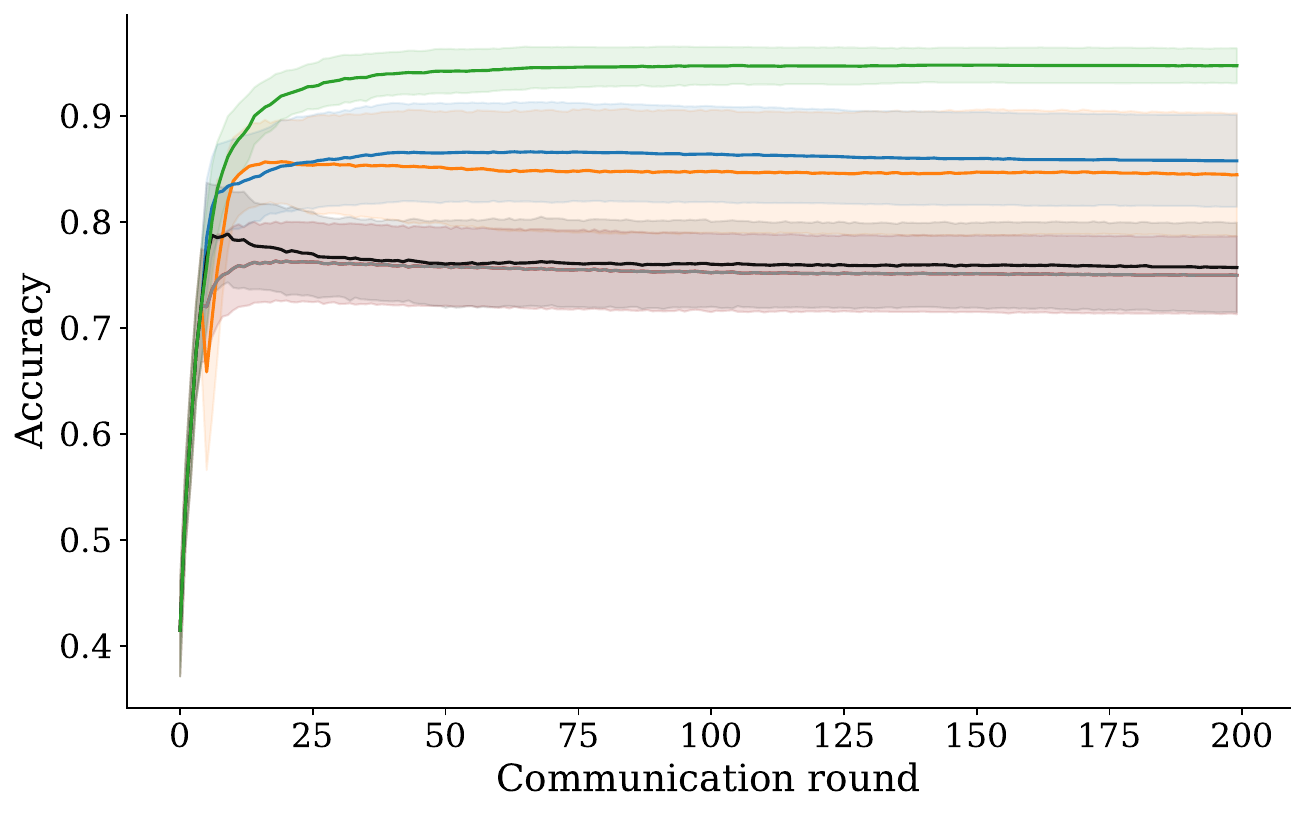}
        \caption{DFedAvgM - non-iid (clusters)}
      
    \end{subfigure}
    \hfill
    \begin{subfigure}[b]{0.33\textwidth}
        \includegraphics[width=\linewidth]{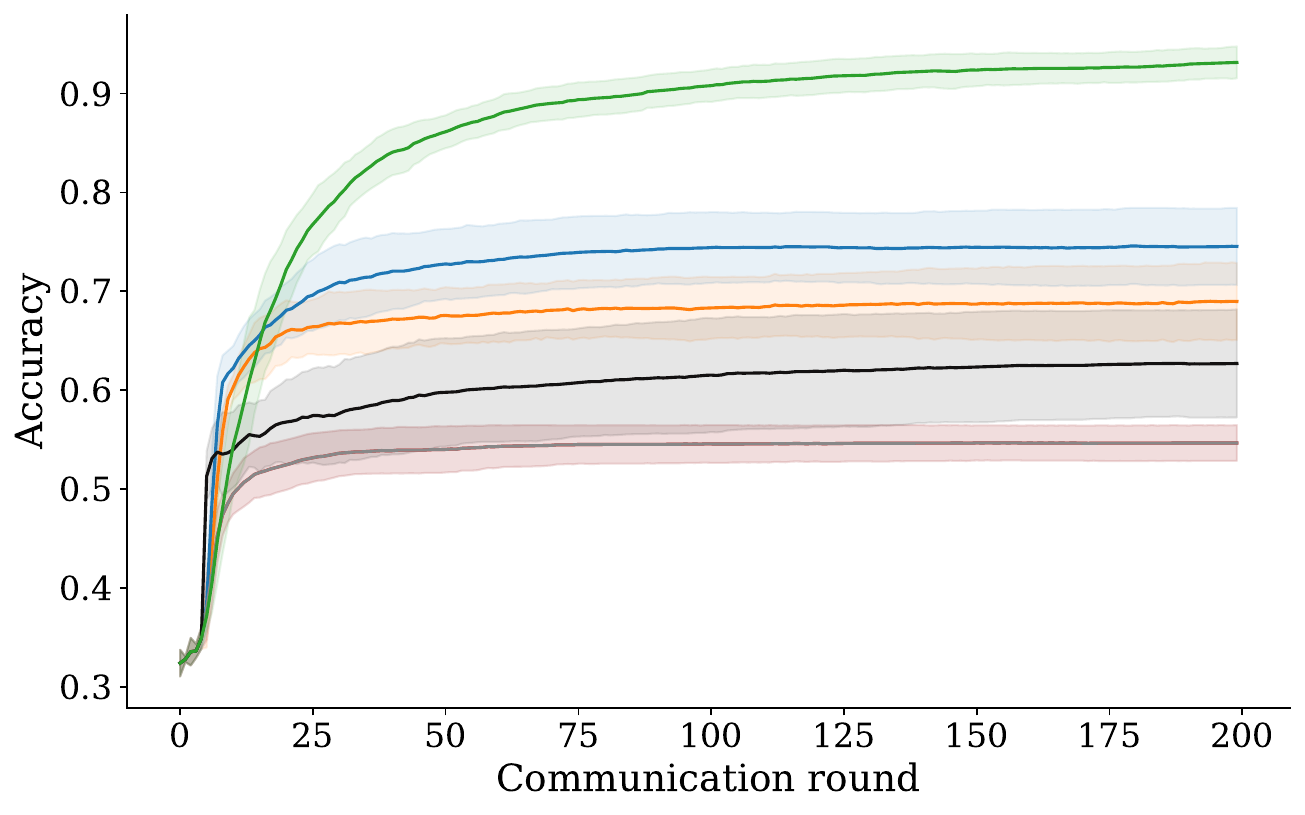}
        \caption{DFedAvgM - non-iid (classes)}
       
    \end{subfigure}

    \caption{Convergence plots for DJAM, FSR, and DFedAvgM algorithms on the digits dataset. Shaded regions represent mean $\pm$ standard deviation across 10 folds, over 200 communication rounds. A random client is dropped persistently after the 5th round. Colors: {\color{modelinversion}model inversion}, {\color{gradientinversion}gradient inversion}, {\color{reference}reference}, {\color{random}random}, {\color{drop}drop}, {\color{noaction}no action}}.
    \label{fig:convergence:digits}
\end{figure*}

In this section we present convergence plots for all combinations of dataset, algorithm and data heterogeneity type. They are grouped by dataset in figures of 9 subplots, where each row has a different algorithm and column, data heterogeneity type. We have plots for wine in \Cref{fig:convergence:wine}, iris in \Cref{fig:convergence:iris}, and digits in \Cref{fig:convergence:digits}.

%% file: chapters/7.2.appendix.tex
\section{All tables}
\label{app:exp:all-tables}

\begin{table*}[h!]
    \centering
    \setlength{\tabcolsep}{2pt}
    \input{figures/fedKDD-tabular2-plots/tables/variant2-table}
    \caption{Mean accuracy of clients on a holdout test set after 200 communication rounds for all algorithms.}
    \label{tab:results:all-algos}
\end{table*}

In this section, we present all results in one table (\Cref{tab:results:all-algos}).

%% file: figures/fedKDD-tabular2-plots/tables/variant2-table.tex
\begin{tabular}{ccl|ccccc|c}

 \textbf{Algorithm}   & \textbf{Dataset}   & \textbf{Distribution}       & \textbf{No action}       & \textbf{Forget}            & \textbf{Random}          & \textbf{Grad inv}        & \textbf{Model inv}       & \textbf{Reference}       \\
\hline \hline
 \multirow{9}{*}{DJAM} & \multirow{3}{*}{wine}  & iid  & $0.53 \pm 0.14$ & $0.80 \pm 0.05$ & $0.87 \pm 0.08$ & $0.74 \pm 0.17$ & $0.88 \pm 0.05$ & $0.91 \pm 0.08$ \\
         &       & non-iid (clusters) & $0.41 \pm 0.08$ & $0.56 \pm 0.02$ & $0.64 \pm 0.03$ & $0.66 \pm 0.05$ & $0.62 \pm 0.05$ & $0.95 \pm 0.04$ \\
         &       & non-iid (class)    & $0.38 \pm 0.08$ & $0.57 \pm 0.04$ & $0.65 \pm 0.02$ & $0.69 \pm 0.04$ & $0.64 \pm 0.05$ & $0.94 \pm 0.05$ \\ \hline
         & \multirow{3}{*}{iris}      & iid                & $0.42 \pm 0.18$ & $0.65 \pm 0.08$ & $0.73 \pm 0.08$ & $0.57 \pm 0.21$ & $0.69 \pm 0.12$ & $0.71 \pm 0.08$ \\
         &       & non-iid (clusters) & $0.49 \pm 0.12$ & $0.55 \pm 0.08$ & $0.57 \pm 0.13$ & $0.59 \pm 0.15$ & $0.63 \pm 0.16$ & $0.66 \pm 0.11$ \\
         &       & non-iid (class)    & $0.56 \pm 0.08$ & $0.59 \pm 0.03$ & $0.65 \pm 0.04$ & $0.67 \pm 0.02$ & $0.64 \pm 0.07$ & $0.78 \pm 0.06$ \\ \hline
         & \multirow{3}{*}{digits}    & iid                & $0.73 \pm 0.08$ & $0.84 \pm 0.03$ & $0.86 \pm 0.02$ & $0.87 \pm 0.02$ & $0.86 \pm 0.02$ & $0.89 \pm 0.02$ \\
         &     & non-iid (clusters) & $0.65 \pm 0.10$ & $0.65 \pm 0.06$ & $0.64 \pm 0.05$ & $0.70 \pm 0.06$ & $0.69 \pm 0.05$ & $0.90 \pm 0.02$ \\
         &     & non-iid (class)    & $0.41 \pm 0.05$ & $0.54 \pm 0.02$ & $0.60 \pm 0.05$ & $0.62 \pm 0.05$ & $0.61 \pm 0.05$ & $0.92 \pm 0.02$ \\\hline \hline
 \multirow{9}{*}{FSR}  & \multirow{3}{*}{wine}  & iid   & $0.65 \pm 0.17$ & $0.86 \pm 0.09$ & $0.90 \pm 0.05$ & $0.94 \pm 0.03$ & $0.96 \pm 0.04$ & $0.95 \pm 0.04$ \\
          &       & non-iid (clusters) & $0.61 \pm 0.11$ & $0.71 \pm 0.12$ & $0.71 \pm 0.11$ & $0.81 \pm 0.13$ & $0.80 \pm 0.14$ & $0.96 \pm 0.04$ \\
          &       & non-iid (class)    & $0.58 \pm 0.08$ & $0.63 \pm 0.06$ & $0.67 \pm 0.05$ & $0.77 \pm 0.12$ & $0.74 \pm 0.13$ & $0.96 \pm 0.03$ \\ \hline
          & \multirow{3}{*}{iris}      & iid                & $0.73 \pm 0.07$ & $0.87 \pm 0.05$ & $0.91 \pm 0.07$ & $0.83 \pm 0.15$ & $0.90 \pm 0.09$ & $0.93 \pm 0.05$ \\
          &       & non-iid (clusters) & $0.71 \pm 0.07$ & $0.75 \pm 0.13$ & $0.76 \pm 0.14$ & $0.79 \pm 0.15$ & $0.85 \pm 0.11$ & $0.91 \pm 0.06$ \\
          &       & non-iid (class)    & $0.67 \pm 0.08$ & $0.65 \pm 0.06$ & $0.64 \pm 0.07$ & $0.71 \pm 0.08$ & $0.68 \pm 0.03$ & $0.95 \pm 0.06$ \\ \hline
          & \multirow{3}{*}{digits}    & iid                & $0.93 \pm 0.02$ & $0.94 \pm 0.01$ & $0.82 \pm 0.03$ & $0.86 \pm 0.06$ & $0.94 \pm 0.01$ & $0.95 \pm 0.01$ \\
          &     & non-iid (clusters) & $0.59 \pm 0.06$ & $0.57 \pm 0.04$ & $0.54 \pm 0.05$ & $0.56 \pm 0.05$ & $0.67 \pm 0.07$ & $0.72 \pm 0.06$ \\
          &     & non-iid (class)    & $0.33 \pm 0.01$ & $0.35 \pm 0.01$ & $0.33 \pm 0.06$ & $0.43 \pm 0.05$ & $0.48 \pm 0.04$ & $0.48 \pm 0.07$ \\ \hline \hline
\multirow{9}{*}{DFedAvgM} & \multirow{3}{*}{wine} & iid & $0.96 \pm 0.03$ & $0.96 \pm 0.03$ & $0.90 \pm 0.06$ & $0.97 \pm 0.03$ & $0.97 \pm 0.03$ & $0.97 \pm 0.03$ \\
     &       & non-iid (clusters) & $0.62 \pm 0.10$ & $0.62 \pm 0.10$ & $0.64 \pm 0.11$ & $0.78 \pm 0.14$ & $0.86 \pm 0.10$ & $0.99 \pm 0.02$ \\
     &       & non-iid (class)    & $0.55 \pm 0.01$ & $0.55 \pm 0.01$ & $0.63 \pm 0.07$ & $0.71 \pm 0.08$ & $0.82 \pm 0.05$ & $0.97 \pm 0.03$ \\ \hline
     & \multirow{3}{*}{iris}      & iid                & $0.90 \pm 0.04$ & $0.90 \pm 0.04$ & $0.89 \pm 0.09$ & $0.92 \pm 0.09$ & $0.95 \pm 0.04$ & $0.97 \pm 0.04$ \\
     &       & non-iid (clusters) & $0.64 \pm 0.11$ & $0.64 \pm 0.11$ & $0.70 \pm 0.17$ & $0.79 \pm 0.17$ & $0.87 \pm 0.12$ & $0.94 \pm 0.05$ \\
     &       & non-iid (class)    & $0.57 \pm 0.04$ & $0.57 \pm 0.04$ & $0.57 \pm 0.13$ & $0.62 \pm 0.10$ & $0.73 \pm 0.08$ & $0.84 \pm 0.04$ \\ \hline
     & \multirow{3}{*}{digits}    & iid                & $0.94 \pm 0.01$ & $0.94 \pm 0.01$ & $0.94 \pm 0.01$ & $0.95 \pm 0.02$ & $0.94 \pm 0.02$ & $0.95 \pm 0.01$ \\
     &     & non-iid (clusters) & $0.75 \pm 0.04$ & $0.75 \pm 0.04$ & $0.76 \pm 0.04$ & $0.84 \pm 0.06$ & $0.86 \pm 0.04$ & $0.95 \pm 0.02$ \\
     &     & non-iid (class)    & $0.55 \pm 0.02$ & $0.55 \pm 0.02$ & $0.63 \pm 0.05$ & $0.69 \pm 0.04$ & $0.75 \pm 0.04$ & $0.93 \pm 0.02$ \\

\end{tabular}

%% file: chapters/7.3.appendix.tex
\section{What Data is Extracted}
\label{app:extracted-data}

This section presents qualitative examples of images reconstructed using model inversion and gradient inversion attacks on the Digits dataset. 

\begin{figure*}[t]
    \centering
    \begin{subfigure}[b]{0.48\textwidth}
        \centering
        \includegraphics[width=\textwidth]{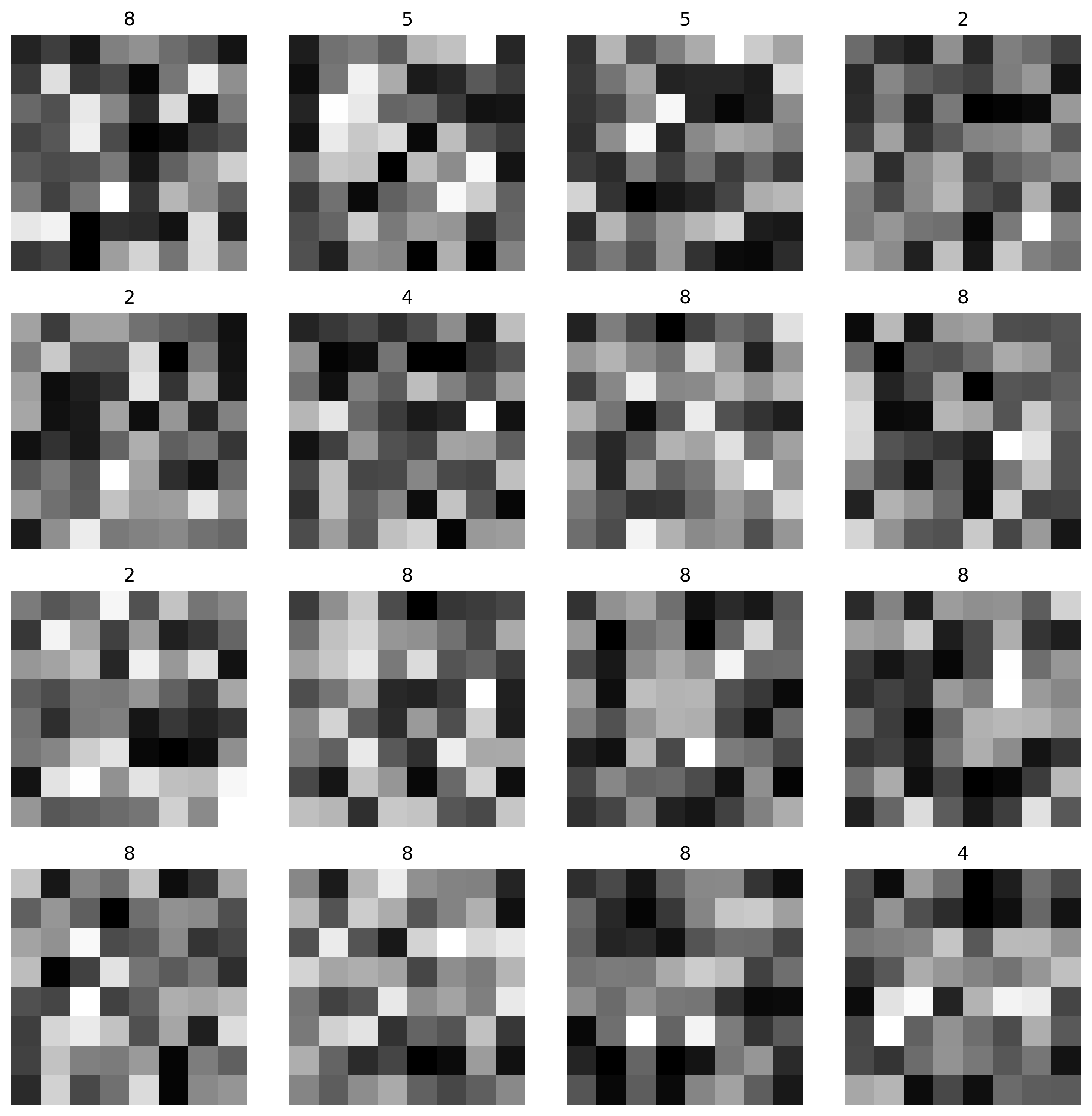}
        \caption{After 1 epoch}
        \label{fig:dfedavgm-mi-1}
    \end{subfigure}
    \hfill
    \begin{subfigure}[b]{0.48\textwidth}
        \centering
        \includegraphics[width=\textwidth]{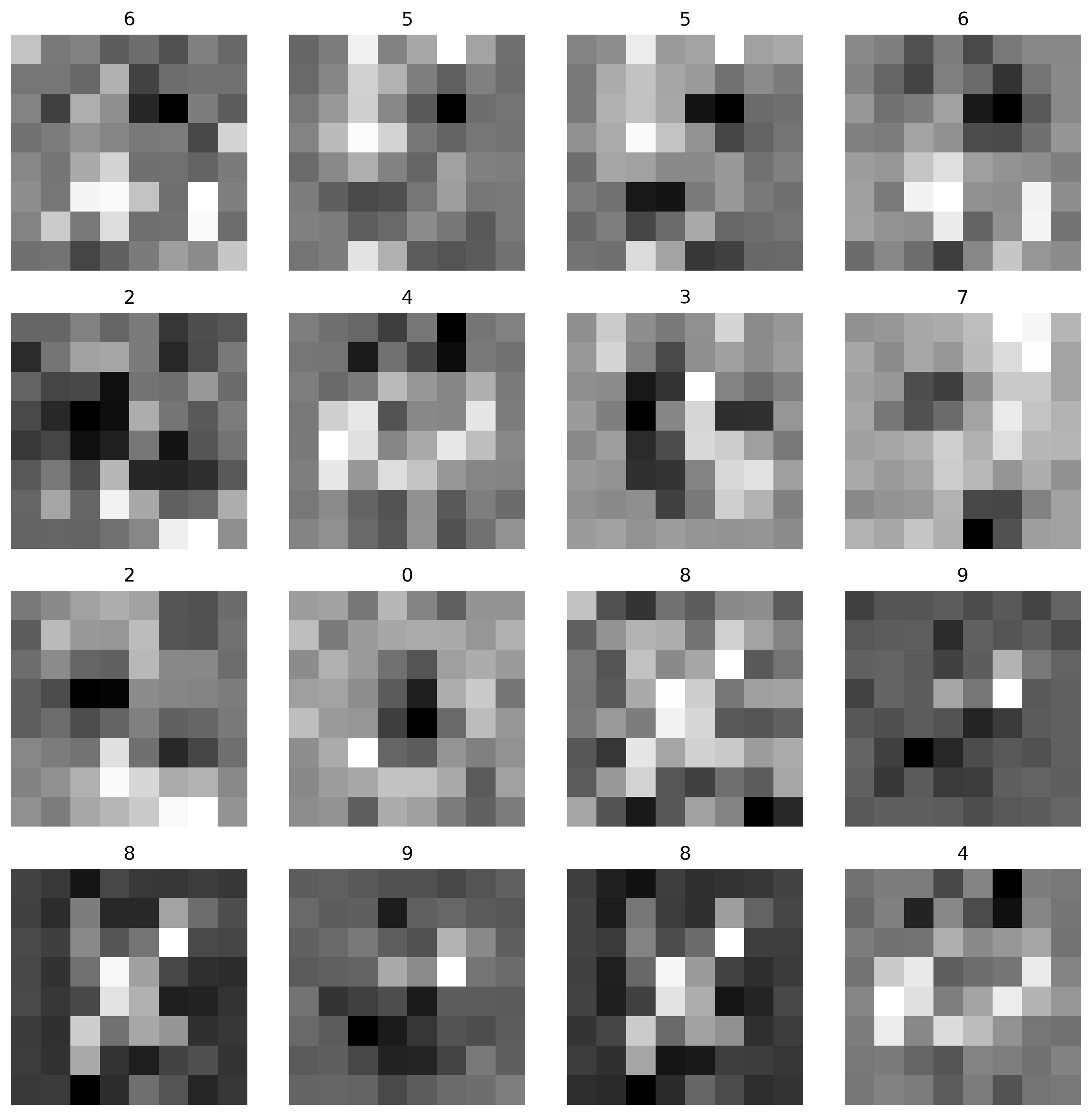}
        \caption{After 800 epochs}
        \label{fig:dfedavgm-mi-800}
    \end{subfigure}
    \caption{Reconstructed images using model inversion on the Digits dataset (DFedAvgM, non-iid (class)).}
    \label{fig:app:images:dfedavg:mi}
\end{figure*}

\begin{figure*}[t]
    \centering
    \begin{subfigure}[b]{0.48\textwidth}
        \centering
        \includegraphics[width=\textwidth]{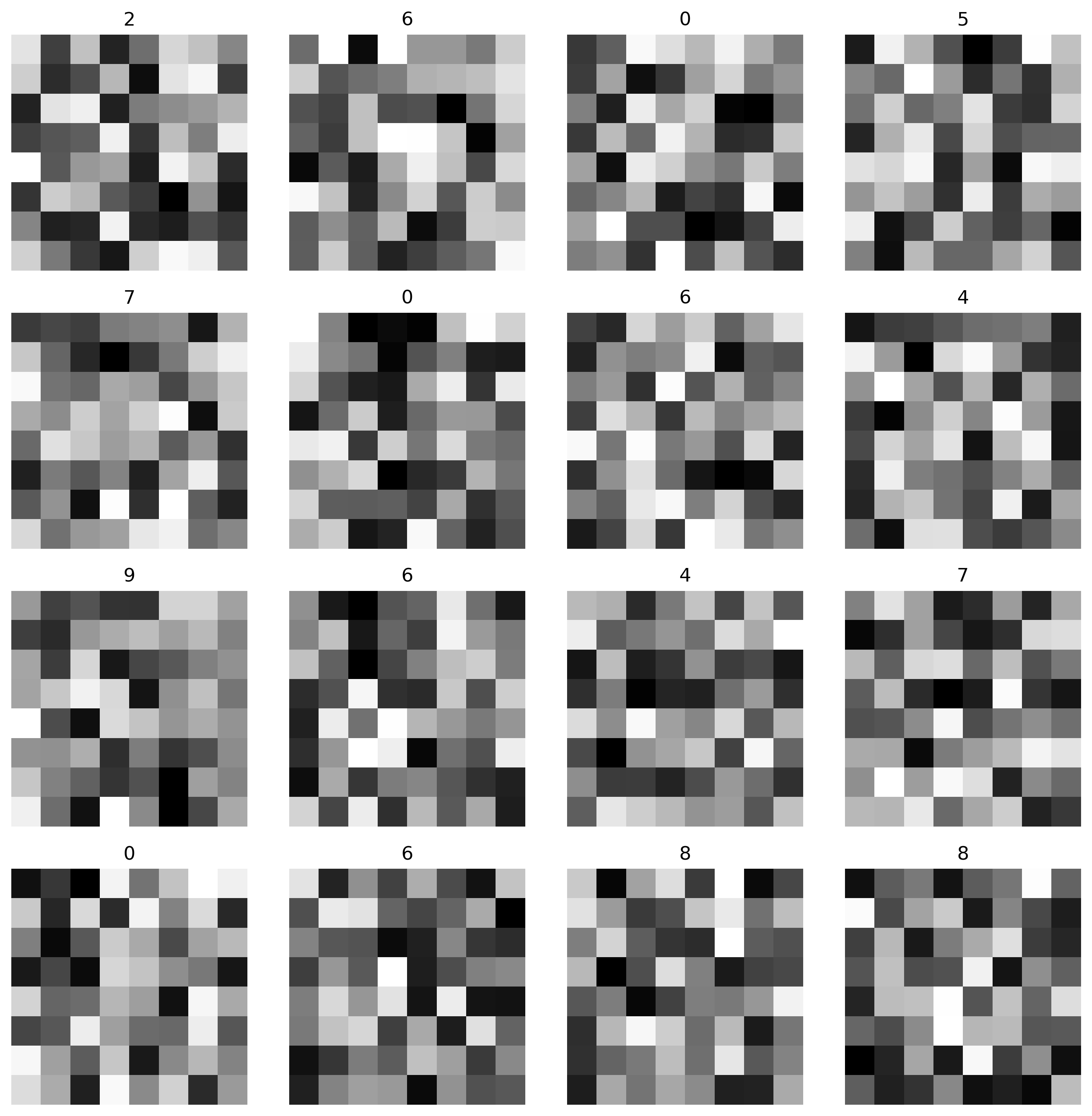}
        \caption{After 1 epoch}
        \label{fig:dfedavgm-grad-1}
    \end{subfigure}
    \hfill
    \begin{subfigure}[b]{0.48\textwidth}
        \centering
        \includegraphics[width=\textwidth]{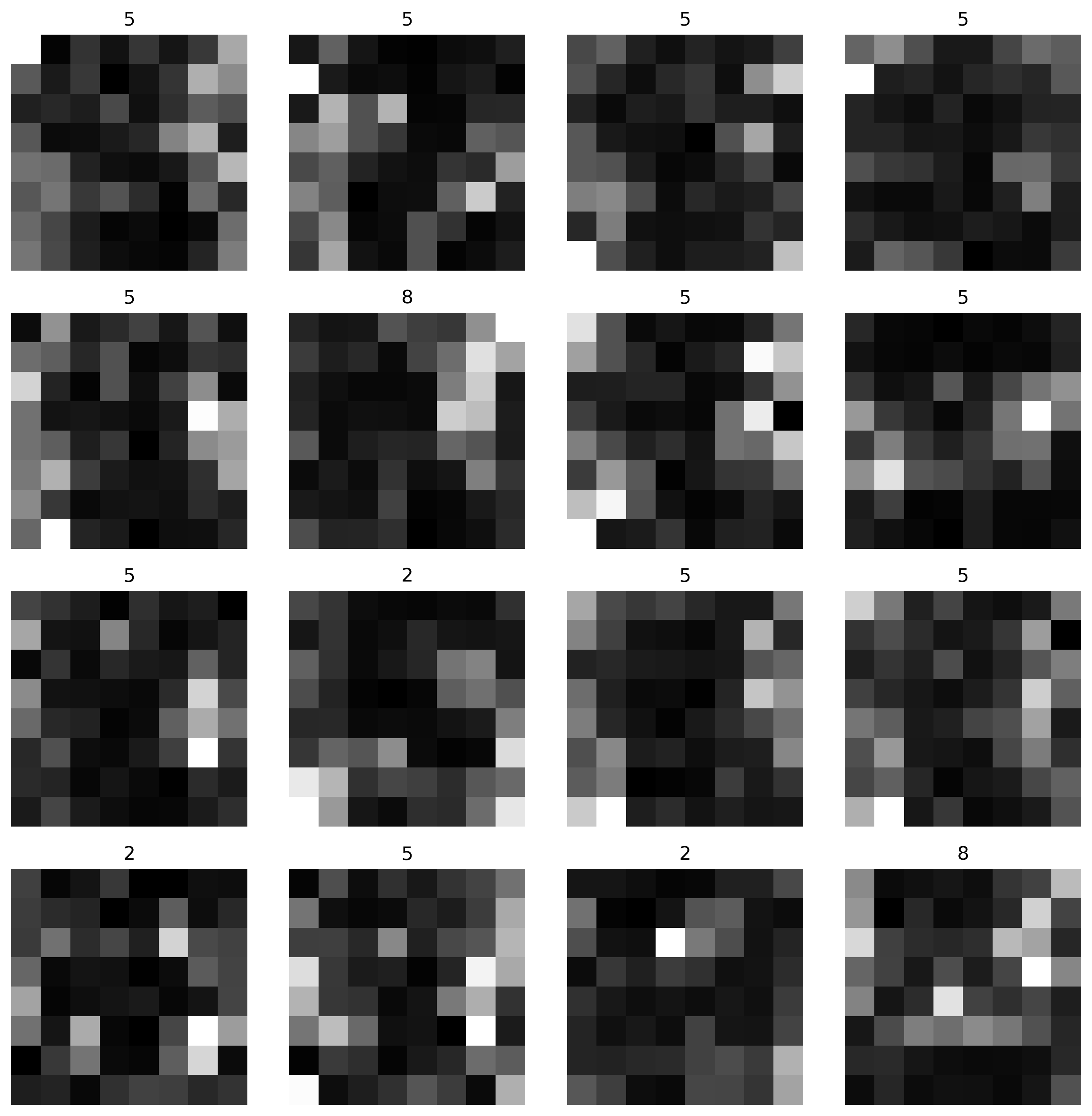}
        \caption{After 1500 epochs}
        \label{fig:dfedavgm-grad-1500}
    \end{subfigure}
    \caption{Reconstructed images using gradient inversion on the Digits dataset (DFedAvgM, non-iid (class)).}
    \label{fig:app:images:dfedavg:gi}
\end{figure*}

\begin{figure*}[t]
    \centering
    \begin{subfigure}[b]{0.48\textwidth}
        \centering
        \includegraphics[width=\textwidth]{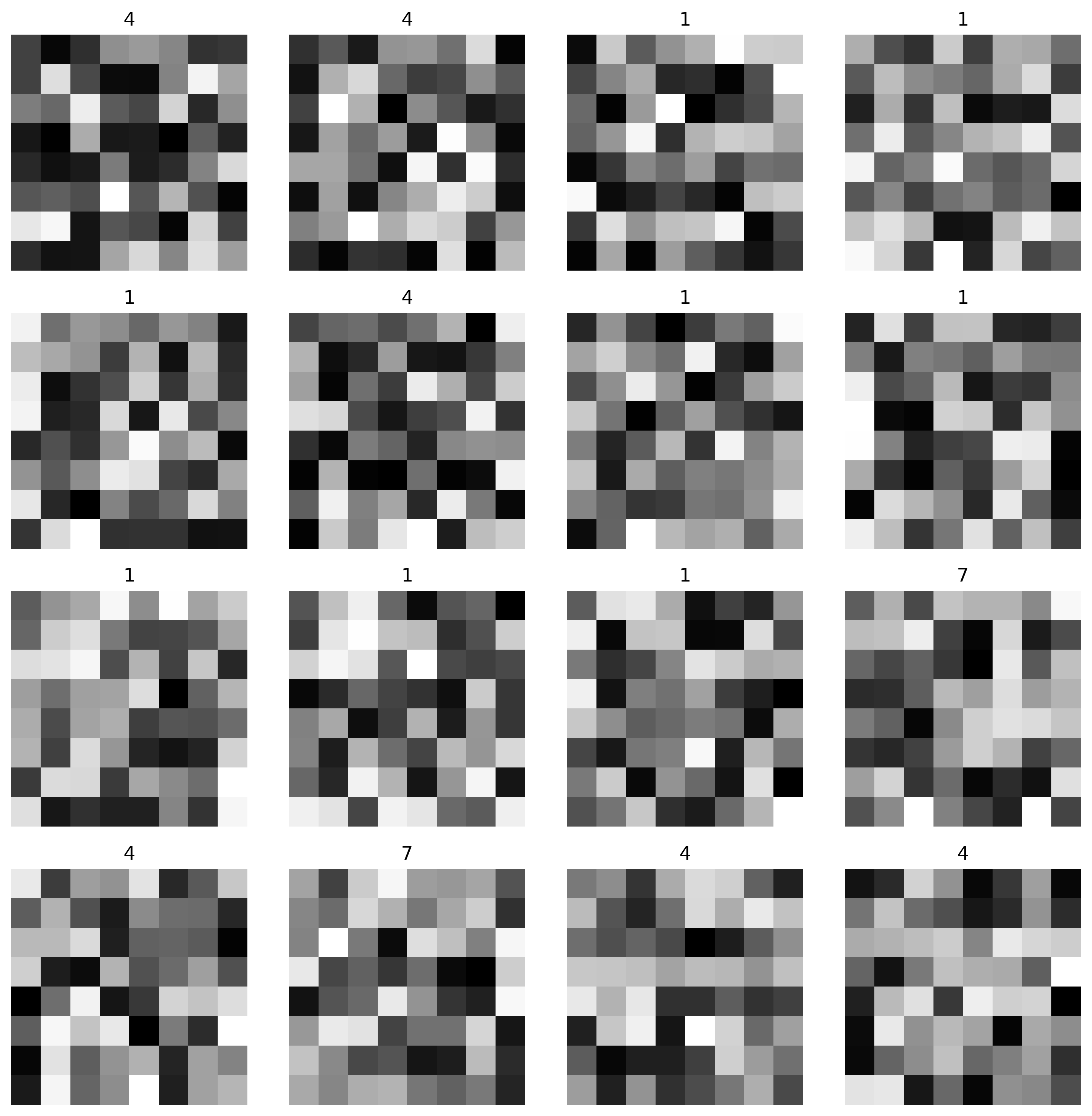}
        \caption{After 1 epoch}
        \label{fig:djam-mi-1}
    \end{subfigure}
    \hfill
    \begin{subfigure}[b]{0.48\textwidth}
        \centering
        \includegraphics[width=\textwidth]{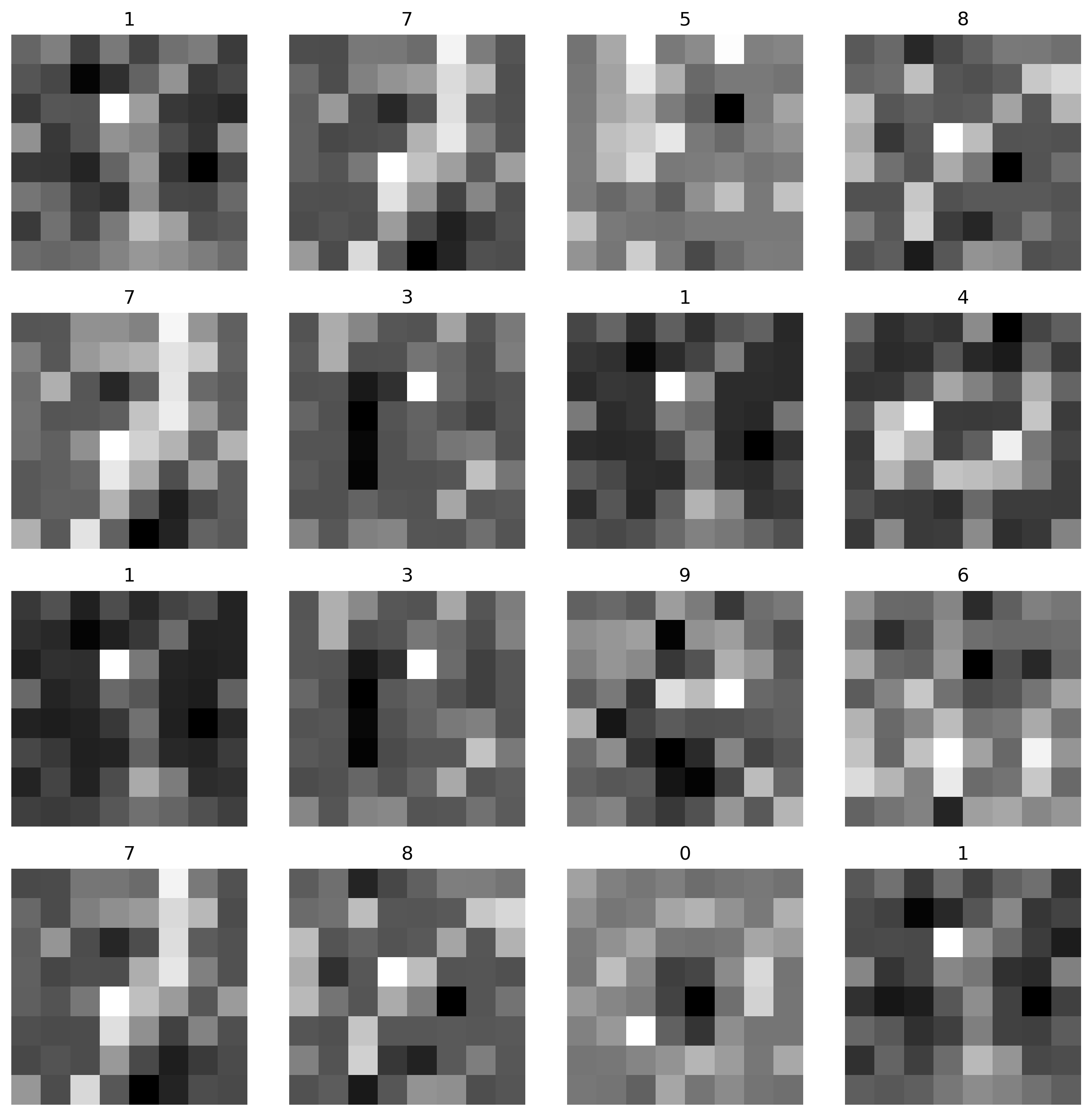}
        \caption{After 800 epochs}
        \label{fig:djam-mi-800}
    \end{subfigure}
    \caption{Reconstructed images using model inversion on the Digits dataset (DJAM, non-iid (class)).}
    \label{fig:app:images:djam:mi}
\end{figure*}

\begin{figure*}[t]
    \centering
    \begin{subfigure}[b]{0.48\textwidth}
        \centering
        \includegraphics[width=\textwidth]{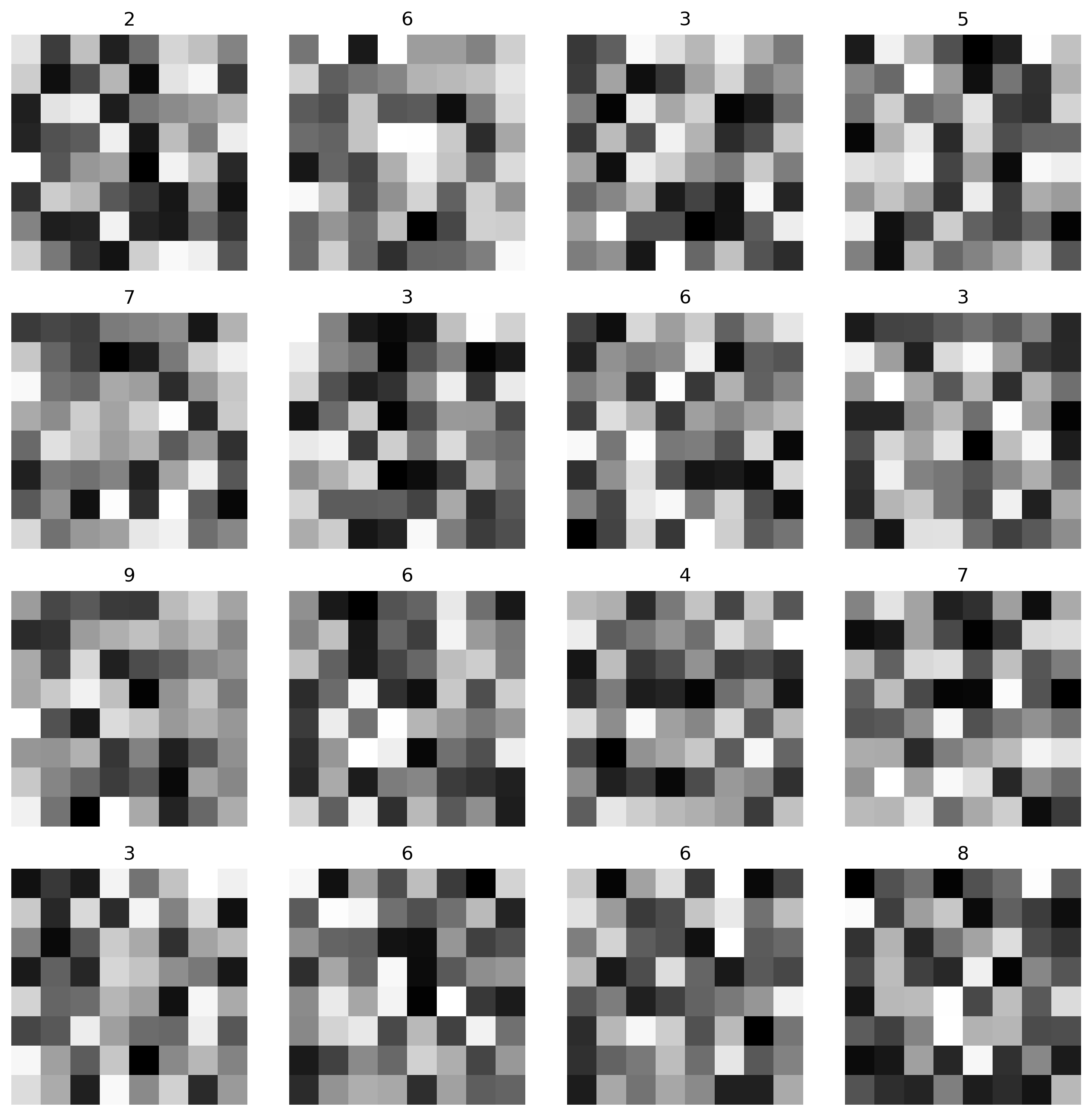}
        \caption{After 1 epoch}
        \label{fig:djam-grad-1}
    \end{subfigure}
    \hfill
    \begin{subfigure}[b]{0.48\textwidth}
        \centering
        \includegraphics[width=\textwidth]{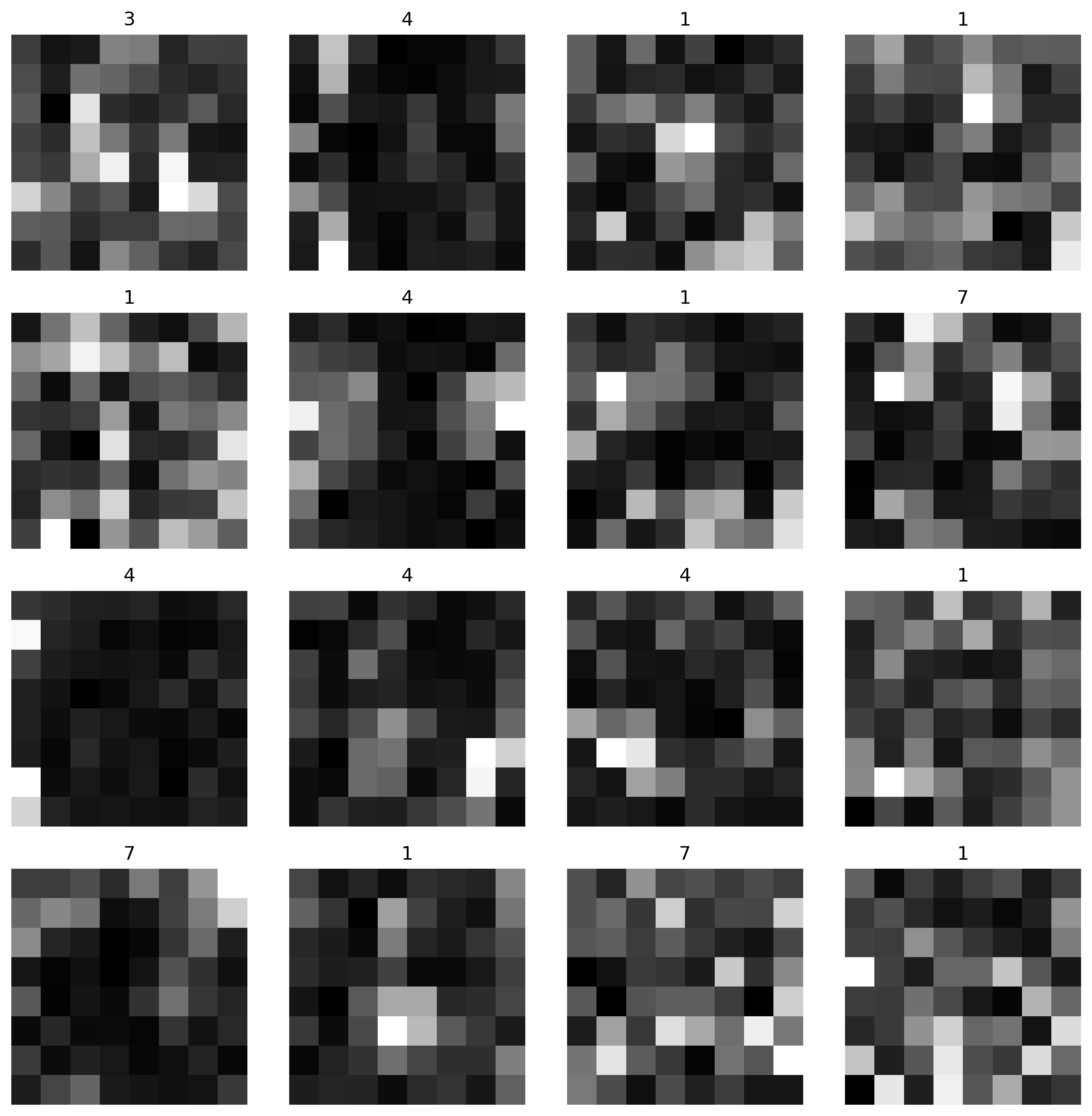}
        \caption{After 1500 epochs}
        \label{fig:djam-grad-1500}
    \end{subfigure}
    \caption{Reconstructed images using gradient inversion on the Digits dataset (DJAM, non-iid (class)).}
    \label{fig:app:images:djam:gi}
\end{figure*}

For model inversion (\Cref{fig:app:images:dfedavg:mi,fig:app:images:djam:mi}), we show reconstructed samples after 1 epoch (left) and 800 epochs (right) of attack optimization. For gradient inversion (\Cref{fig:app:images:dfedavg:gi,fig:app:images:djam:gi}), we present reconstructions at 1 epoch (left) and 1500 epochs (right). 

In general, model inversion appears to recover structural patterns resembling noisy or averaged digit shapes. While the reconstructions are imperfect, they are not random and reflect underlying data structure. Gradient inversion yields darker, lower-contrast images with less visible structure, though still non-random. 

Based on both this qualitative inspection and the quantitative performance gains shown in \Cref{sec:experiments}, we hypothesize that the reconstructed images—though degraded—retain useful statistical biases, enabling the virtual client to partially preserve the contribution of the original client lost due to persistent dropout.

%% file: chapters/7.4.appendix.tex
\section{Similarity Plots}
\label{app:sec:similarity}

\begin{figure*}[htbp]
    \centering
    \begin{subfigure}[b]{0.33\textwidth}
        \includegraphics[width=\linewidth]{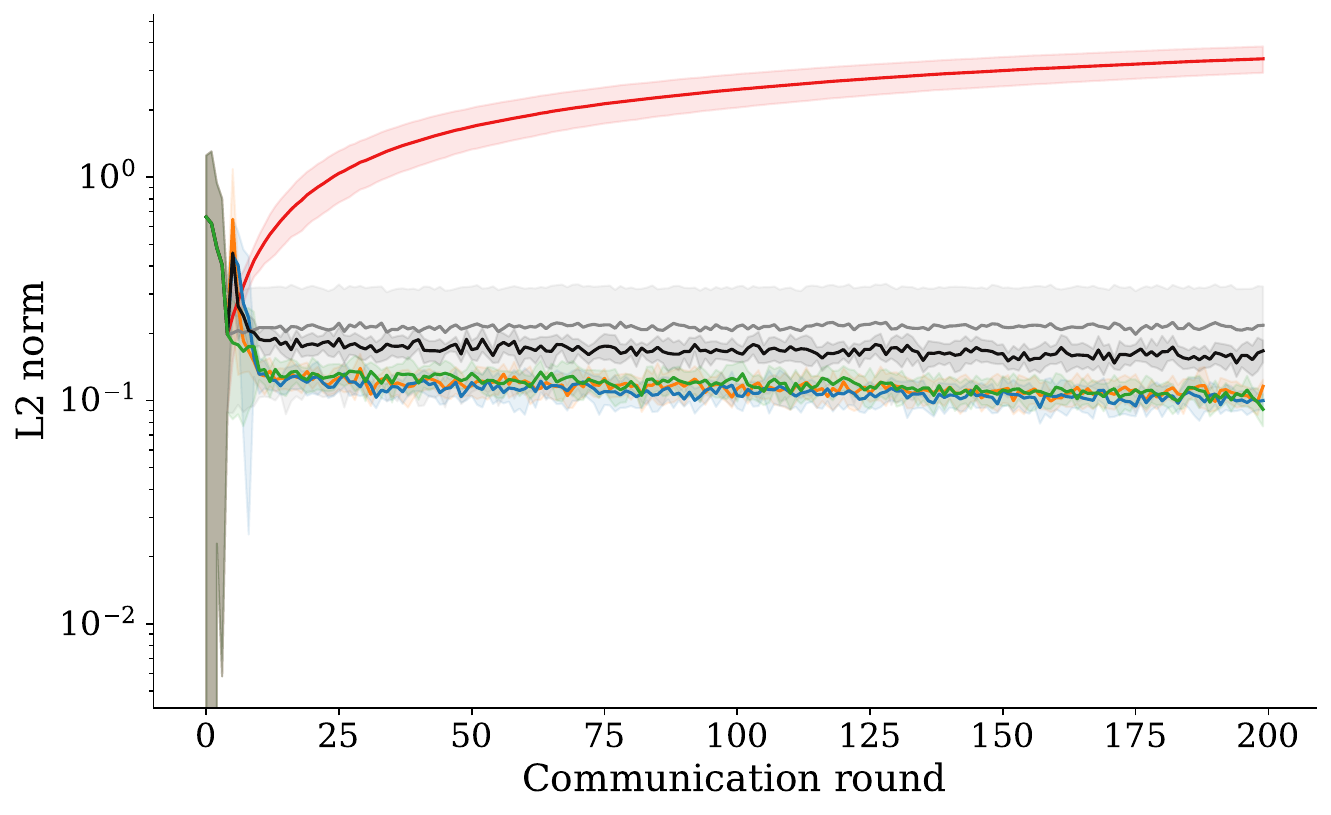}
        \caption{DJAM - iid}
       
    \end{subfigure}
    \hfill
    \begin{subfigure}[b]{0.33\textwidth}
        \includegraphics[width=\linewidth]{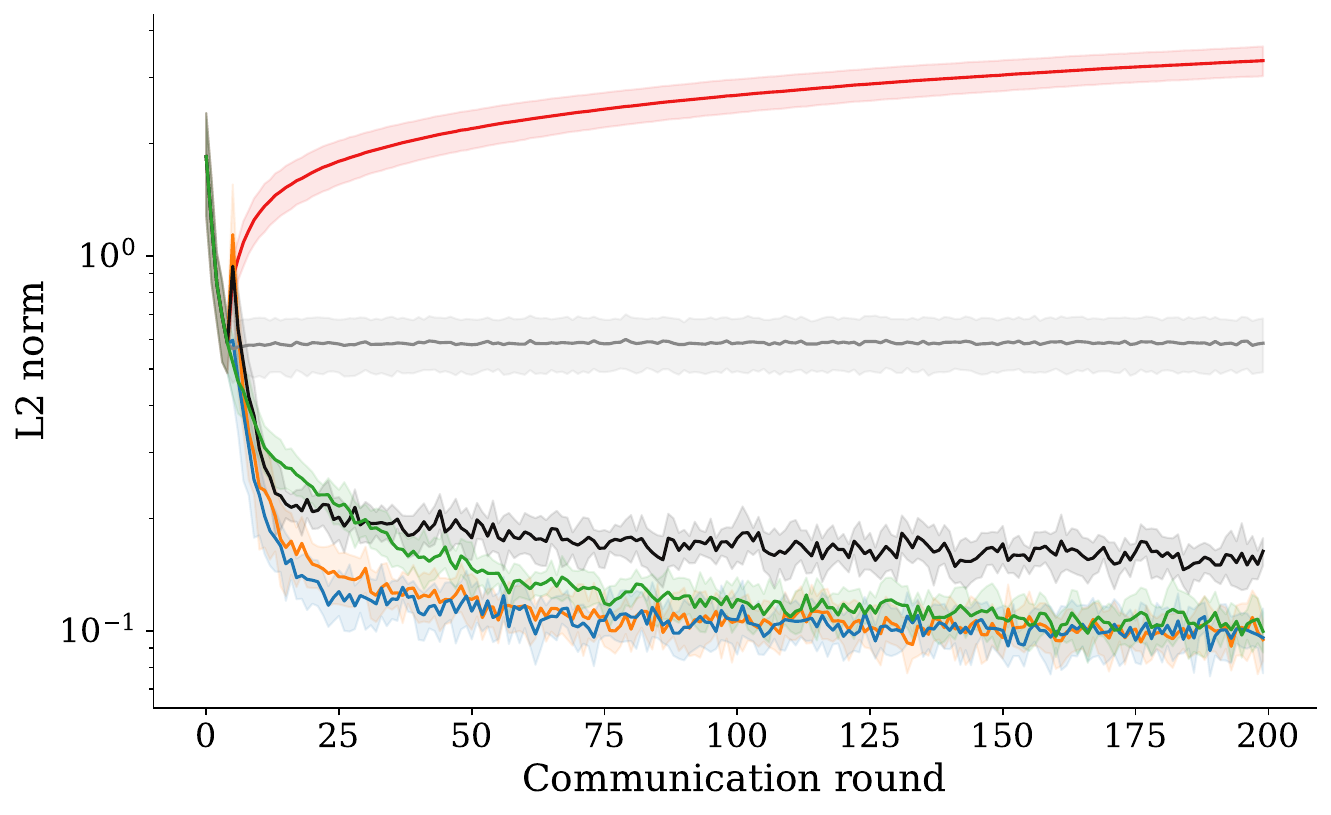}
        \caption{DJAM - non-iid (clusters)}
       
    \end{subfigure}
    \hfill
    \begin{subfigure}[b]{0.33\textwidth}
        \includegraphics[width=\linewidth]{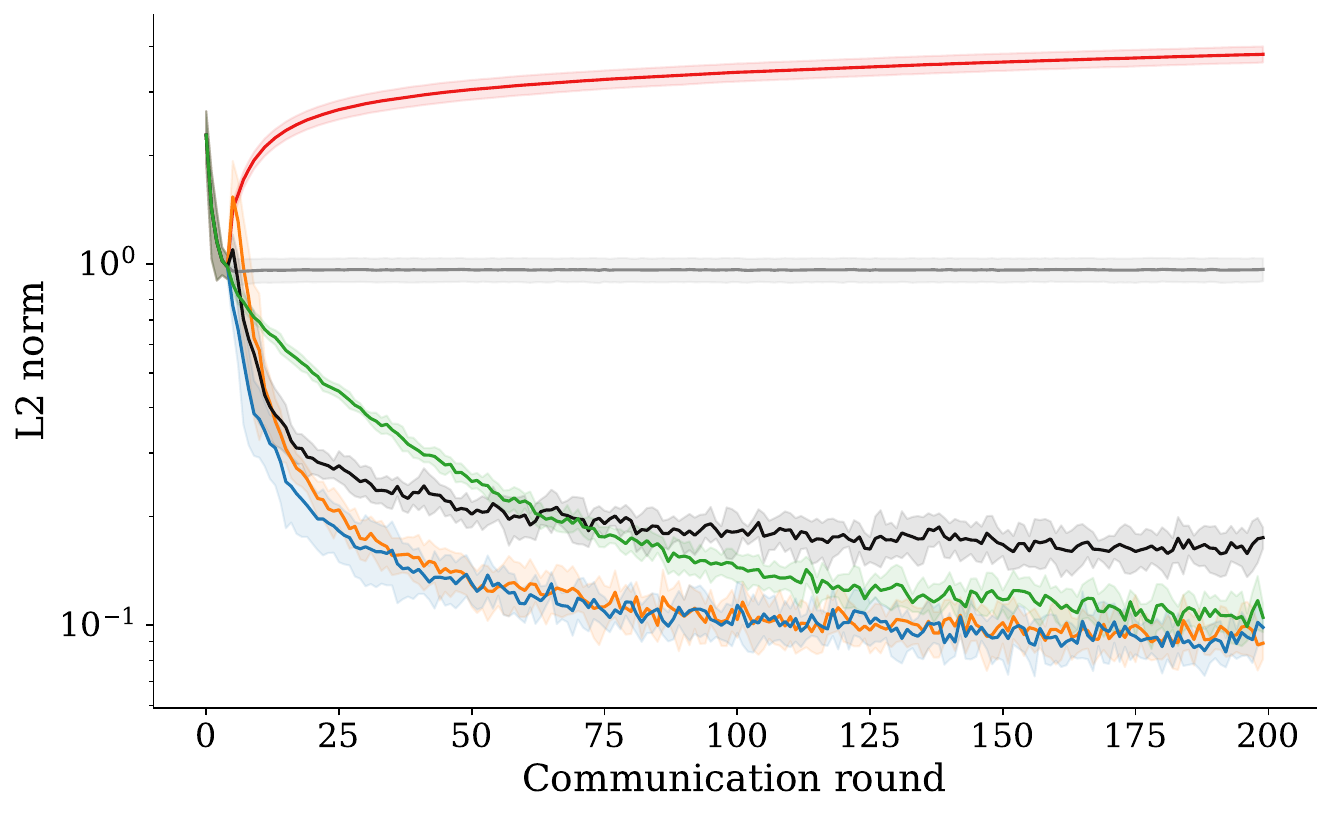}
        \caption{DJAM - non-iid (classes)}
 
    \end{subfigure}

    \begin{subfigure}[b]{0.33\textwidth}
        \includegraphics[width=\linewidth]{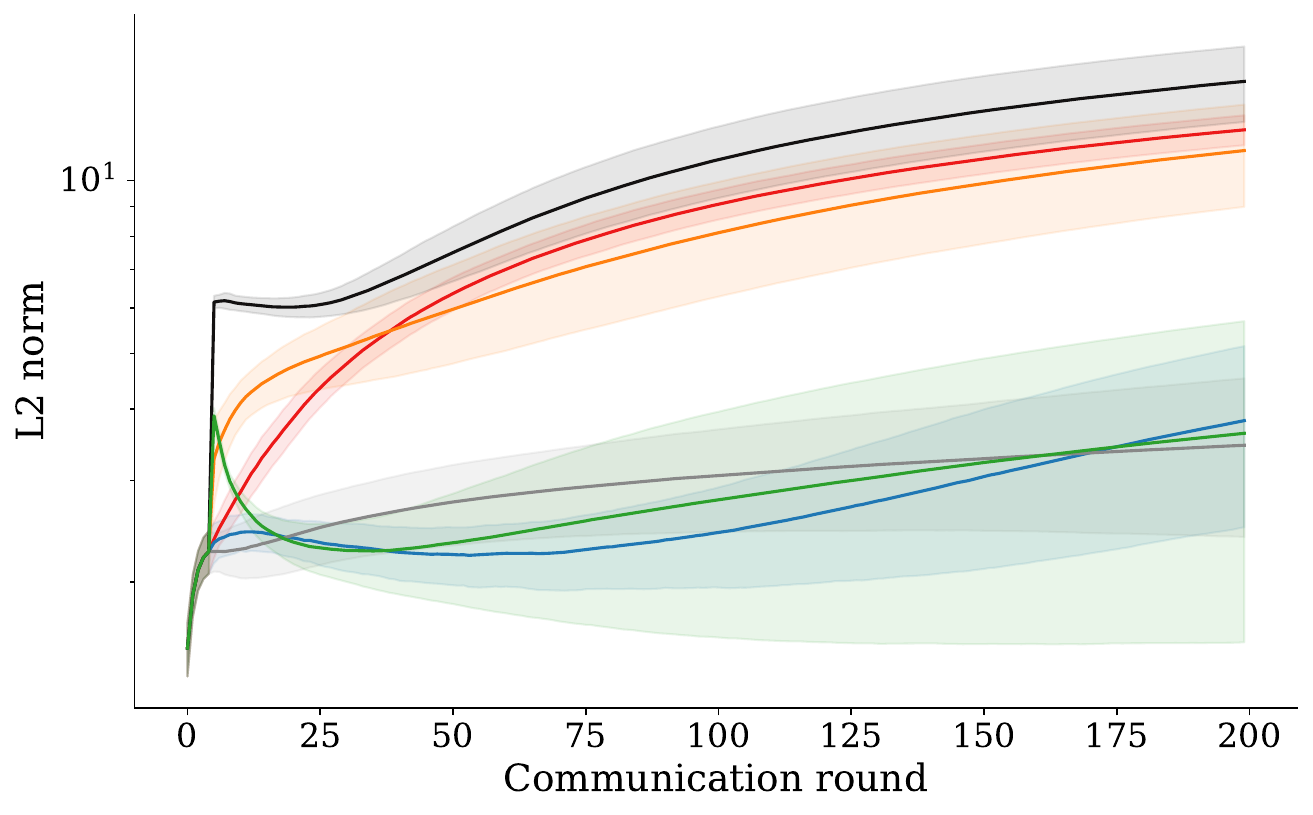}
        \caption{FSR - iid}
     
    \end{subfigure}
    \hfill
    \begin{subfigure}[b]{0.33\textwidth}
        \includegraphics[width=\linewidth]{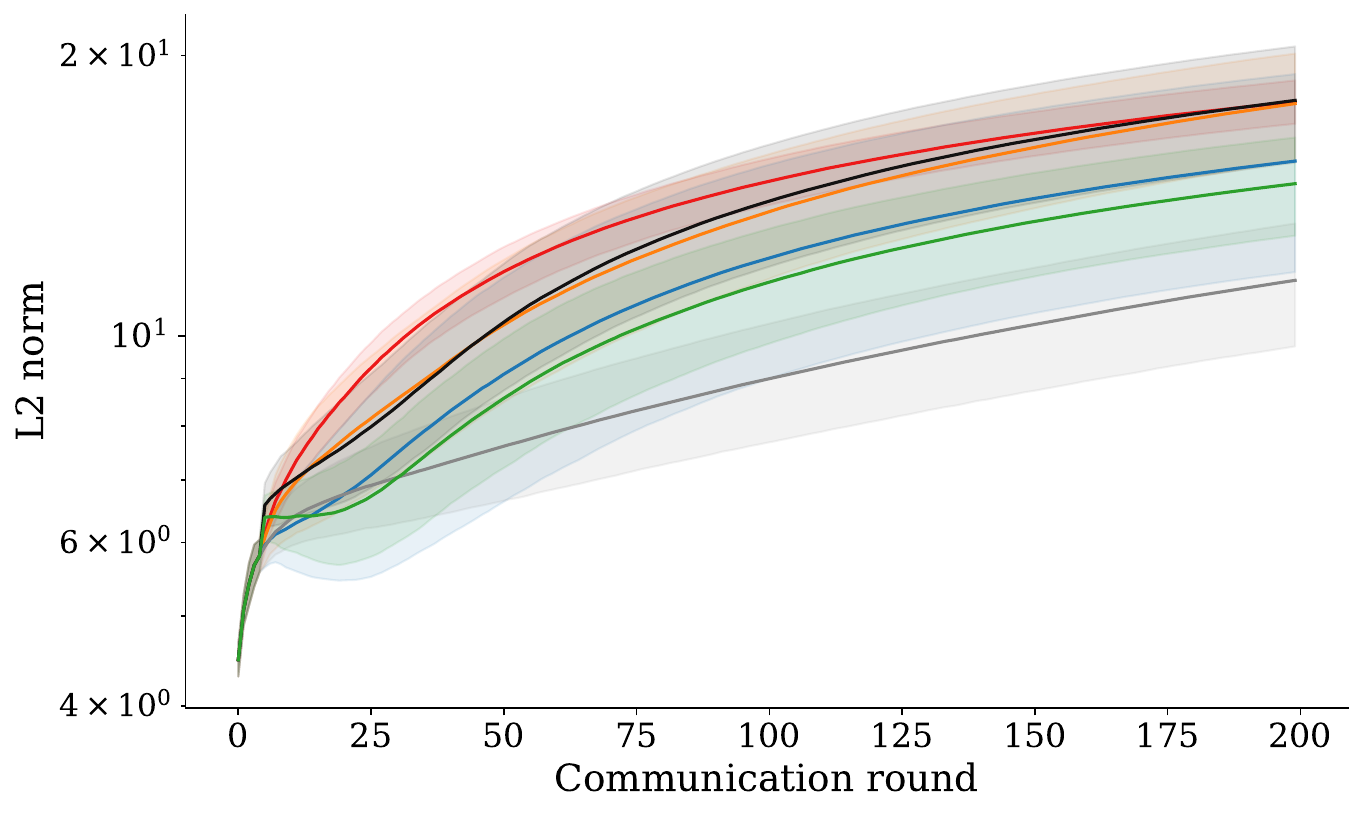}
        \caption{FSR - non-iid (clusters)}
        
    \end{subfigure}
    \hfill
    \begin{subfigure}[b]{0.33\textwidth}
        \includegraphics[width=\linewidth]{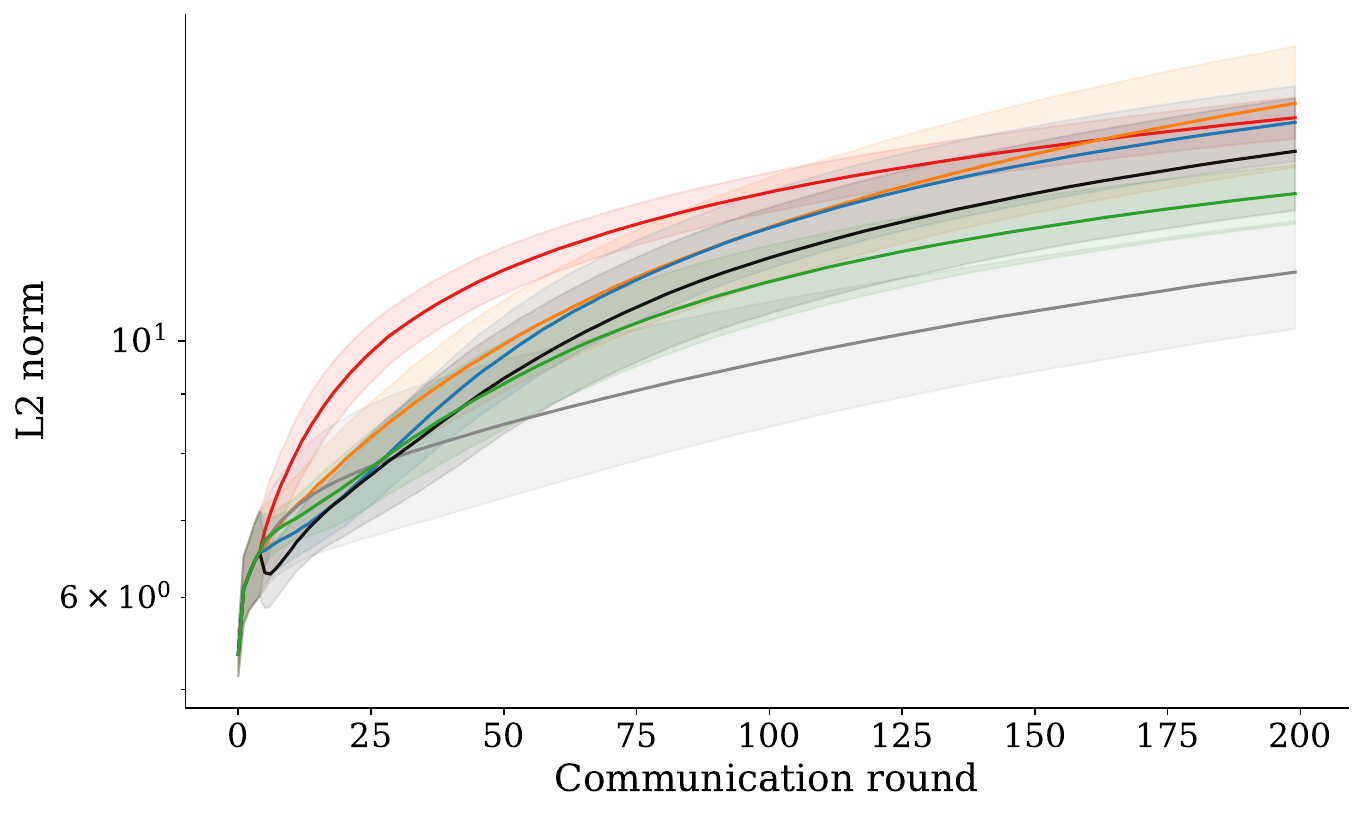}
        \caption{FSR - non-iid (classes)}
       
    \end{subfigure}

    \begin{subfigure}[b]{0.33\textwidth}
        \includegraphics[width=\linewidth]{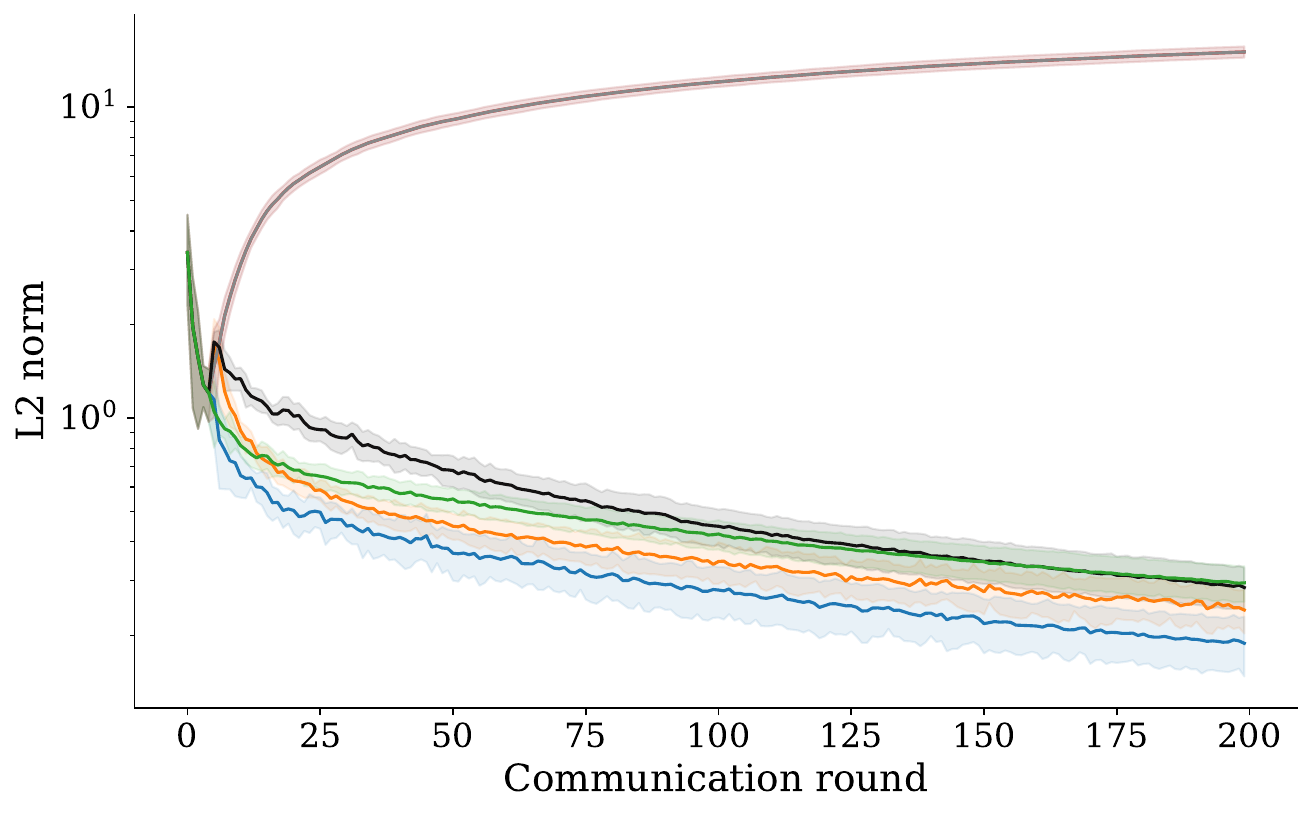}
        \caption{DFedAvgM - iid}
        
    \end{subfigure}
    \hfill
    \begin{subfigure}[b]{0.33\textwidth}
        \includegraphics[width=\linewidth]{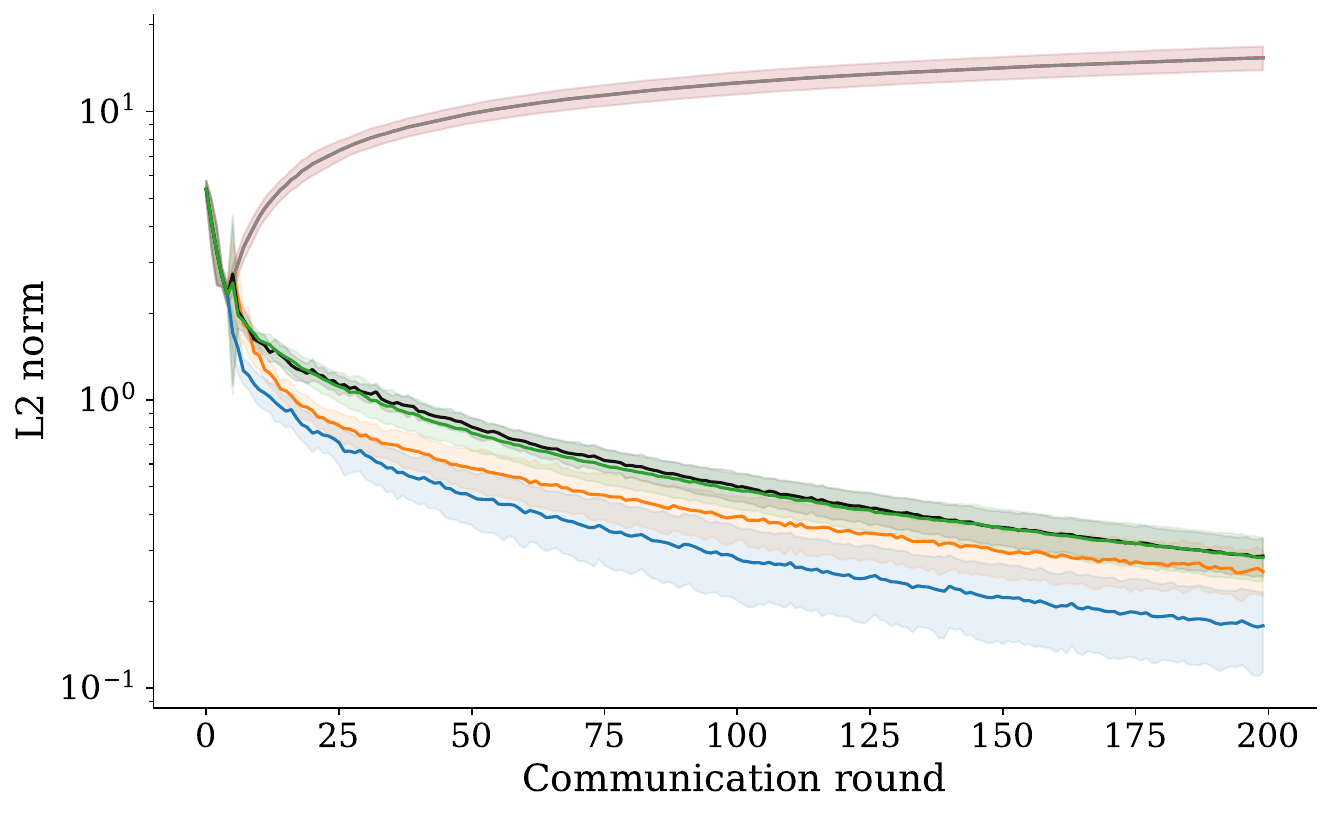}
        \caption{DFedAvgM - non-iid (clusters)}
       
    \end{subfigure}
    \hfill
    \begin{subfigure}[b]{0.33\textwidth}
        \includegraphics[width=\linewidth]{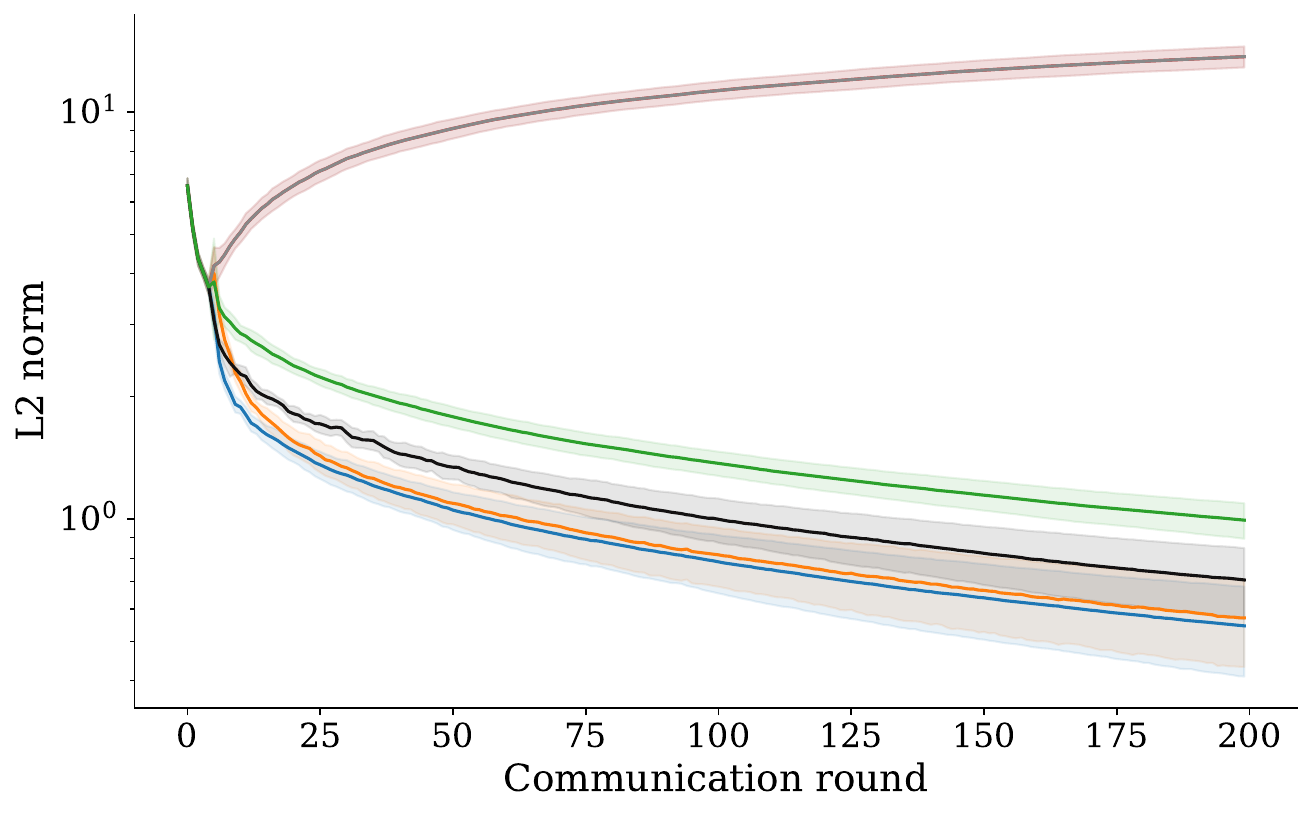}
        \caption{DFedAvgM - non-iid (classes)}
   
    \end{subfigure}

    \caption{Similarity plots for DJAM, FSR, and DFedAvgM algorithms on the digits dataset. Shaded regions represent mean $\pm$ standard deviation across 10 folds, over 200 communication rounds. A random client is dropped persistently after the 5th round. Colors:{\color{modelinversion}model inversion}, {\color{gradientinversion}gradient inversion}, {\color{reference}reference}, {\color{random}random}, {\color{drop}drop}, {\color{noaction}no action}}.
    \label{fig:similarity:digits}
\end{figure*}

\begin{figure*}[htbp]
    \centering
    \begin{subfigure}[b]{0.33\textwidth}
        \includegraphics[width=\linewidth]{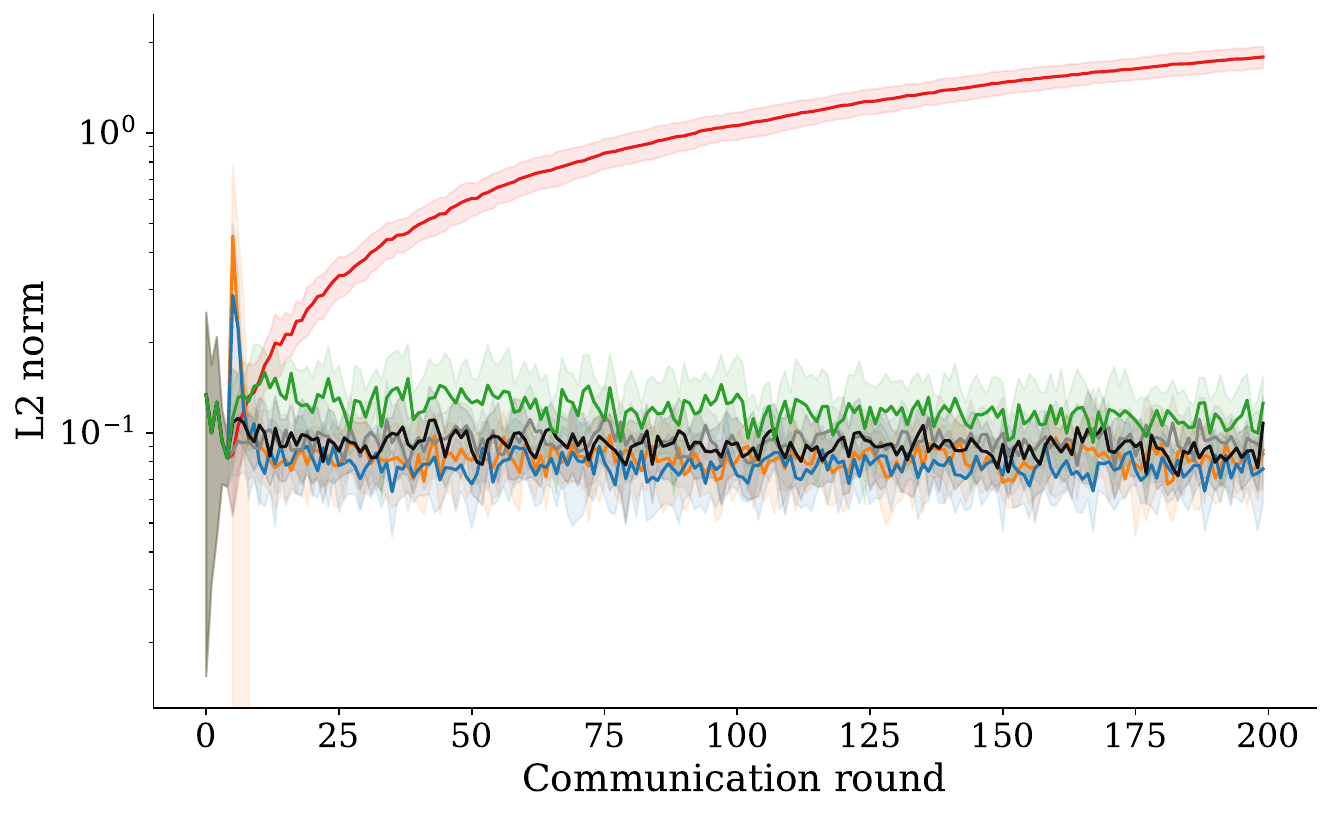}
        \caption{DJAM - iid}
    \end{subfigure}
    \hfill
    \begin{subfigure}[b]{0.33\textwidth}
        \includegraphics[width=\linewidth]{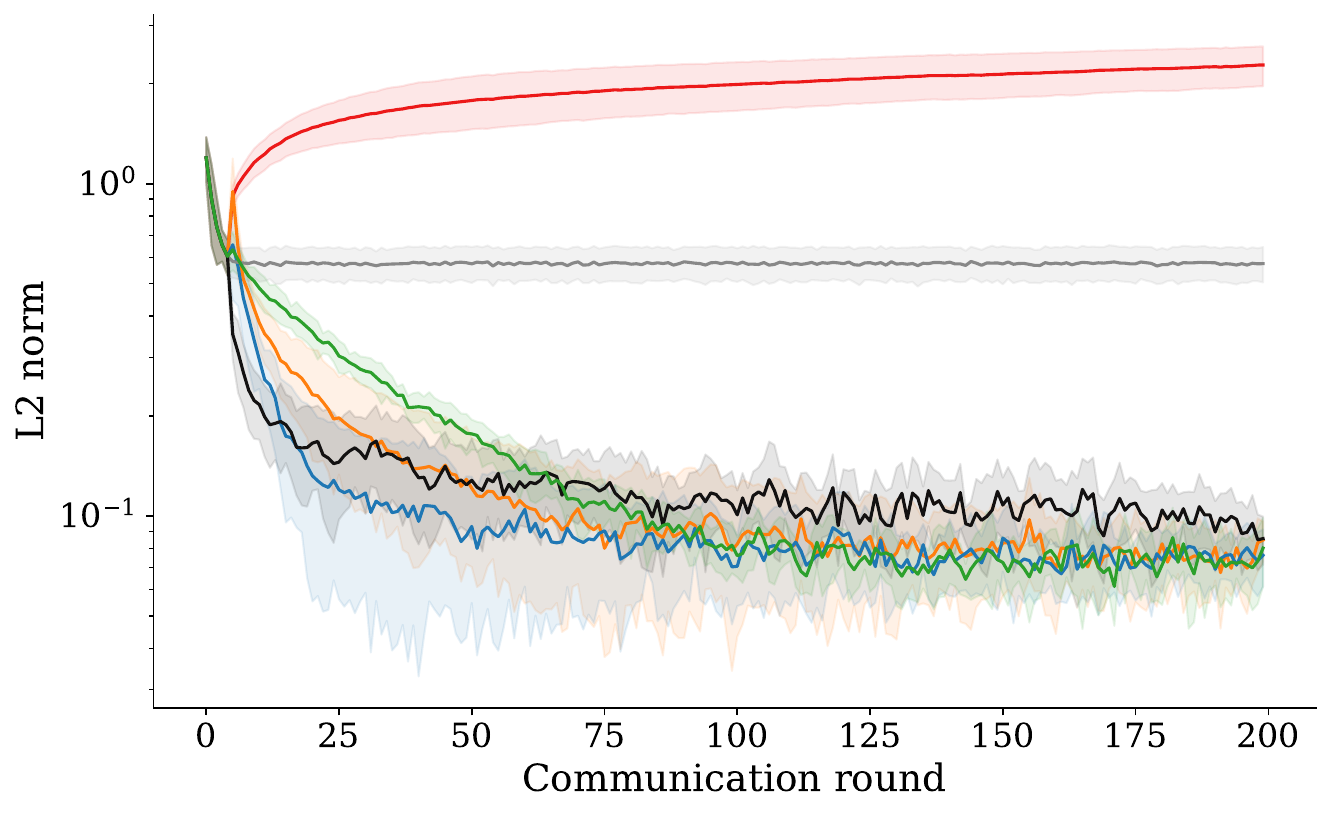}
        \caption{DJAM - non-iid (clusters)}
    \end{subfigure}
    \hfill
    \begin{subfigure}[b]{0.33\textwidth}
        \includegraphics[width=\linewidth]{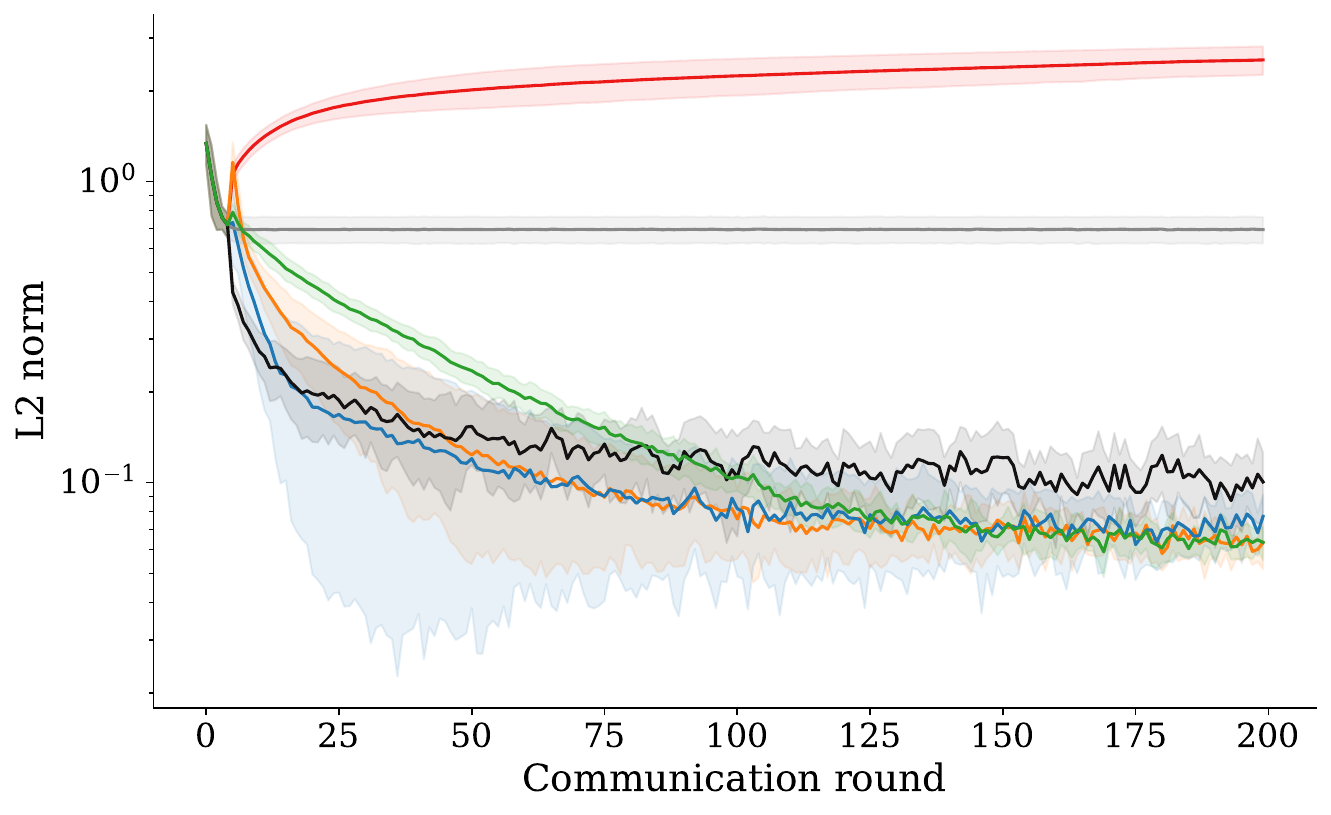}
        \caption{DJAM - non-iid (classes)}
    \end{subfigure}

    \begin{subfigure}[b]{0.33\textwidth}
        \includegraphics[width=\linewidth]{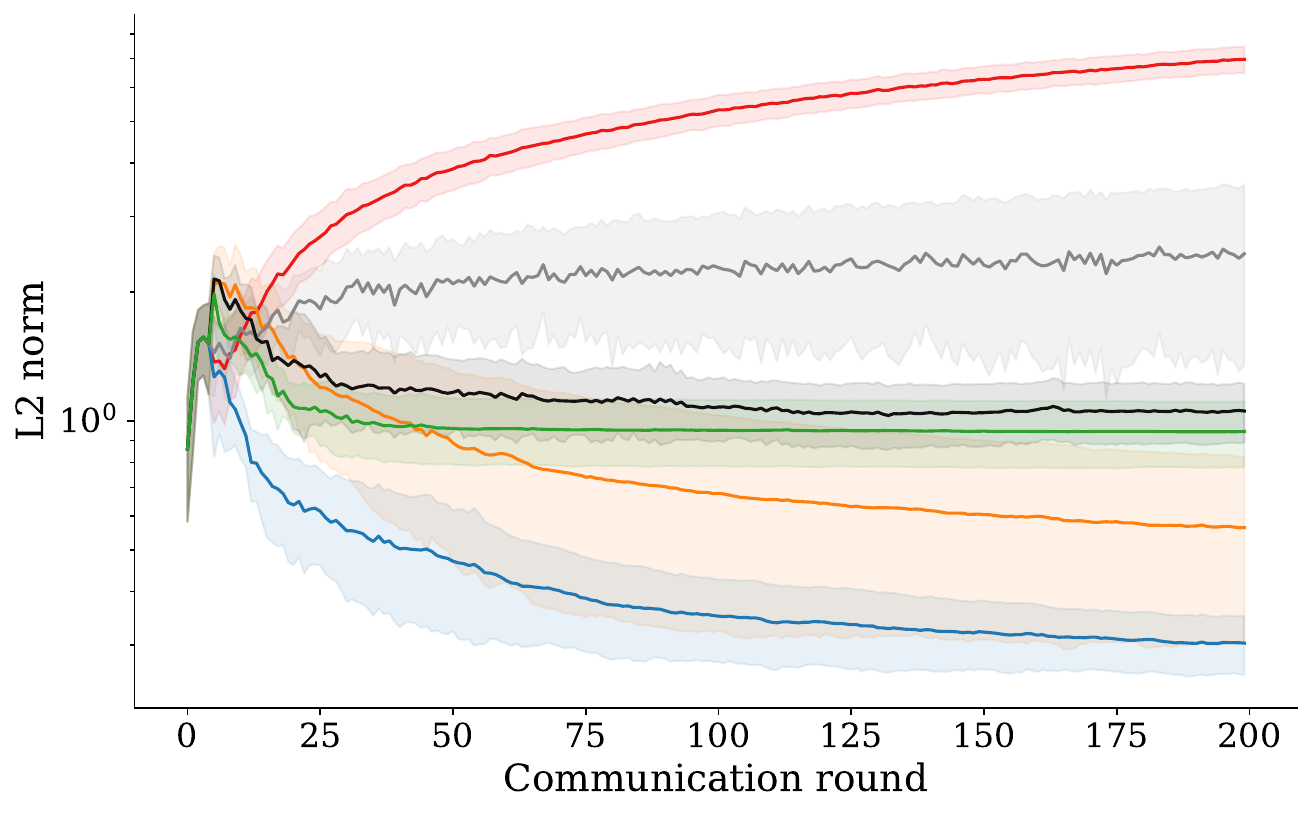}
        \caption{FSR - iid}
    \end{subfigure}
    \hfill
    \begin{subfigure}[b]{0.33\textwidth}
        \includegraphics[width=\linewidth]{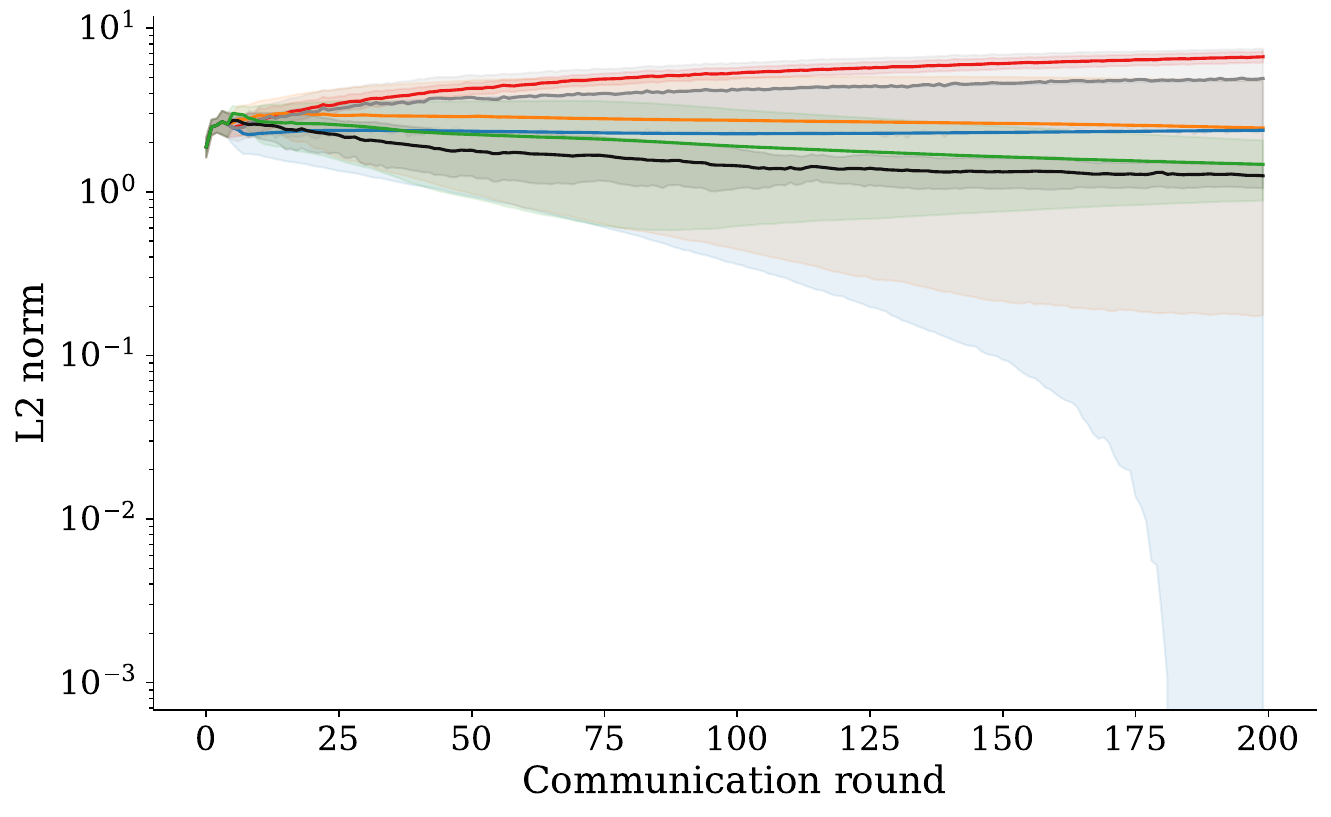}
        \caption{FSR - non-iid (clusters)}
    \end{subfigure}
    \hfill
    \begin{subfigure}[b]{0.33\textwidth}
        \includegraphics[width=\linewidth]{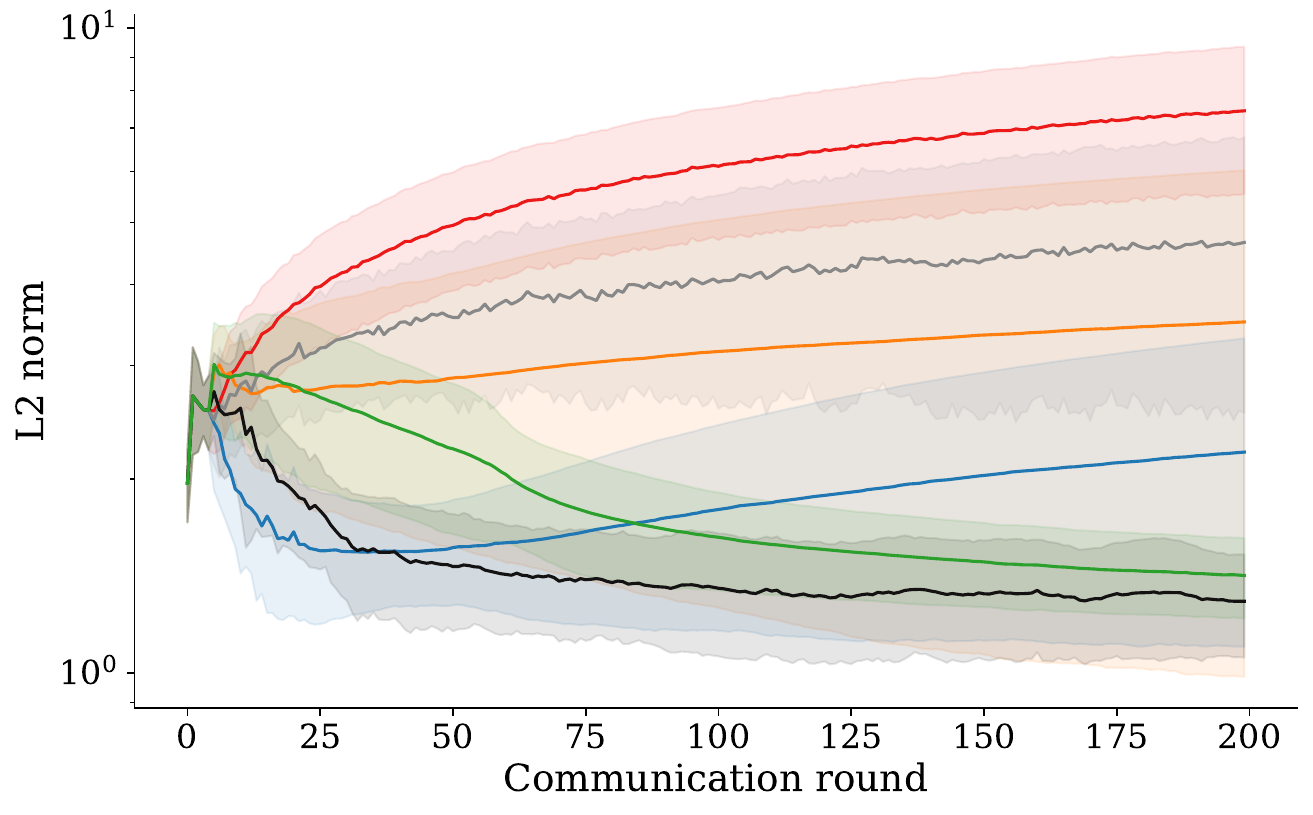}
        \caption{FSR - non-iid (classes)}
    \end{subfigure}

    \begin{subfigure}[b]{0.33\textwidth}
        \includegraphics[width=\linewidth]{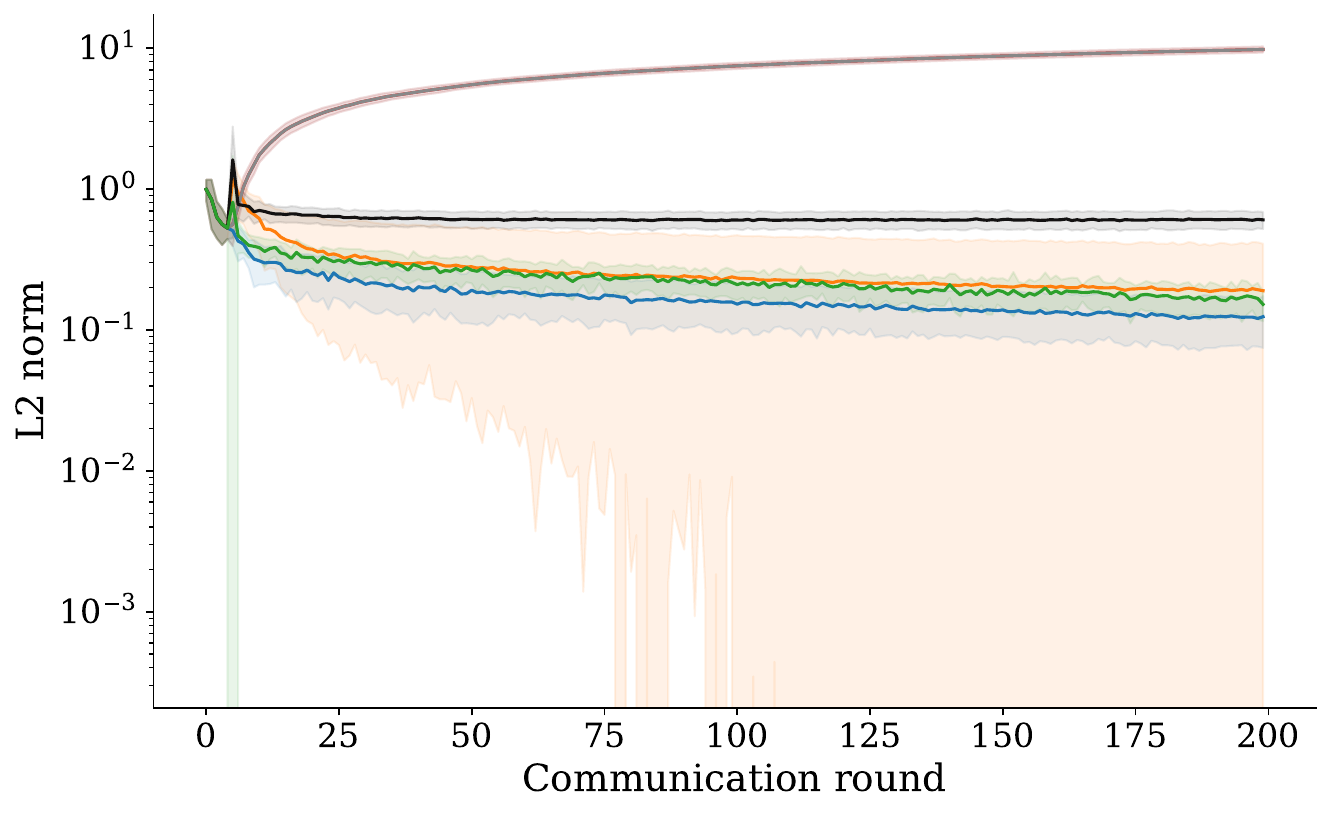}
        \caption{DFedAvgM - iid}
    \end{subfigure}
    \hfill
    \begin{subfigure}[b]{0.33\textwidth}
        \includegraphics[width=\linewidth]{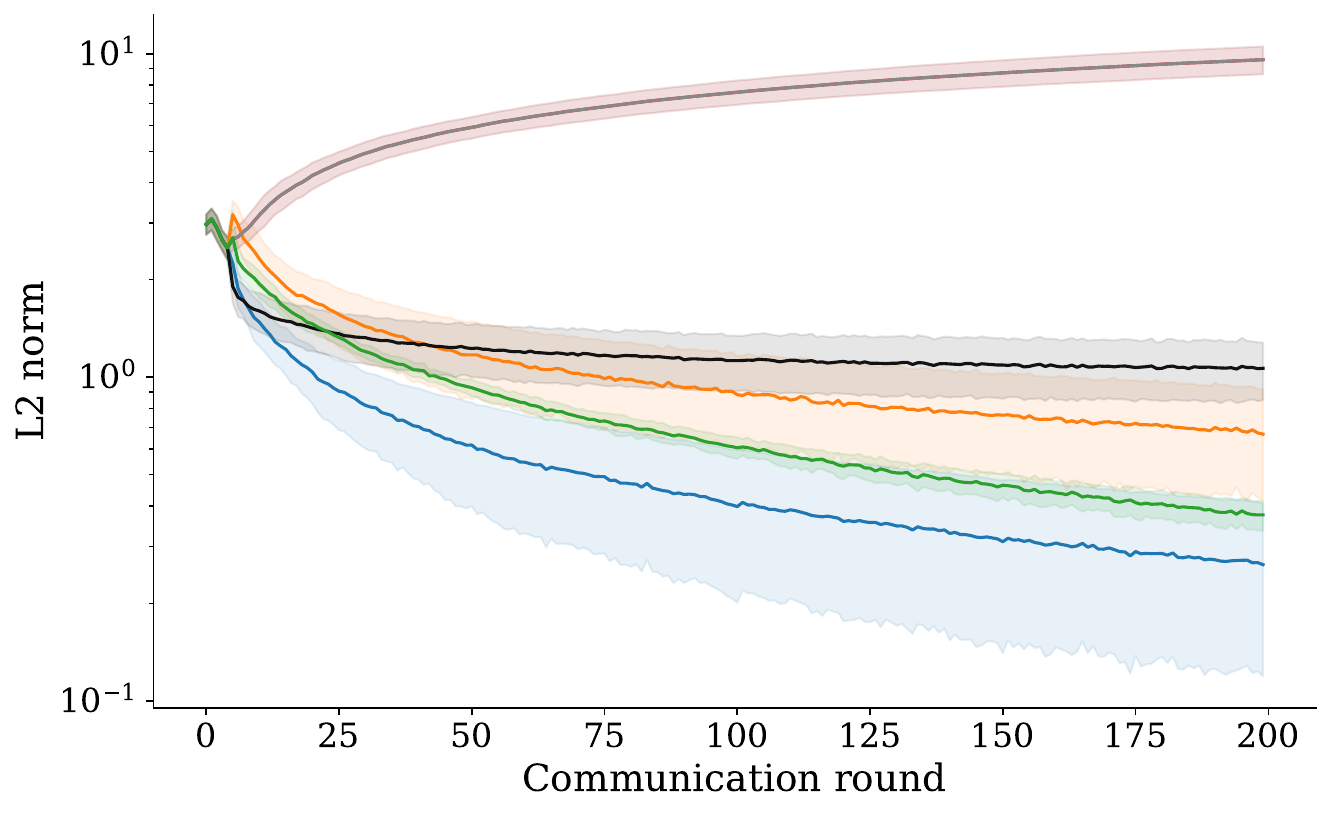}
        \caption{DFedAvgM - non-iid (clusters)}
    \end{subfigure}
    \hfill
    \begin{subfigure}[b]{0.33\textwidth}
        \includegraphics[width=\linewidth]{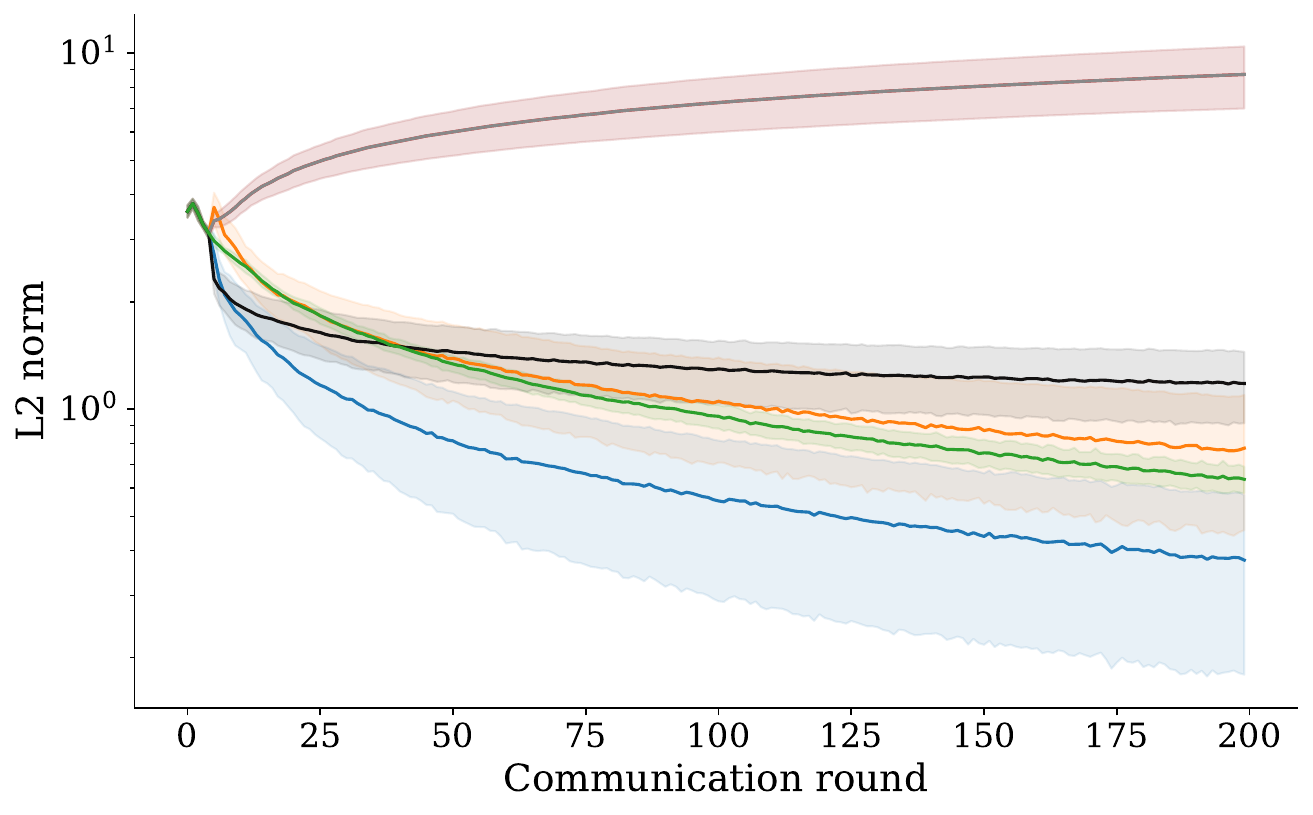}
        \caption{DFedAvgM - non-iid (classes)}
    \end{subfigure}

    \caption{Similarity plots for DJAM, FSR, and DFedAvgM algorithms on the wine dataset. Shaded regions represent mean $\pm$ standard deviation across 10 folds, over 200 communication rounds. A random client is dropped persistently after the 5th round. Colors: {\color{modelinversion}model inversion}, {\color{gradientinversion}gradient inversion}, {\color{reference}reference}, {\color{random}random}, {\color{drop}drop}, {\color{noaction}no action}}.
    \label{fig:similarity:wine}
\end{figure*}

\begin{figure*}[htbp]
    \centering
    \begin{subfigure}[b]{0.33\textwidth}
        \includegraphics[width=\linewidth]{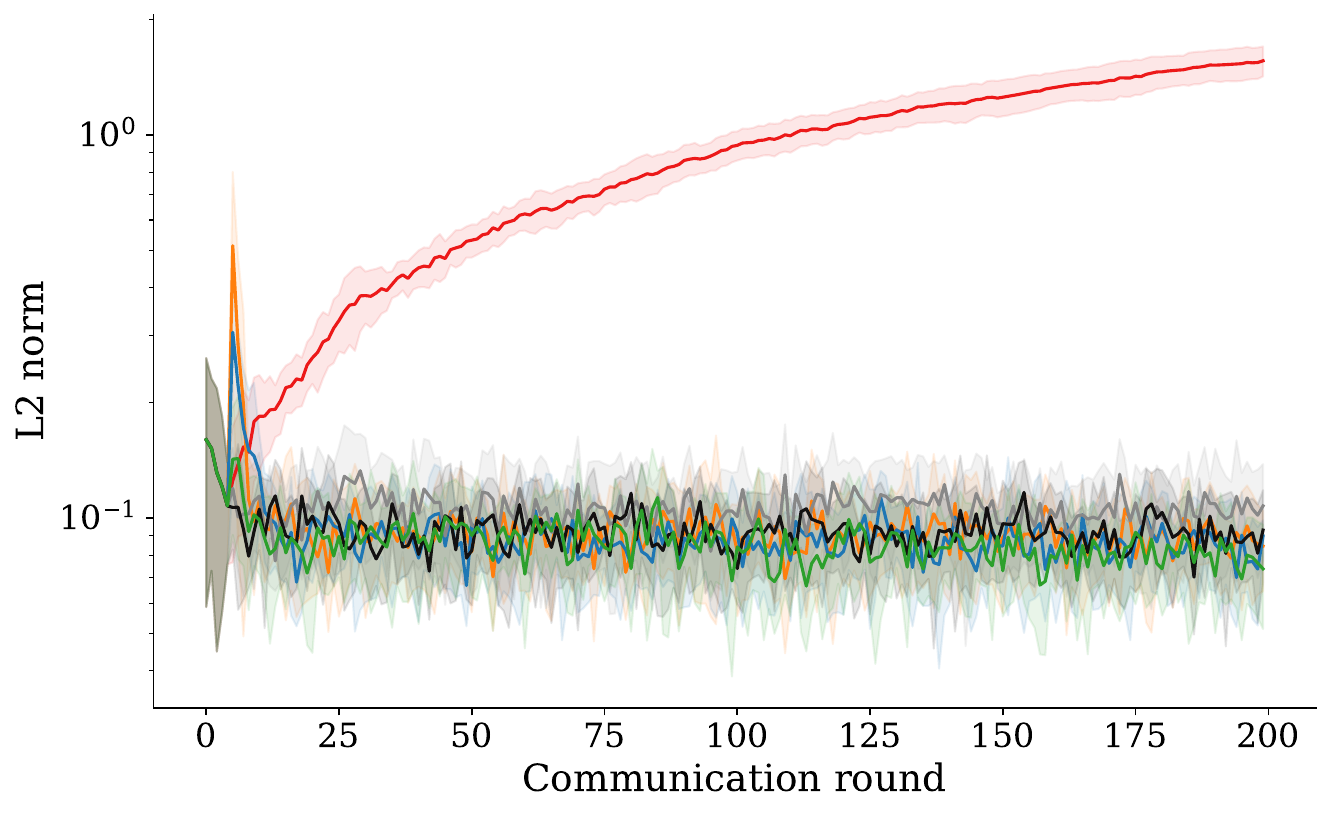}
        \caption{DJAM - iid}
    \end{subfigure}
    \hfill
    \begin{subfigure}[b]{0.33\textwidth}
        \includegraphics[width=\linewidth]{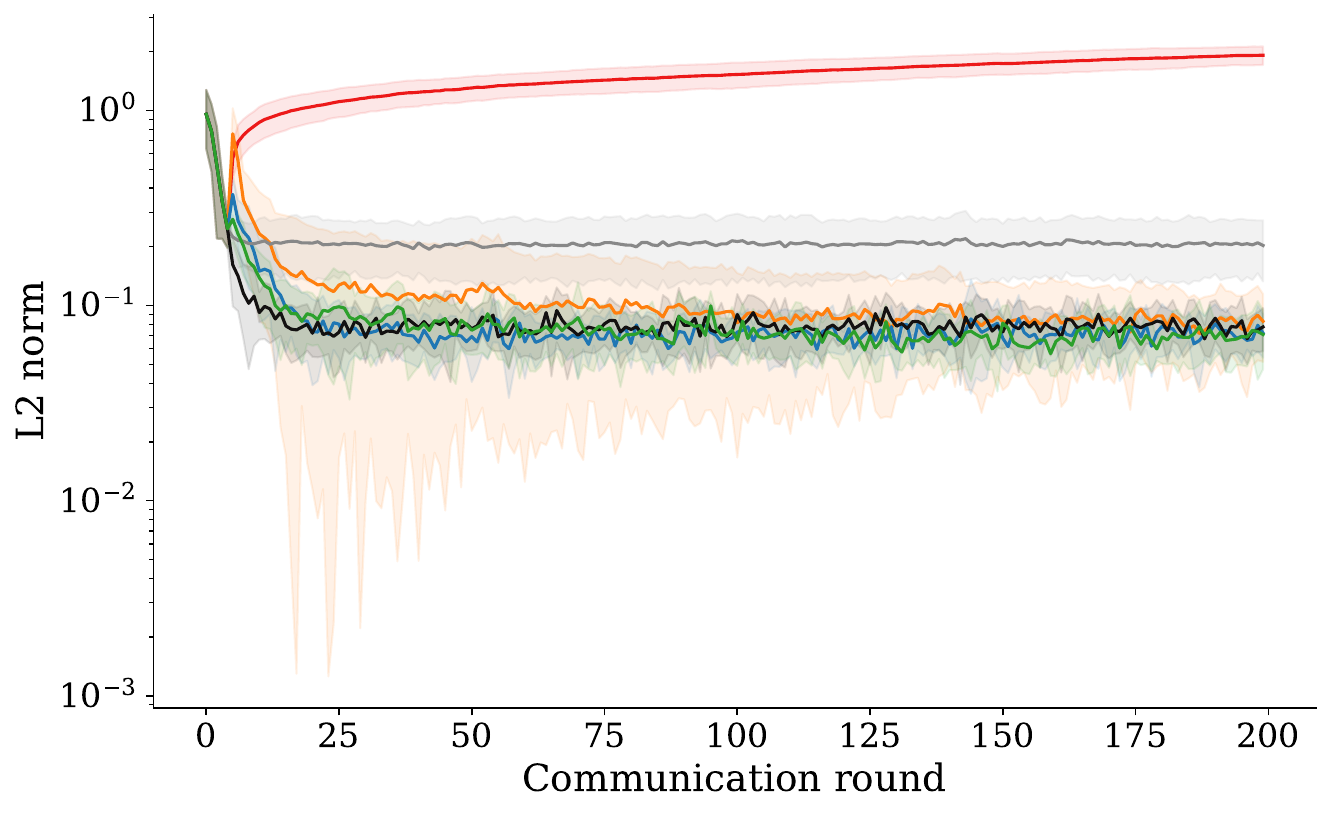}
        \caption{DJAM - non-iid (clusters)}
      
    \end{subfigure}
    \hfill
    \begin{subfigure}[b]{0.33\textwidth}
        \includegraphics[width=\linewidth]{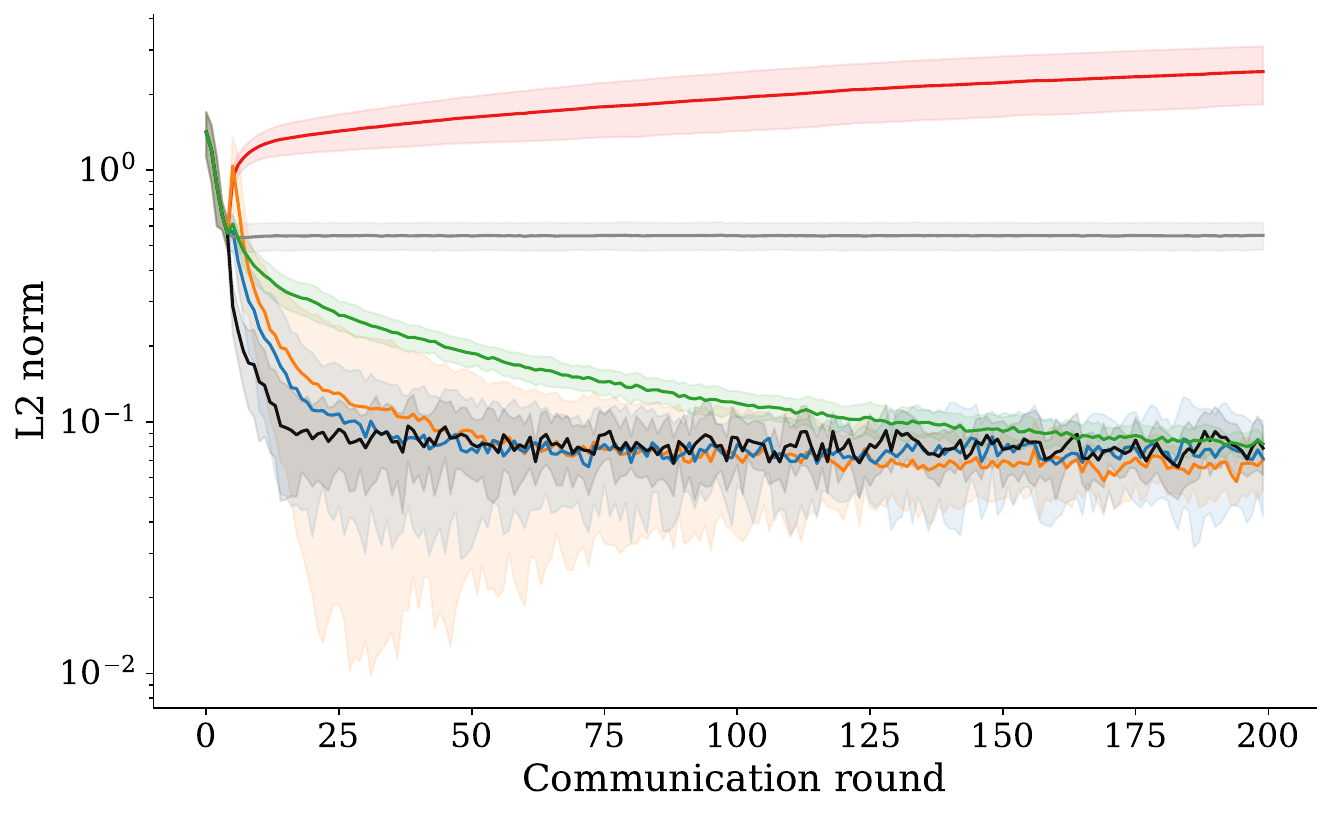}
        \caption{DJAM - non-iid (classes)}
      
    \end{subfigure}

    \begin{subfigure}[b]{0.33\textwidth}
        \includegraphics[width=\linewidth]{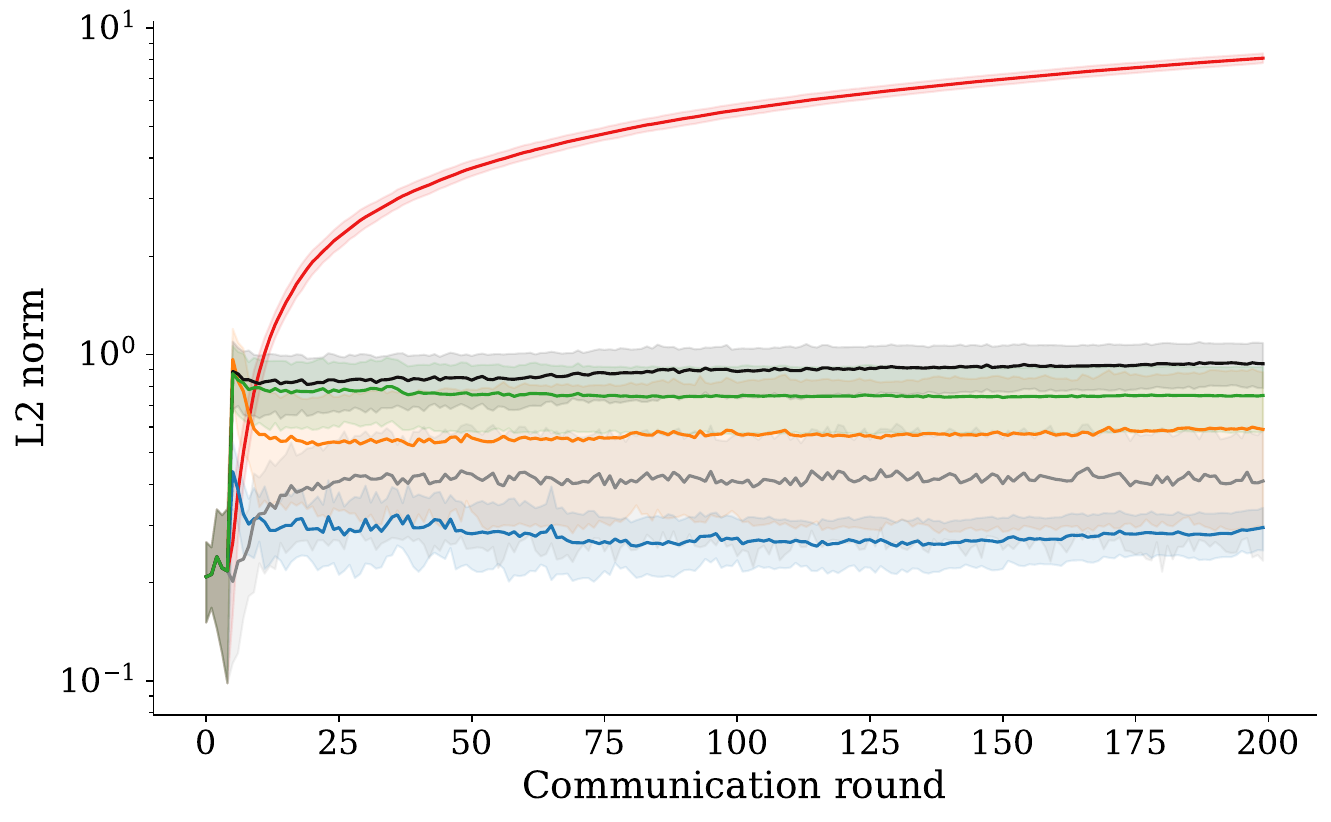}
        \caption{FSR - iid}
     
    \end{subfigure}
    \hfill
    \begin{subfigure}[b]{0.33\textwidth}
        \includegraphics[width=\linewidth]{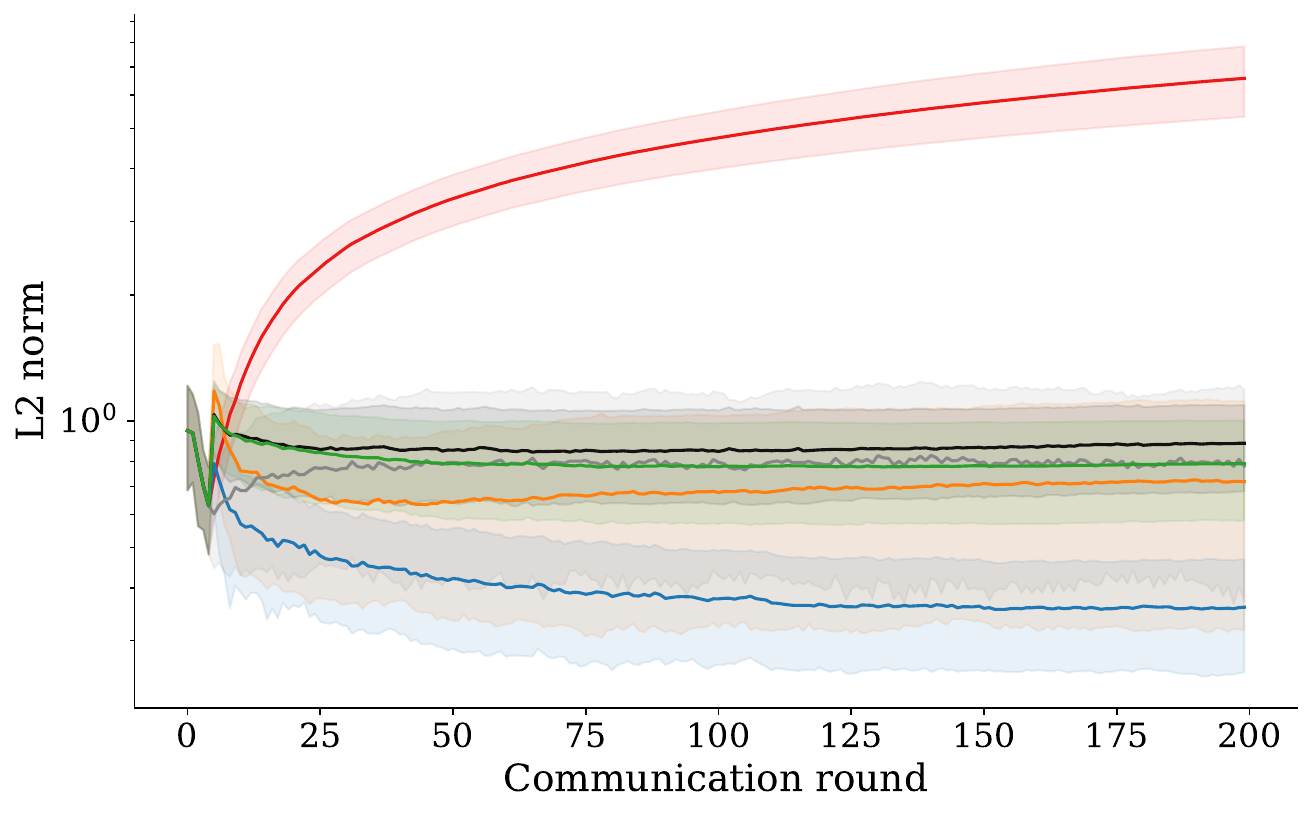}
        \caption{FSR - non-iid (clusters)}
        
    \end{subfigure}
    \hfill
    \begin{subfigure}[b]{0.33\textwidth}
        \includegraphics[width=\linewidth]{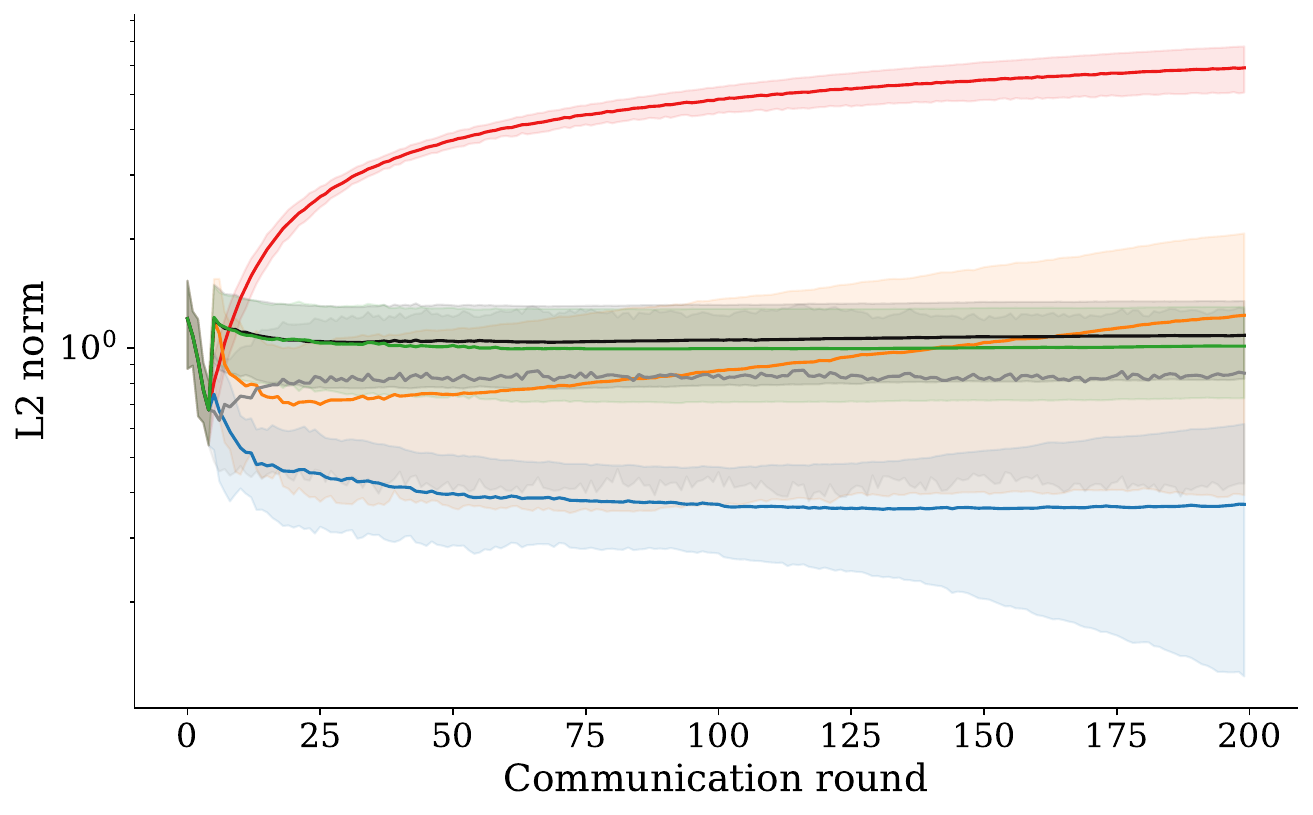}
        \caption{FSR - non-iid (classes)}
     
    \end{subfigure}

    \begin{subfigure}[b]{0.33\textwidth}
        \includegraphics[width=\linewidth]{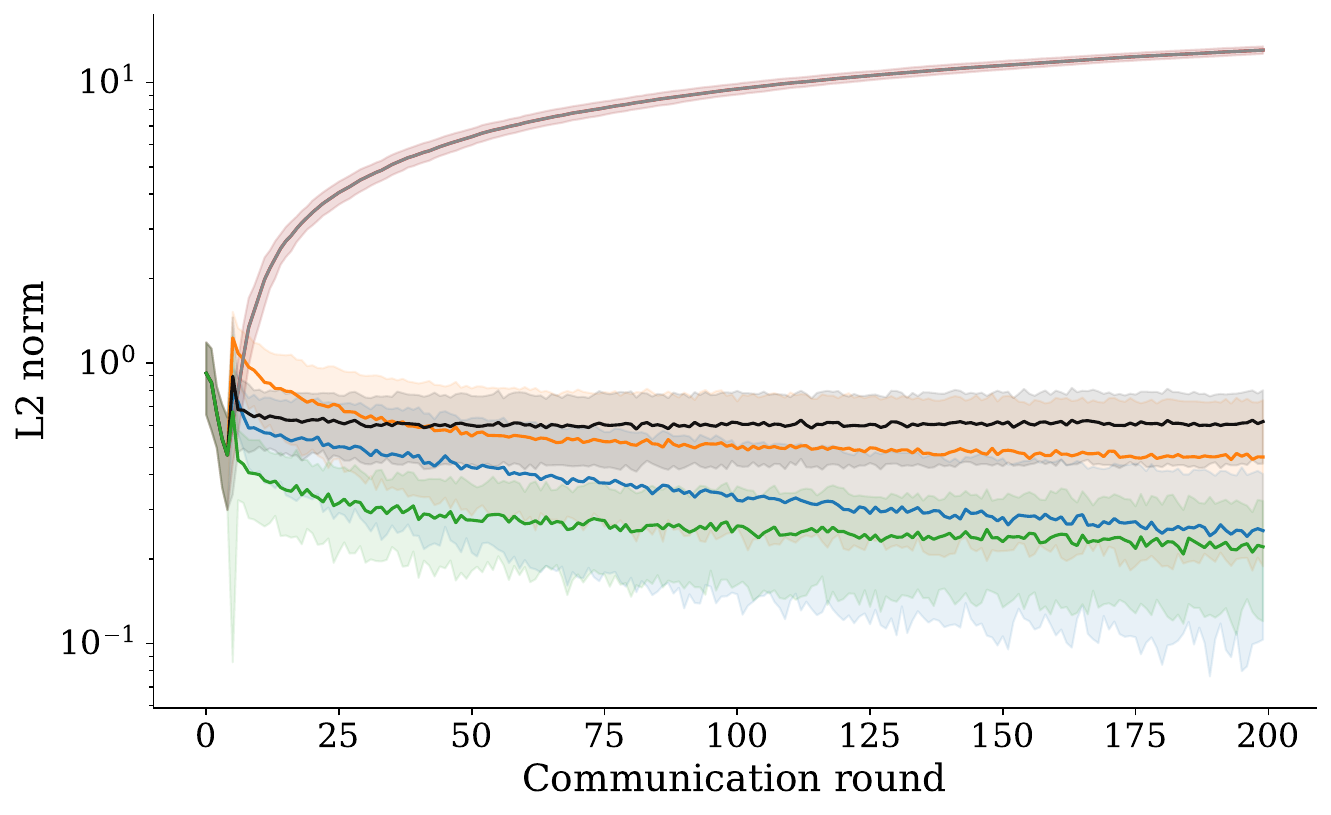}
        \caption{DFedAvgM - iid}
      
    \end{subfigure}
    \hfill
    \begin{subfigure}[b]{0.33\textwidth}
        \includegraphics[width=\linewidth]{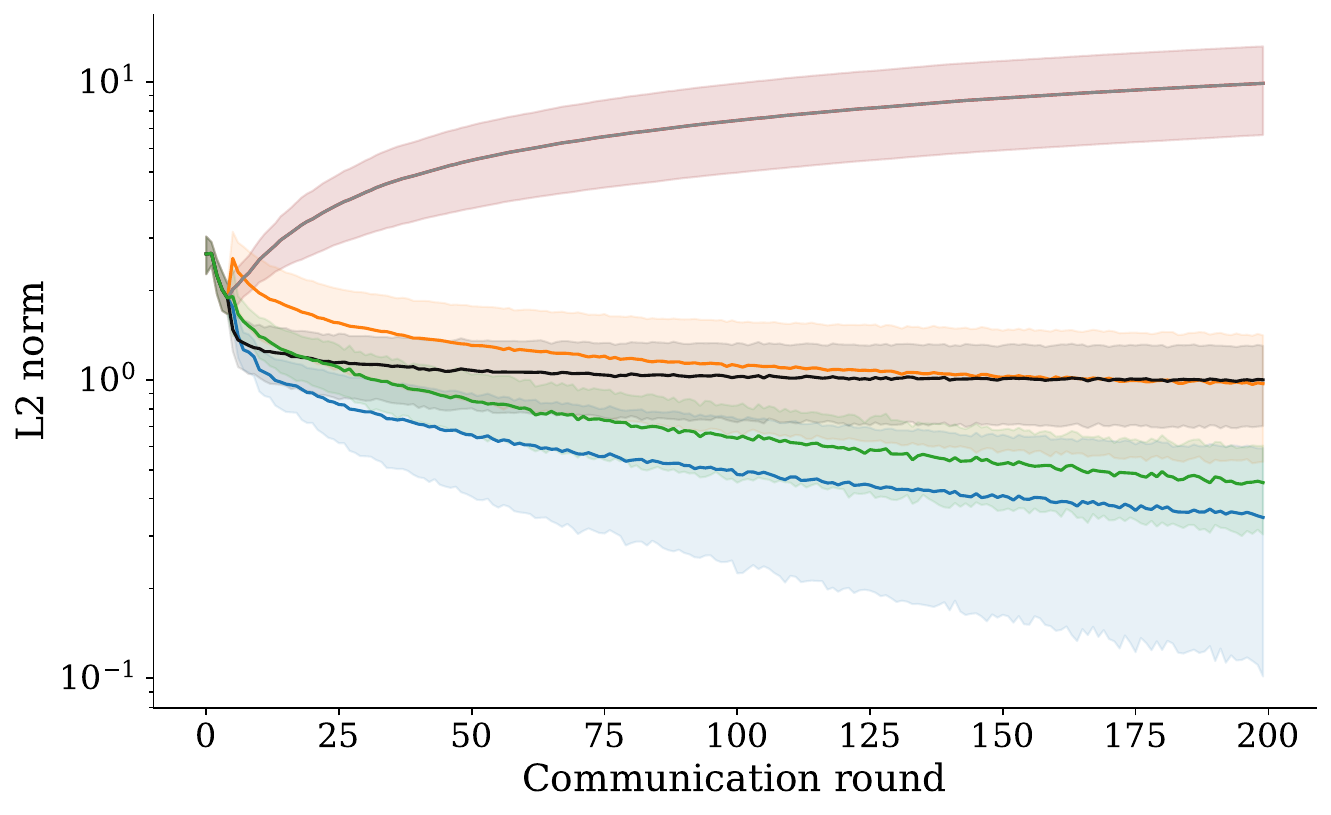}
        \caption{DFedAvgM - non-iid (clusters)}
       
    \end{subfigure}
    \hfill
    \begin{subfigure}[b]{0.33\textwidth}
        \includegraphics[width=\linewidth]{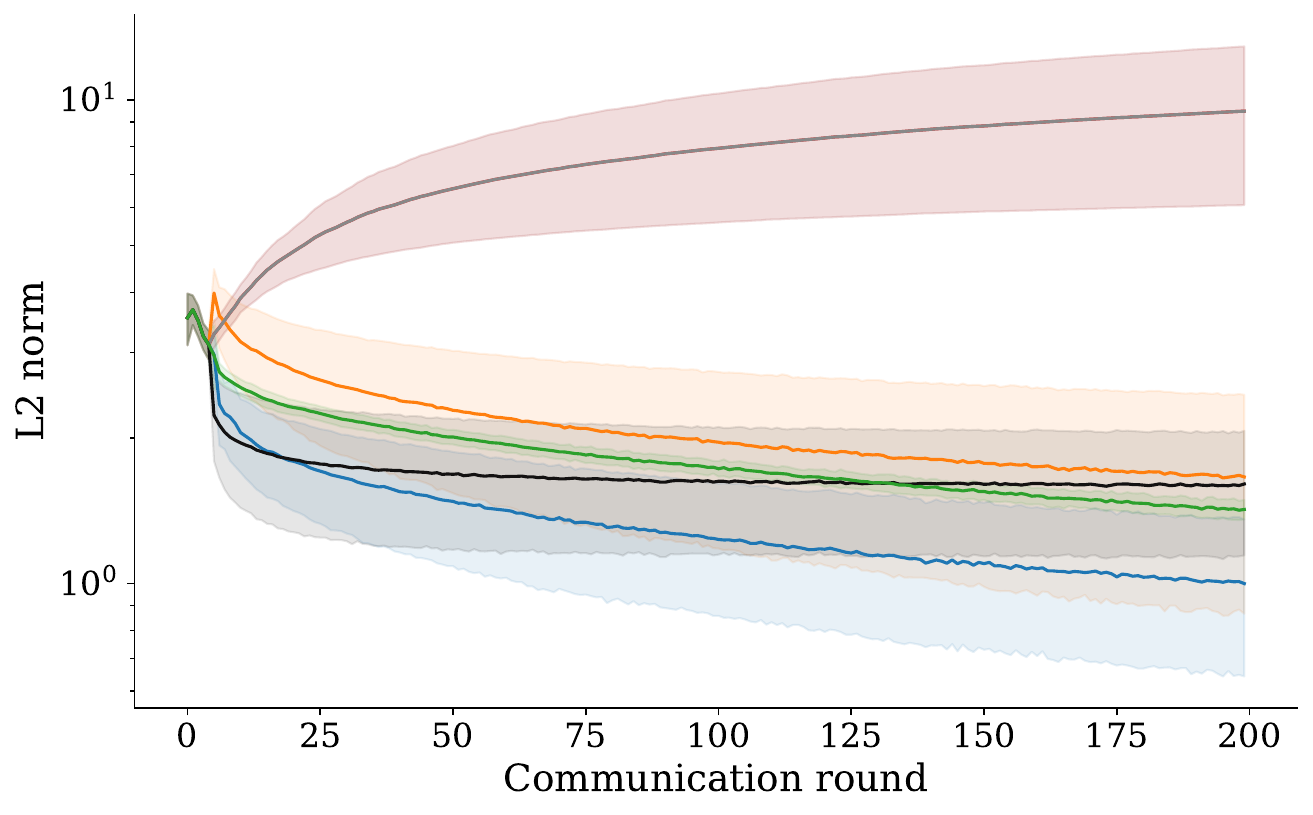}
        \caption{DFedAvgM - non-iid (classes)}
       
    \end{subfigure}

    \caption{Similarity plots for DJAM, FSR, and DFedAvgM algorithms on the iris dataset. Shaded regions represent mean $\pm$ standard deviation across 10 folds, over 200 communication rounds. A random client is dropped persistently after the 5th round. Colors: {\color{modelinversion}model inversion}, {\color{gradientinversion}gradient inversion}, {\color{reference}reference}, {\color{random}random}, {\color{drop}drop}, {\color{noaction}no action}}.
    \label{fig:similarity:iris}
\end{figure*}

This section presents model similarity plots corresponding to the experiments and convergence plots shown in \Cref{fig:similarity:digits,fig:similarity:iris,fig:similarity:wine} and detailed in \Cref{app:exp:all-conv-plots,app:exp:all-tables}. 

A key observation is that for the FSR algorithm, models do not converge in parameter space during optimization. This is not a surprise, as FSR explicitly optimizes for similarity in function space rather than parameter space. Aside from this, the similarity trends across experiments align with our findings in the main text and provide further evidence that the proposed adaptive strategies lead to more cohesive and aligned models across the federation.

%% file: chapters/7.5.appendix.tex
\section{Results with Neural Networks}
\label{app:sec:neural-network}

\begin{figure*}[htbp]
    \centering
    \begin{subfigure}[b]{0.33\textwidth}
        \includegraphics[width=\linewidth]{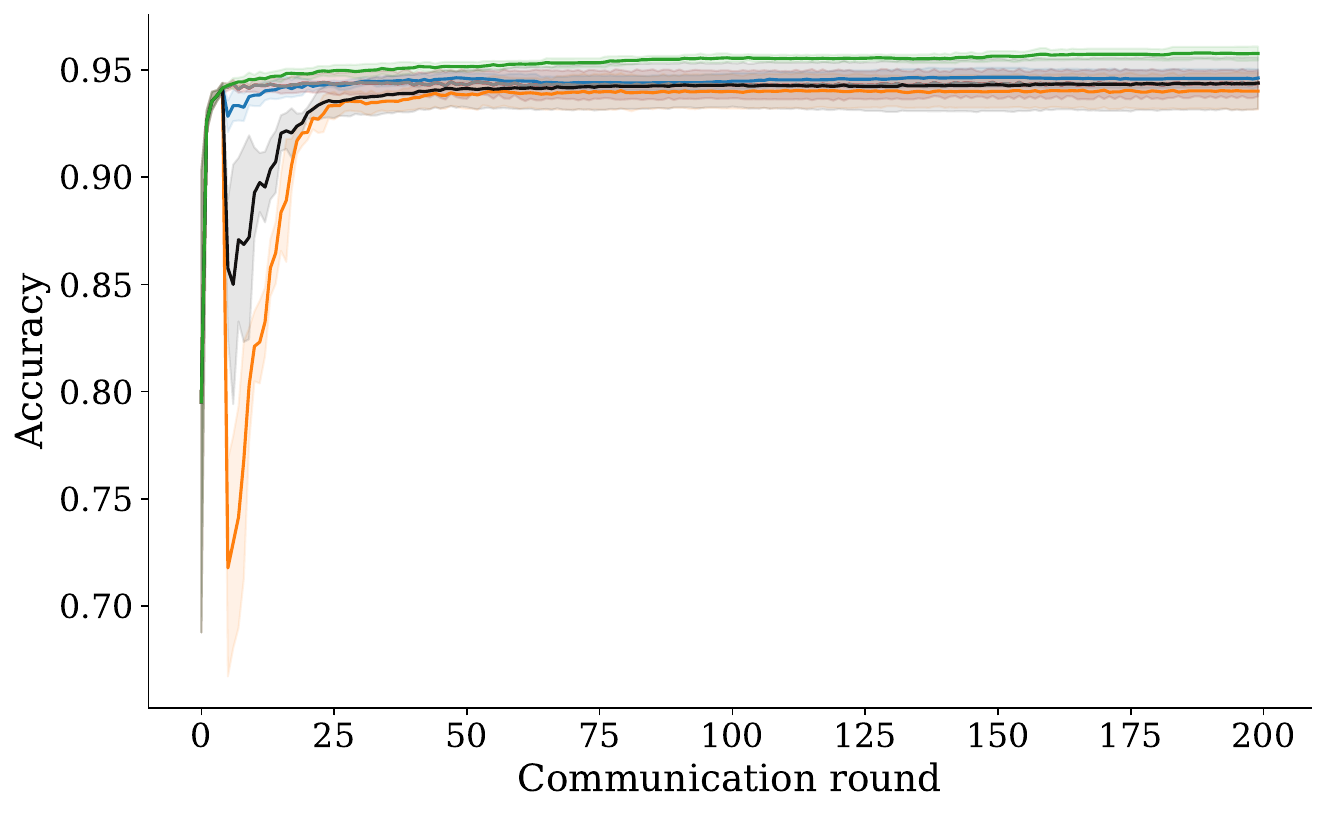}
        \caption{DFedAvgM - iid}
       
    \end{subfigure}
    \hfill
    \begin{subfigure}[b]{0.33\textwidth}
        \includegraphics[width=\linewidth]{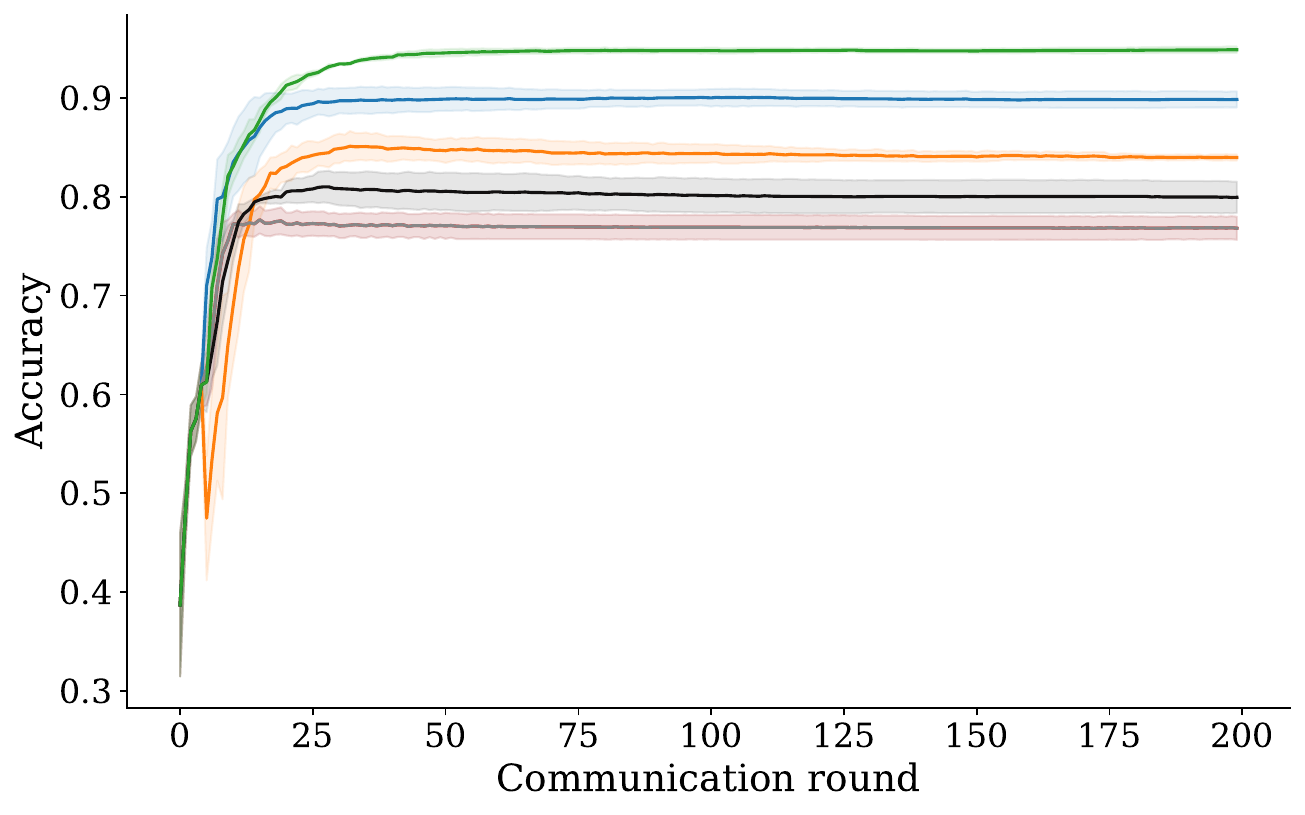}
        \caption{DFedAvgM}
     
    \end{subfigure}
    \hfill
    \begin{subfigure}[b]{0.33\textwidth}
        \includegraphics[width=\linewidth]{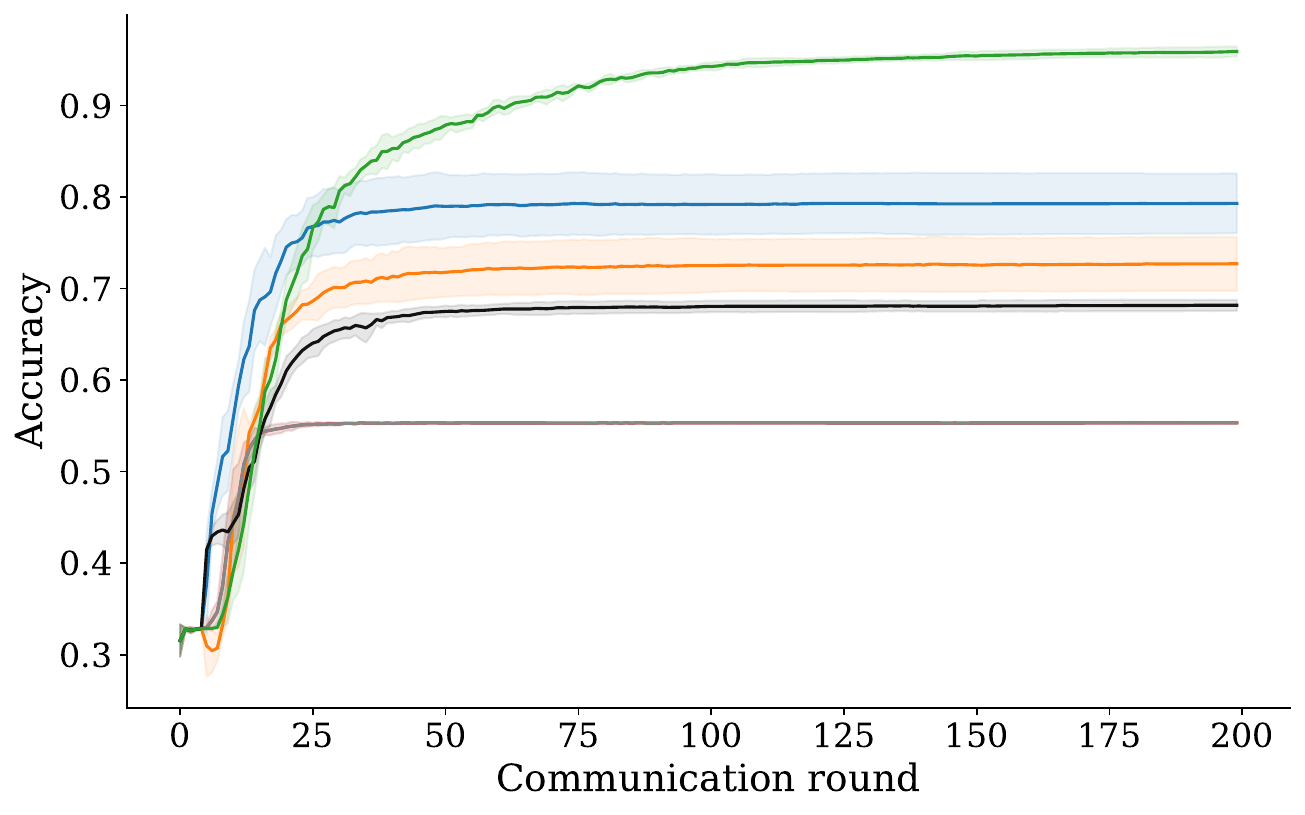}
        \caption{DFedAvgM - non-iid (classes)}
       
    \end{subfigure}

    \caption{Convergence plots for DFedAvgM on the digits dataset using a neural network model (3-layer MLP with 128 hidden units per layer). Shaded regions represent mean $\pm$ standard deviation across 3 folds, over 200 communication rounds. A random client is dropped persistently after the 5th round. Colors: {\color{modelinversion}model inversion}, {\color{gradientinversion}gradient inversion}, {\color{reference}reference}, {\color{random}random}, {\color{drop}drop}, {\color{noaction}no action}.}
    
    \label{fig:appendix:extra:nn}
\end{figure*}

This section presents a smaller set of experiments where the underlying model is a neural network instead of logistic regression. In \Cref{fig:appendix:extra:nn}, we show results on the digits dataset using the DFedAvgM algorithm with a 3-layer multilayer perceptron (MLP), where each hidden layer has 128 units.

The key takeaway is that performance trends are consistent with those observed for simpler models. Adaptive strategies, particularly model and gradient inversion, remain effective even when using a neural network as the base model. 

%% file: paper.bbl

\begin{thebibliography}{17}


\ifx \showCODEN    \undefined \def \showCODEN     #1{\unskip}     \fi
\ifx \showISBNx    \undefined \def \showISBNx     #1{\unskip}     \fi
\ifx \showISBNxiii \undefined \def \showISBNxiii  #1{\unskip}     \fi
\ifx \showISSN     \undefined \def \showISSN      #1{\unskip}     \fi
\ifx \showLCCN     \undefined \def \showLCCN      #1{\unskip}     \fi
\ifx \shownote     \undefined \def \shownote      #1{#1}          \fi
\ifx \showarticletitle \undefined \def \showarticletitle #1{#1}   \fi
\ifx \showURL      \undefined \def \showURL       {\relax}        \fi
\providecommand\bibfield[2]{#2}
\providecommand\bibinfo[2]{#2}
\providecommand\natexlab[1]{#1}
\providecommand\showeprint[2][]{arXiv:#2}

\bibitem[Almeida and Xavier(2018)]%
        {almeida_djam_2018}
\bibfield{author}{\bibinfo{person}{Inês Almeida} {and} \bibinfo{person}{João Xavier}.} \bibinfo{year}{2018}\natexlab{}.
\newblock \showarticletitle{{DJAM}: {Distributed} {Jacobi} {Asynchronous} {Method} for {Learning} {Personal} {Models}}.
\newblock \bibinfo{journal}{\emph{IEEE Signal Process. Lett.}} \bibinfo{volume}{25}, \bibinfo{number}{9} (\bibinfo{year}{2018}), \bibinfo{pages}{1389--1392}.
\newblock
\showISSN{1558-2361}


\bibitem[Fang et~al\mbox{.}(2024)]%
        {fang2024privacy}
\bibfield{author}{\bibinfo{person}{Hao Fang}, \bibinfo{person}{Yixiang Qiu}, \bibinfo{person}{Hongyao Yu}, \bibinfo{person}{Wenbo Yu}, \bibinfo{person}{Jiawei Kong}, \bibinfo{person}{Baoli Chong}, \bibinfo{person}{Bin Chen}, \bibinfo{person}{Xuan Wang}, \bibinfo{person}{Shu-Tao Xia}, {and} \bibinfo{person}{Ke Xu}.} \bibinfo{year}{2024}\natexlab{}.
\newblock \showarticletitle{Privacy leakage on dnns: A survey of model inversion attacks and defenses}.
\newblock \bibinfo{journal}{\emph{arXiv preprint arXiv:2402.04013}} (\bibinfo{year}{2024}).
\newblock


\bibitem[Geiping et~al\mbox{.}(2020)]%
        {geiping_inverting_2020}
\bibfield{author}{\bibinfo{person}{Jonas Geiping}, \bibinfo{person}{Hartmut Bauermeister}, \bibinfo{person}{Hannah Dröge}, {and} \bibinfo{person}{Michael Moeller}.} \bibinfo{year}{2020}\natexlab{}.
\newblock \showarticletitle{Inverting {Gradients} -- {How} easy is it to break privacy in federated learning?}
\newblock \bibinfo{journal}{\emph{Advances in Neural Information Processing Systems}}  \bibinfo{volume}{33} (\bibinfo{year}{2020}), \bibinfo{pages}{16937--16937}.
\newblock


\bibitem[Good(2024)]%
        {good2024trustworthy}
\bibfield{author}{\bibinfo{person}{Jack Good}.} \bibinfo{year}{2024}\natexlab{}.
\newblock \emph{\bibinfo{title}{Trustworthy Learning using Uncertain Interpretation of Data}}.
\newblock \bibinfo{thesistype}{Ph.\,D. Dissertation}. \bibinfo{school}{Carnegie Mellon University}.
\newblock


\bibitem[Haim et~al\mbox{.}(2022)]%
        {haim_reconstructing_2022}
\bibfield{author}{\bibinfo{person}{Niv Haim}, \bibinfo{person}{Gal Vardi}, \bibinfo{person}{Gilad Yehudai}, \bibinfo{person}{Ohad Shamir}, {and} \bibinfo{person}{Michal Irani}.} \bibinfo{year}{2022}\natexlab{}.
\newblock \showarticletitle{Reconstructing {Training} {Data} {From} {Trained} {Neural} {Networks}}.
\newblock \bibinfo{journal}{\emph{Advances in Neural Information Processing Systems}}  \bibinfo{volume}{35} (\bibinfo{date}{Dec.} \bibinfo{year}{2022}), \bibinfo{pages}{22911--22924}.
\newblock


\bibitem[Ji and Telgarsky(2020)]%
        {ji_directional_2020}
\bibfield{author}{\bibinfo{person}{Ziwei Ji} {and} \bibinfo{person}{Matus Telgarsky}.} \bibinfo{year}{2020}\natexlab{}.
\newblock \showarticletitle{Directional convergence and alignment in deep learning}.
\newblock \bibinfo{journal}{\emph{arXiv preprint arXiv:2006.06657}} (\bibinfo{year}{2020}).
\newblock


\bibitem[McMahan et~al\mbox{.}(2017)]%
        {mcmahan_communication-efficient_2023}
\bibfield{author}{\bibinfo{person}{Brendan McMahan}, \bibinfo{person}{Eider Moore}, \bibinfo{person}{Daniel Ramage}, \bibinfo{person}{Seth Hampson}, {and} \bibinfo{person}{Blaise~Aguera y Arcas}.} \bibinfo{year}{2017}\natexlab{}.
\newblock \showarticletitle{Communication-efficient learning of deep networks from decentralized data}. In \bibinfo{booktitle}{\emph{Artificial intelligence and statistics}}.
\newblock


\bibitem[Reddi et~al\mbox{.}(2016)]%
        {reddi_aide_2016}
\bibfield{author}{\bibinfo{person}{Sashank~J Reddi}, \bibinfo{person}{Jakub Kone{\v{c}}n{\`y}}, \bibinfo{person}{Peter Richt{\'a}rik}, \bibinfo{person}{Barnab{\'a}s P{\'o}cz{\'o}s}, {and} \bibinfo{person}{Alex Smola}.} \bibinfo{year}{2016}\natexlab{}.
\newblock \showarticletitle{Aide: Fast and communication efficient distributed optimization}.
\newblock \bibinfo{journal}{\emph{arXiv preprint arXiv:1608.06879}} (\bibinfo{year}{2016}).
\newblock


\bibitem[Shao et~al\mbox{.}(2022)]%
        {shao_dres-fl_2022}
\bibfield{author}{\bibinfo{person}{Jiawei Shao}, \bibinfo{person}{Yuchang Sun}, \bibinfo{person}{Songze Li}, {and} \bibinfo{person}{Jun Zhang}.} \bibinfo{year}{2022}\natexlab{}.
\newblock \showarticletitle{Dres-fl: Dropout-resilient secure federated learning for non-iid clients via secret data sharing}.
\newblock \bibinfo{journal}{\emph{Advances in Neural Information Processing Systems}}  \bibinfo{volume}{35} (\bibinfo{year}{2022}), \bibinfo{pages}{10533--10545}.
\newblock


\bibitem[Sun et~al\mbox{.}(2022)]%
        {sun_decentralized_2021}
\bibfield{author}{\bibinfo{person}{Tao Sun}, \bibinfo{person}{Dongsheng Li}, {and} \bibinfo{person}{Bao Wang}.} \bibinfo{year}{2022}\natexlab{}.
\newblock \showarticletitle{Decentralized federated averaging}.
\newblock \bibinfo{journal}{\emph{IEEE Transactions on Pattern Analysis and Machine Intelligence}} \bibinfo{volume}{45}, \bibinfo{number}{4} (\bibinfo{year}{2022}).
\newblock


\bibitem[Sun et~al\mbox{.}(2024)]%
        {sun_mimic_2024}
\bibfield{author}{\bibinfo{person}{Yuchang Sun}, \bibinfo{person}{Yuyi Mao}, {and} \bibinfo{person}{Jun Zhang}.} \bibinfo{year}{2024}\natexlab{}.
\newblock \showarticletitle{{MimiC}: {Combating} {Client} {Dropouts} in {Federated} {Learning} by {Mimicking} {Central} {Updates}}.
\newblock \bibinfo{journal}{\emph{IEEE Transactions on Mobile Computing}} \bibinfo{volume}{23}, \bibinfo{number}{7} (\bibinfo{date}{July} \bibinfo{year}{2024}), \bibinfo{pages}{7572--7584}.
\newblock
\showISSN{1536-1233, 1558-0660, 2161-9875}


\bibitem[Vanhaesebrouck et~al\mbox{.}(2017)]%
        {vanhaesebrouck_decentralized_2017}
\bibfield{author}{\bibinfo{person}{Paul Vanhaesebrouck}, \bibinfo{person}{Aur{\'e}lien Bellet}, {and} \bibinfo{person}{Marc Tommasi}.} \bibinfo{year}{2017}\natexlab{}.
\newblock \showarticletitle{Decentralized collaborative learning of personalized models over networks}. In \bibinfo{booktitle}{\emph{Artificial Intelligence and Statistics}}. PMLR, \bibinfo{pages}{509--517}.
\newblock


\bibitem[Wang and Xu(2024)]%
        {wang_friends_2024}
\bibfield{author}{\bibinfo{person}{Heqiang Wang} {and} \bibinfo{person}{Jie Xu}.} \bibinfo{year}{2024}\natexlab{}.
\newblock \showarticletitle{Friends to {Help}: {Saving} {Federated} {Learning} from {Client} {Dropout}}. In \bibinfo{booktitle}{\emph{{ICASSP} 2024 - 2024 {IEEE} {International} {Conference} on {Acoustics}, {Speech} and {Signal} {Processing} ({ICASSP})}}. \bibinfo{publisher}{IEEE}, \bibinfo{pages}{8896--8900}.
\newblock
\showISBNx{979-8-3503-4485-1}


\bibitem[Wang et~al\mbox{.}(2019)]%
        {wang_beyond_2019}
\bibfield{author}{\bibinfo{person}{Zhibo Wang}, \bibinfo{person}{Mengkai Song}, \bibinfo{person}{Zhifei Zhang}, \bibinfo{person}{Yang Song}, \bibinfo{person}{Qian Wang}, {and} \bibinfo{person}{Hairong Qi}.} \bibinfo{year}{2019}\natexlab{}.
\newblock \showarticletitle{Beyond {Inferring} {Class} {Representatives}: {User}-{Level} {Privacy} {Leakage} {From} {Federated} {Learning}}. In \bibinfo{booktitle}{\emph{{IEEE} {INFOCOM} 2019}}. \bibinfo{publisher}{IEEE}, \bibinfo{pages}{2512--2520}.
\newblock


\bibitem[Yin et~al\mbox{.}(2021)]%
        {yin_see_2021}
\bibfield{author}{\bibinfo{person}{Hongxu Yin}, \bibinfo{person}{Arun Mallya}, \bibinfo{person}{Arash Vahdat}, \bibinfo{person}{Jose~M Alvarez}, \bibinfo{person}{Jan Kautz}, {and} \bibinfo{person}{Pavlo Molchanov}.} \bibinfo{year}{2021}\natexlab{}.
\newblock \showarticletitle{See through Gradients: Image Batch Recovery via GradInversion}.
\newblock \bibinfo{journal}{\emph{arXiv preprint arXiv:2104.07586}} (\bibinfo{year}{2021}).
\newblock


\bibitem[Zhu et~al\mbox{.}(2022)]%
        {zhu_client_2023}
\bibfield{author}{\bibinfo{person}{Hongbin Zhu}, \bibinfo{person}{Junqian Kuang}, \bibinfo{person}{Miao Yang}, {and} \bibinfo{person}{Hua Qian}.} \bibinfo{year}{2022}\natexlab{}.
\newblock \showarticletitle{Client selection with staleness compensation in asynchronous federated learning}.
\newblock \bibinfo{journal}{\emph{IEEE Transactions on Vehicular Technology}} \bibinfo{volume}{72}, \bibinfo{number}{3} (\bibinfo{year}{2022}), \bibinfo{pages}{4124--4129}.
\newblock


\bibitem[Zhu et~al\mbox{.}(2019)]%
        {zhu_deep_2019}
\bibfield{author}{\bibinfo{person}{Ligeng Zhu}, \bibinfo{person}{Zhijian Liu}, {and} \bibinfo{person}{Song Han}.} \bibinfo{year}{2019}\natexlab{}.
\newblock \showarticletitle{Deep {Leakage} from {Gradients}}.
\newblock \bibinfo{journal}{\emph{Advances in Neural Information Processing Systems}}  \bibinfo{volume}{32} (\bibinfo{date}{Dec.} \bibinfo{year}{2019}).
\newblock


\end{thebibliography}
